%% file: main.tex
\PassOptionsToPackage{table, dvipsnames, xcdraw}{xcolor}
\DocumentMetadata{}
\documentclass[
  manuscript,
  nonacm=false,
  xelatex,
  acmthm,
  authorversion=true,
]{acmart}

\setcopyright{none}
\usepackage{multicol}
\usepackage{caption}
\usepackage{placeins}
\usepackage[english]{babel}
\usepackage[utf8]{inputenc}
\usepackage[english]{babel}
\usepackage{graphicx}
\usepackage{amsmath}
\usepackage{amstext}
\usepackage{amssymb}
\usepackage{latexsym}
\usepackage{dsfont}
\usepackage{mathrsfs}
\usepackage{lscape}
\usepackage{afterpage}
\usepackage{booktabs}
\usepackage{multirow}
\usepackage{longtable}
\usepackage{adjustbox}
\usepackage{array}
\usepackage[ruled,vlined, linesnumbered]{algorithm2e}
\SetKwRepeat{Do}{do}{while}
\SetKw{Return}{Return}
\SetKwComment{Comment}{\#{} }{}
\SetKwInput{KwStore}{Store}

\newcolumntype{R}[2]{%
  >{\adjustbox{angle=#1,lap=\width-(#2)}\bgroup}%
  l%
  <{\egroup}%
}

\newcommand{\review}[1]{{\color{black} #1}}

\newcommand{\transpose}[1]{\ensuremath{#1^\top}}
\newcommand{\Matrix}[1]{\ensuremath{\boldsymbol{\mathbf{#1}}}}
\newcommand{\Vector}[1]{\ensuremath{\boldsymbol{\mathbf{#1}}}}

\newcommand{\ov}\overline
\newcommand{\what}{\widehat}

\newcommand{\diag}{\text{diag}}

\newcommand{\WLS}{\text{WLS}}

\newcommand{\opt}{\ensuremath{\text{opt}}}
\newcommand{\hdmul}{\ensuremath{\odot}}
\newcommand{\hddiv}{\ensuremath{\oslash}}

\newcommand{\argmin}{\mathop{\rm arg\,min}}
\newcommand{\middledot}{\mathbin{\vcenter{\hbox{\scalebox{0.5}{$\bullet$}}}}}

\definecolor{darkred}{rgb}{0.55, 0, 0}
\definecolor{darkcyan}{rgb}{0.0, 0.55, 0.55}
\definecolor{darkpurple}{rgb}{0.55, 0, 0.55}
\definecolor{darkblue}{rgb}{0, 0, 0.55}
\definecolor{darkgreen}{rgb}{0, 0.55, 0}

\newtheorem{theorem}{Theorem}

\newtheorem{remark}[theorem]{Remark}

\newtheorem{definition}[theorem]{Definition}

\begin{document}

\date{January 14, 2026}
\title{Online Distributional Regression}
\author{Simon Hirsch}
\orcid{0009-0008-1409-9677}
\email{simon.hirsch@stud.uni-due.de}
\email{simon.hirsch@statkraft.com}
\affiliation{%
  \institution{Statkraft Trading GmbH}
  \country{Germany}
}
\affiliation{%
  \institution{Data Science in Energy and Environment, University of Duisburg-Essen}
  \country{Germany}
}

\author{Jonathan Berrisch}
\orcid{0000-0002-4944-9074}
\email{jonathan.berrisch@uni-due.de}
\author{Florian Ziel}
\email{florian.ziel@uni-due.de}
\orcid{0000-0002-2974-2660}
\affiliation{%
  \institution{Data Science in Energy and Environment, University of Duisburg-Essen}
  \country{Germany}
}
\renewcommand{\shortauthors}{Hirsch et al.}
\date{\today}

\begin{abstract}%   <- trailing '%' for backward compatibility of .sty file
  Large-scale streaming data are common in modern machine learning applications and have led to the development of online learning algorithms. Many fields, such as supply chain management, weather and meteorology, energy markets, and finance, have pivoted toward probabilistic forecasting. \review{This results in} the need not only for accurate learning of the expected value but also for learning the conditional heteroskedasticity and conditional moments. Against this backdrop, we present a methodology for online estimation of regularized, linear distributional models. The proposed algorithm combines recent developments in online estimation of LASSO models with the well-known GAMLSS framework. We provide a case study on day-ahead electricity price forecasting, in which we show the competitive performance of the incremental estimation combined with strongly reduced computational effort. Our algorithms are implemented in a computationally efficient \texttt{Python} package \review{\texttt{ondil}}.
\end{abstract}

\maketitle

\textbf{Working Paper -- This Version:} January 14, 2026

\section{Introduction}\label{sec:introduction}

Large-scale streaming data are common in modern applications of machine learning and have led to the development of online learning algorithms \citep{cesa2021online}. For processes driven by a high-dimensional covariate space, regularized algorithms have been presented by, e.g., \cite{angelosante2010online, yang2010online, monti2018adaptive, yang2023online}. In many settings, online statistical algorithms are used to issue forecasts. The advent of probabilistic forecasting in many fields, such as supply chain management, weather and meteorology, energy markets, and finance, yields the need not only for accurate learning of the conditional expected value but also for learning the conditional moments \citep[for a review on distributional regression see, e.g. \cite{kneib2023rage} and \cite{klein2024distributional}, for probabilistic forecasting see][]{gneiting2014probabilistic, nowotarski2018recent, alvarez2021probabilistic, petropoulos2022forecasting, ziel2022m5}. However, online learning approaches for distributional regression remain sparse in the literature and can be grouped as:
\begin{enumerate}
  \item Adaptive tracking of the \review{variance-controlling} scale parameter \citep[see e.g.][]{alvarez2021probabilistic, vilmarest2024viking},
  \item Adaptive estimation of the conditional heteroskedasticity \cite[see e.g.][]{priouret2005recursive, dahlhaus2007recursive, hendrych2018self, cipra2018robust,werge2022adavol, wintenberger2024stochastic},
  \item Adaptive conformal prediction approaches \citep[see e.g.][]{zaffran2022adaptive, bhatnagar2023improved, gibbs2024conformal, dutot2024adaptive, brusaferri2024line},
\end{enumerate}
and, to the best of our knowledge, there are no regularized online distributional regression approaches suitable for high-dimensional covariate processes available so far.

In this paper, we provide a regularized online learning algorithm for the conditional distribution parameters of the response variable $Y$ based on a combination of the \textit{online coordinate descent} (OCD) algorithm introduced by \cite{angelosante2010online} and \cite{messner2019online} and the \textit{generalized additive models for location, scale and shape} (GAMLSS) introduced by \cite{rigby2005generalized, stasinopoulos2008generalized, stasinopoulos2018gamlss}. Formally, the GAMLSS framework assumes that \review{ the $N$-dimensional vector of response variables $\Vector{y} = \transpose{(y_1, \ldots, y_N)}$} \review{follows} the probability density function (PDF)
$$f_{y} \left(y_n \mid \mu_n, \sigma_n, \nu_n, \tau_n \right) \quad \review{\Leftrightarrow \quad f_Y\left(y_n \mid \theta_{n,1}, \theta_{n,2}, \theta_{n,3}, \theta_{n,4} \right)} $$
with (up to) four distribution parameters, each of which can be a (linear) function of explanatory variables. The first two parameters, $\mu_n$ and $\sigma_n$, commonly characterize the location and scale of the distribution, while  $\nu_n$ and $\tau_n$ are commonly denoted as the shape parameters describing the skewness and kurtosis. \review{Based on } \cite{rigby2005generalized} we define the parametric GAMLSS as follows:
\begin{equation}\label{eq:y_distributed_f_theta}
  \review{y_n \sim
  \mathcal{F}\left(\theta_{n,1} ,\ldots, \theta_{n,K}\right) = \mathcal{F}\left(\Vector{\theta}_{n, \middledot}\right) }
  %\mathcal{F}\left(\mu_n, \sigma_n, \nu_n, \tau_n \right) \Leftrightarrow \mathcal{F}\left(\theta_{n,1}, \theta_{n,2}, \theta_{n,3}, \theta_{n,4} \right)
\end{equation}
where $\mathcal{F}$ is \review{a user-chosen} distribution with \review{$K$ distribution parameters $\theta_{n,k}$ and $\Vector{\theta}_{n, \middledot} = (\theta_{n,1}, \ldots, \theta_{n,K})$.} \review{Generally, we use subscript $\middledot$ for indexing a column/row of a matrix. Depending on the domain of $Y$, the distribution's $\mathcal{F}$ support should match $\mathbb{R}$, the positive line $\mathbb{R}^+$, or an appropriate subset.} Let $g_k(\middledot)$ be a known, monotonic link function relating a distribution parameter to a predictor $\Vector{\eta}_{\middledot, k}$ by \review{
  \begin{equation}
    g_k(\Vector{\theta}_{\middledot, k}) = \Vector{\eta}_{\middledot, k} = \beta_0 +
    \underbrace{ \sum_{j\in \mathcal{J}_k} {\beta_{k,j} \Vector{x}_{j}}}_{\text{linear part}} +
    \underbrace{ \sum_{j\in\mathcal{B}_k}  b_{k,j}(\Vector{x}_{j})}_{
      %}_{=\sum^{d_{kj}}_{l=1} \beta_{kjl} \phi_{kjl}(\Vector{x}_{j})
    \text{spline part}}
    \label{eq:distreg_model}
  \end{equation}
  where $\Vector{x}_j$ are the columns of the covariate matrix $\Matrix{X}_k$, $\mathcal{J}_k$ is the index set for the linear effects, $\beta_{k,j}$ the distributional regression coefficients to be estimated. $b_{k,j}(\Vector{x}_{j})$ defines non-linear additive effects that can be represented by a linear combination of basis functions, $$b_{k,j}(\Vector{x}_{j}) = \sum^{D_{k,j}}_{l=1} \beta_{k,j,l} \phi_{k,j,l}(\Vector{x}_{j})$$
  for some basis function $\phi_{k,j,l}(\middledot)$ and the chosen set of additive effects $\mathcal{B}_k$; yielding the basis functions $\Vector{\phi}_{k,j}= \{\phi_{k,j,1}, \phi_{k,j,2},\ldots,\phi_{k,j,D_{k,j}}\}$ with $ |\Vector{\phi}_{k,j}| = D_{k,j}$. This class of additive effects includes e.g., B-Splines,
  or (smooth) ReLU-based splines \citep[see e.g.][]{muschinski2022cholesky, kneib2023rage, klein2024distributional} and allows us to remain linear in the coefficients. Therefore, without loss of generality, we can rewrite the model in Equation \ref{eq:distreg_model} as
  \begin{equation}
    g_k(\Vector{\theta}_{\middledot, k}) = \Vector{\eta}_{\middledot, k} = \Matrix{X}_k\Vector{\beta}_k
  \end{equation}
where the model matrix $\Matrix{X}_k = (\Vector{1}, (\Vector{x}_{j})_{j\in \mathcal{J}_k}, (\Vector{\phi}_{kjl}(\Vector{x}_j) )_{j\in \mathcal{B}_k, l=1,\ldots, D_{k,j} }  )$ is of size $N \times J_k$ where $J_k = 1+ |\mathcal{J}_k| + \sum_{j\in \mathcal{B}_k} D_{k,j}$ and $\Vector{\beta}_k$ is the according linear coefficient vector and $|\middledot|$ is the cardinality.}
The \review{parametric GAMLSS model} therefore allows the modeling of all conditional distribution parameters as \review{non-}linear, additive functions of the explanatory variables in $\Matrix{X}_k$. \review{\cite{rigby2005generalized} introduced estimation via iteratively reweighted least squares (IRLS), which has been extended to LASSO-type penalties by \cite{groll2019lasso}. Further regularized estimation approaches have been proposed by \cite{ziel2021gamlss}.}
\review{
  Alternatively, estimation via Bayesian approaches (direct and via Bayesian-IRLS), direct likelihood minimization using stochastic gradient descent \citep[SGD, see e.g.][]{thielmann2024neural,umlauf2024scalable} and gradient-boosted decision trees \citep[see e.g.][]{maerz2019xgboostlss, cevid2022distributional} are popular.
}
\begin{figure}[htb!]
  \Description[Building blocks of the algorithm]{The paper combines two strands of literature by merging online estimation of linear models and iteratively reweighted least squares estimation for distributional regression models.}
  \centering
  \includegraphics[width=0.8\textwidth]{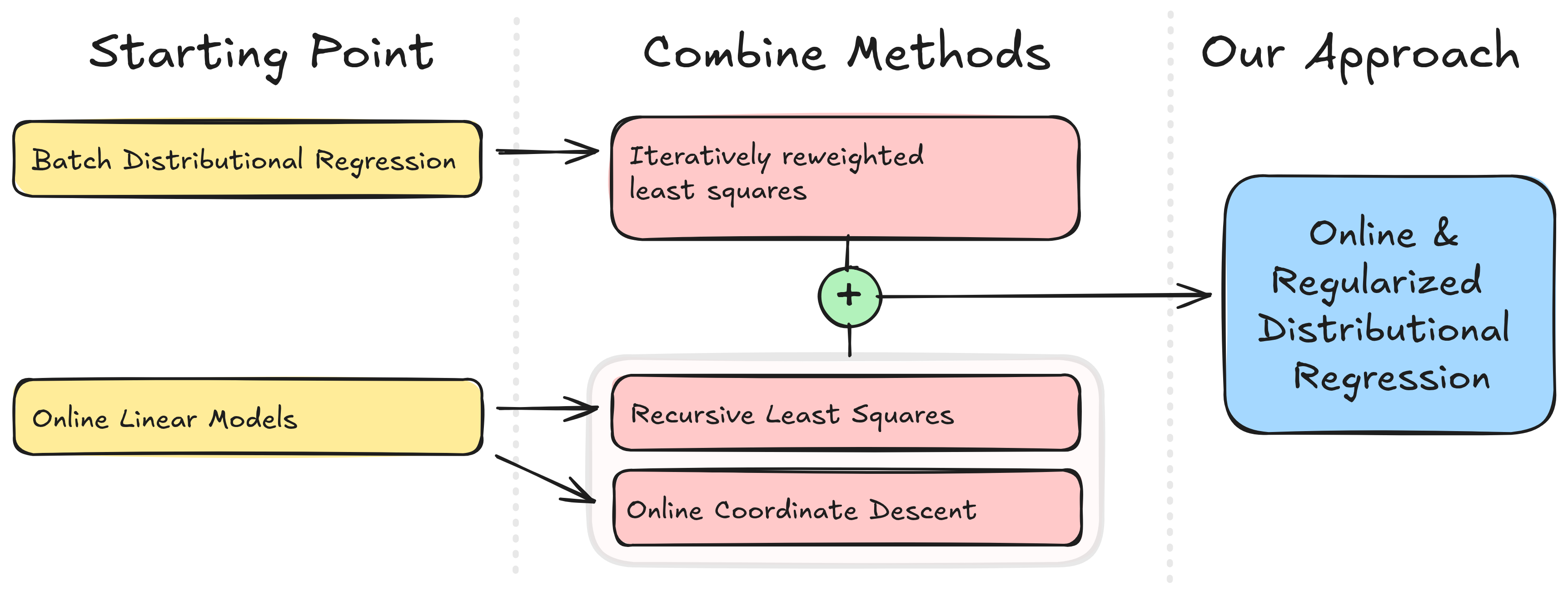}
  \caption{The building blocks of our algorithm give the flow of the paper. Section \ref{sec:batch_gamlss} introduces batch distributional regression and estimation using IRLS, Section \ref{sec:online_regression} introduces incremental estimation methods for linear models. Finally, Section \ref{sec:method} combines both to yield the online distributional regression model.}
  \label{fig:paper_flow}
\end{figure}

\review{The aforementioned methods concern the batch estimation of distributional regression models. However, in many applications, we regularly receive new data. Formally,} after having received the first $N$ observations, we receive a new pair set of data $y_{N+1}$ and $\Vector{x}_{N+1, k}$.
In the online setting, we are interested in updating the \review{coefficients} of the model~$\Vector{\beta}_k$, without recalling all previous observations. We \review{hence} consider an algorithm for a strict online setting in which we can discard all observations after updating the model coefficients. Our proposed algorithm \review{builds on} the IRLS algorithm proposed by \cite{rigby2005generalized} and exploits its agnosticism towards the estimation method \review{ within each iteration. This allows us to exchange the used batch estimation methods with online estimation methods --- including sparse and regularized estimation approaches --- and thereby propose an algorithm suitable for streaming data.}

Figure \ref{fig:paper_flow} gives an idea of the building blocks of our algorithm. In summary, we make the following contributions to the current literature:
\begin{enumerate}
  \item We propose \review{an \emph{online} and \emph{regularized} algorithm enabling} scalable distributional regression for \review{streaming data.
    We implement LASSO regularization ($L_1$-regularization), ridge-based ($L_2$-regularization) and the elastic-net for combined $L_1$ and $L_2$ penalties.}
    We discuss the implementation of the incremental update step, including regularization and online model selection in-depth  (Section \ref{sec:method}).
  \item We formally analyze the relationship between the online estimation and repeated batch estimation and discuss the influence of the exponential discounting, the running time, and the initial sample size. Due to the generic setting of our algorithm, exact results depend on the chosen distributional assumption and link functions. Nevertheless, we aim to provide some intuitive guidance and validate our results in an extensive simulation study (Section~\ref{sec:simulation}).
  \item We validate the proposed method in a forecasting study for electricity prices and demonstrate the competitive performance of our model, combined with a strongly reduced estimation time compared to batch estimation. \review{Furthermore, we conduct ablation studies to analyze the impact of several algorithmic and hyperparameters in a real-world example} (Section~\ref{sec:forecasting}).
  \item We provide an open-source, ready-to-use \texttt{Python} implementation of the online distributional regression model \review{in the package \texttt{ondil}}.
    Our implementation is based on \texttt{numpy} and \texttt{scipy} and employs \texttt{numba} just-in-time compilation for high computational efficiency.
    The code can be accessed on \texttt{GitHub},\footnote{See: \url{https://github.com/simon-hirsch/ondil}.} and the package is available on PyPI\footnote{See: \url{https://pypi.org/project/ondil}.}
    \review{(Section \ref{sec:python})}.
\end{enumerate}

Our work opens multiple avenues for future research. First, our implementation is constrained to \review{parametric GAMLSS models, and hence only allows for non-linear additive effects that can be expressed as a sum of basis functions and linear coefficients. While this class includes a wide array of non-linear effects and can be estimated in a regularized way by the proposed sparse estimation methods,} the inclusion of smooth terms, random effects, and regularized P-splines seems like a worthwhile extension \citep[as implemented in the batch case by][]{rigby2005generalized, eilers1996flexible}. Furthermore, the inclusion of autoregressive and cross-moments effects can provide valuable tools for modeling highly complex \review{time series} processes such as weather, electricity markets, or supply chain applications. Lastly, a thorough theoretical analysis of the error bounds of the proposed online approximation compared to the batch setting should further increase the trust in the presented methods.

\review{The remainder of this paper is structured by the \review{flow given in Figure \ref{fig:paper_flow}.} The following two sections give the building blocks of our algorithm: First, Section \ref{sec:batch_gamlss} briefly reviews the algorithm for the parametric GAMLSS. Afterwards, Section \ref{sec:online_regression} reviews online estimation of regularized linear models. Section \ref{sec:method} presents our main contribution, the online distributional regression. The following Sections \ref{sec:simulation} and \ref{sec:forecasting} present a simulation study on the properties of the online algorithm and our real-world application for energy markets. Section \ref{sec:python} presents our open-source implementation. Finally, Section \ref{sec:conclusion} discusses our results and concludes the paper.}

\review{
\section{Batch Distributional Regression} \label{sec:batch_gamlss}}
\review{We denote scalar values as lowercase letters $a$, vectors as bold lowercase letters $\Vector{a} = (a_1, \ldots, a_N)$ and matrices as bold uppercase letters $\Matrix{A} = (\Vector{a}_1, \ldots, \Vector{a}_M)$. For an arbitrary matrix $\Matrix{A}$, $\Matrix{A}_{n, \middledot}$ denotes accessing the $n$-th row, while  $\Matrix{A}_{\middledot, i}$ denotes accessing the $i$-th column. We denote element-wise or Hadamard multiplication using~$\odot$ and element-wise division using~$\oslash$, that is~$\Vector{a} \oslash \Vector{b} = \Vector{a} \odot \Vector{b}^{-1}$ for vectors.} %We use calligraphic $\mathcal{F}$ and $\mathcal{D}$ to denote arbitrary distributions and reserve $\mathcal{N}$ and $\mathcal{U}$ for the Gaussian and uniform distributions respectively. $\mathcal{L}$ denotes the likelihood function and $\mathcal{\ell}$ the log-likelihood. Some further calligraphic letters are used to denote index sets. Some indices are used throughout the paper: We use $n = 1, \ldots , N, N+1, \ldots$ as index to denote the $n$-th observation from a data stream. $j = 1, \ldots, J$ is the index for the $j$-th covariate. $k = 1, \ldots, K$ is the index for the $k$-th distribution parameter. We start indices generally at 1.}
\review{The following exposition of the IRLS algorithm for fitting distributional regression models largely follows \citet[Appendix B]{rigby2005generalized} and \cite{stasinopoulos2008generalized}. The full algorithm} is outlined in Algorithm \ref{alg:batch_gamlss}. \review{Our algorithm follows the RS-Algorithm (Rigby \& Stasinopoulos Algorithm), although an implementation of the GC-Algorithm (Cole \& Green Algorithm) is possible as well.}
%Let us introduce some notation first. Generally, $\mathcal{L}_\mathcal{D}$ and $l_\mathcal{D}$ denote the likelihood and log-likelihood function of the distribution $\mathcal{D}$ given the data $Y_i$, $l(Y_i \mid \Vector{\theta}_i)$, where $\Vector{\theta} = (\theta_1, \ldots, \theta_k)$ for $k$ distribution parameters.
\review{%
  % \begin{equation*}
  %     Y_n \sim \mathcal{F}(\Vector{\theta}_n)  \quad \text{and }\quad g_k(\Vector{\theta}_k) = \Vector{\eta}_k = \beta_0 + \Matrix{X}_k\Vector{\beta}_k +\review{ \sum_j^{J_k} b_{kj}(\Vector{x}_{kj})} \quad
  % \end{equation*}
  \review{
    We start with the distributional regression model from the introduction given in Equations \ref{eq:y_distributed_f_theta} and \ref{eq:distreg_model}:
    \begin{equation*}
      y_n \sim \mathcal{F}(\Vector{\theta}_{n, \middledot})  \quad \text{and }\quad
      g_k(\Vector{\theta}_{\middledot, k}) = \Vector{\eta}_{\middledot, k} = \Matrix{X}_k\Vector{\beta}_k
    \end{equation*}
  }%
  and, without loss of generality, we assume that the same set of covariates is used for all distribution parameters $k \in 1, \ldots, K$ and therefore $\Matrix{X}_k = (\Vector{1}, (\Vector{x}_{j})_{j\in \mathcal{J}_k}, (\Vector{\phi}_{k,j,l}(\Vector{x}_j) )_{j\in \mathcal{B}_k, l=1,\ldots, D_{kj} }  )$. Let $\Vector{\theta}_k = \Vector{\theta}_{\middledot,k}$, $\Vector{\eta_k} = \Vector{\eta}_{\middledot,k} = \Vector{\eta}_{\middledot,k}(\Vector{\theta}_k)$ and let $\ell(y_n \mid \Vector{\theta}_{n, \middledot}) = \log(\mathcal{L}(y_n \mid \Vector{\theta}_{n, \middledot}))$ denote the log-likelihood function of the assumed distribution $\mathcal{F}$. We use the shorthand $\Vector{\ell} = \Vector{\ell}(\Vector{\theta}) =\ell(\Vector{y} \mid \Vector{\theta})$. Then
  \begin{equation}\label{eq:gamlss_vector_u}
    \Vector{u}_k
    = \dfrac{\partial \Vector{\ell}}{\partial \Vector{\eta}_k}
    = \dfrac{\partial \Vector{\eta}}{\partial \Vector{\theta}_k}
    \odot \left(
      \dfrac{\partial \Vector{\eta}_k}{\partial \Vector{\theta}_k}
    \right)^{-1}
  \end{equation}%
}%
denotes the score vectors. Generally, partial derivatives are evaluated element-wise, hence~$\partial \Vector{\ell} / \partial \Vector{\eta}_k$ is a vector of length $N$ with the elements~$\partial \ell(y_n \mid \Vector{\theta}_{n, \middledot}) / \partial \eta_{nk}$. Let
\begin{equation}\label{eq:gamlss_score}
  \Vector{z}_k = \Vector{\eta}_k + \Matrix{W}_{k,k}^{-1}\Vector{u}_k
\end{equation}
be the adjusted dependent variables, also called the working vector. $\Matrix{W}_{k,s}$ is a diagonal, iterative weight matrix, which can have one of the following forms: \review{
  %  \begin{subequations}
  \begin{align}
    \Matrix{W}^{\text{NR}}_{k,s} &= - \diag \left(\dfrac{
        \partial^2 \Vector{\ell}
      }{
        \partial \Vector{\eta}_k \partial \transpose{\Vector{\eta}}_s
    }\right), \quad %\label{eq:gamlss_weight_a} \\
    \Matrix{W}^{\text{FS}}_{k,s} &= - \diag \left(\mathbb{E}\left[\dfrac{
          \partial^2 \Vector{\ell}
        }{
          \partial \Vector{\eta}_k \partial \transpose{\Vector{\eta}}_s
    }\right] \right), \quad %\label{eq:gamlss_weight_b} \\
    \Matrix{W}^{\text{qNR}}_{k,s} &= - \diag\left( \dfrac{
        \partial \Vector{\ell}
      }{
        \partial \Vector{\eta}_k
      } \hdmul
      \dfrac{
        \partial \Vector{\ell}
      }{
        \partial \Vector{\eta}_s
      }
    \right)     \label{eq:gamlss_weight}
    %\label{eq:gamlss_weight_c}
  \end{align}%
  %  \end{subequations}%
  giving the observed (Newton-Raphson), expected (Fisher-Scoring) or product (quasi Newton-Raphson) score function \citep[see Appendix B of][]{rigby2005generalized}. Note that differentiation is again evaluated element-wise and the full Hessian tensor $\Matrix{H}$ is of size $N \times K \times K$.
  Combining Equations \ref{eq:gamlss_weight} %to \ref{eq:gamlss_weight_c}
  and the RS and CG algorithms gives six possible combinations of the RS and CG algorithms and the three score types. In the following, we focus on the RS algorithm with Fisher scoring, but the other combinations can be implemented similarly.
  Using the chain rule and Fisher's identity, we have that:%
  \begin{equation*}
    \mathbb{E}\left[
      \frac{\partial^2 \Vector{\ell}}{\partial \Vector{\eta}_k^2}
    \right]
    = \mathbb{E}\left[
      \frac{\partial \Vector{\ell}}{\partial \Vector{\eta}_k} \odot
      \frac{\partial \Vector{\ell}}{\partial \Vector{\eta}_k}
    \right]
    = \mathbb{E}\left[
      \frac{\partial \Vector{\ell}}{\partial \Vector{\theta}_k} \odot
      \frac{\partial \Vector{\ell}}{\partial \Vector{\theta}_k}
    \right]
    \hddiv \left(\frac{\partial \Vector{\eta}_k}{\partial \Vector{\theta}_k}\right)^{2}.
  \end{equation*}
  The Fisher information matrix can be recovered as the variance of the score function or as the second derivative of the log-likelihood with respect to the predictor.
  Hence, the adjusted observation vector or working vector $\Vector{z}_k$ reads:
  \begin{align}\label{eq:gamlss_working_vector}
    % This is a tiny bit to big for a one-liner
    \Vector{z}_k
    &= \Vector{\eta}_k + \Matrix{W}_{k,k}^{-1}  \left(\dfrac{\partial
      \Vector{\eta}}{\partial \Vector{\theta}_k}
      \odot \left(
        \dfrac{\partial \Vector{\eta}_k}{\partial \Vector{\theta}_k}
    \right)^{-1}\right) \\
    &= \Vector{\eta}_k +\left(
      \mathbb{E}\left[
        \frac{\partial \Vector{\ell}}{\partial \Vector{\theta}_k} \odot
        \frac{\partial \Vector{\ell}}{\partial \Vector{\theta}_k}
      \right]
    \hddiv \left(\frac{\partial \Vector{\eta}_k}{\partial \Vector{\theta}_k}\right)^{2}\right)^{-1} \hdmul
    \left(\dfrac{\partial
      \Vector{\eta}}{\partial \Vector{\theta}_k} \odot \left(\dfrac{\partial \Vector{\eta}_k}{\partial \Vector{\theta}_k}
      \right)^{-1}
    \right),
  \end{align}%
where we note that the weight matrix $\Matrix{W}_{k,k}$ evaluated at the diagonal only. Therefore its inverse corresponds to $\operatorname{diag}(\Matrix{W}_{k,k})^{-1}$ and the multiplication with the score vector is element-wise.} In the implementation, we regularly employ that:
\begin{equation}
  \dfrac{\partial \Vector{\eta}_k}{\partial \Vector{\theta}_k}
  %  = \dfrac{\partial \Vector{\eta}_k}{\partial g_k^{-1}(\Vector{\eta}_k)}
  %  = \dfrac{1}{\left(\dfrac{\partial g_k^{-1}(\Vector{\eta}_k)}{\partial \Vector{\eta}_k}\right)}
  = \left(\dfrac{\partial g_k^{-1}(\Vector{\eta}_k)}{\partial \Vector{\eta}_k}\right)^{-1}
\end{equation}
where $g_k^{-1}(\Vector{\eta}_k) = \Vector{\theta}_k$ is the inverse of the link function. For pseudo-Newton methods, similar results hold. This allows for a mix-and-match implementation of link functions and distributions, since we only need the Jacobian, potentially the Fisher Information and the first derivative of $g_k^{-1}(\cdot)$ \citep{rigby2005generalized, stasinopoulos2008generalized, stasinopoulos2024generalized}, however, \cite{hirsch2025online} show that true Newton-Raphson methods additionally need the second derivative of $g_k^{-1}(\cdot)$ to allow for a similar mix-and-match implementation of the link functions and distributions.

\begin{algorithm}[h]
  \caption{\review{Iteratively reweighted least squares (IRLS) algorithm for the estimation of distributional regression models, see \cite[][RS-algorithm]{rigby2005generalized}}.}
  \label{alg:batch_gamlss}
  \DontPrintSemicolon
  \KwIn{\text{Distribution $\mathcal{F}$, response variable $\Vector{y}$ and covariates $\Matrix{X}$}}
  Initialize the fitted values $\widehat{\Vector{\theta}}_k^{[0, 0]}$ for $k = 1, \ldots, K$. \;
  Evaluate the linear predictors ${\widehat{\Vector{\eta}}^{[0, 0]}_k = g_k(\widehat{\Vector{\theta}}_k^{[0, 0]})}$ for $k = 1, \ldots, K$. \;
  \For{$r = 1, \ldots, \text{\normalfont until convergence}$}{
    \For{$k = 1, \ldots, K$}{
      Start the inner cycle. \;
      \For{$i = 0, 1, \ldots$}{
        Evaluate $\Vector{u}_k^{[r,i]}$, $\Matrix{W}_{k,k}^{[r,i]}$ and $\Vector{z}_k^{[r,i]}$. \;
        Regress $\Vector{z}_k^{[r,i]}$ on $\Matrix{X}$ using weights $\Matrix{W}_{k,k}^{[r,i]}$ to obtain $\widehat{\Vector{\beta}}_k^{[r,i+1]}$. \;
        Calculate the updated $\widehat{\Vector{\eta}}_k^{[r,i+1]}$ and $\widehat{\Vector{\theta}}_k^{[r,i+1]}$. \;
        Evaluate the convergence.\;
      }
      \review{\textbf{break} if $| \ell(\Vector{y} \mid \Vector{\theta}^{[r, i+1]}) - \ell(\Vector{y} \mid \Vector{\theta}^{[r, i]} )|$ is sufficiently small.} \;
      Set
      $\widehat{\Vector{\beta}}_k^{[r + 1, i]} \gets \widehat{\Vector{\beta}}_k^{[r, i]}$, set
      $\widehat{\Vector{\eta}}_k^{[r + 1, i]} \gets \widehat{\Vector{\eta}}_k^{[r, i]}$, set
      $\widehat{\Vector{\theta}}_k^{[r + 1, i]} \gets \widehat{\Vector{\theta}}_k^{[r, i]}$. \;
    }
    \review{\textbf{break} if $| \ell(\Vector{y} \mid \Vector{\theta}^{[r+1]}) - \ell(\Vector{y} \mid \Vector{\theta}^{[r]}) |$ is sufficiently small.}  \;
  }
  \KwOut{$\widehat{\Vector{\beta}}_k=\widehat{\Vector{\beta}}_{k}^{[r,i]}$, $\widehat{\Vector{\eta}}_k=\widehat{\Vector{\eta}}_{k}^{[r,i]}$ and $\widehat{\Vector{\theta}}_k = \widehat{\Vector{\theta}}_{k}^{[r,i]}$ for all $k=1,\ldots,K$}
\end{algorithm}

\review{The iteratively reweighted least squares algorithm \citep[][]{rigby2005generalized} for the estimation of distributional regression models is given in Algorithm \ref{alg:batch_gamlss} and consists of two nested cycles. Let $r$ and $i$ be the indices for the outer and inner cycle. The outer cycle maximizes the penalized likelihood with respect to $\Vector{\beta}_k$. Within each outer cycle, the algorithm iterates through the distribution parameters $k = 1, \ldots, K$. For each distribution parameter $\Vector{\theta}_k$ the inner cycle consists of repeated, weighted regression of the score {vector~$\Vector{z}_k^{[r,i]}$} on the design matrix ~$\Matrix{X}_k$ using the iterative weights~$\Matrix{W}_{k,k}^{[r,i]}$. We denote iterations as superscript, i.e.~$\Vector{\theta}^{[r, i]}_k$ is the value of $\Vector{\theta}_k$ at the outer iteration~$r$ and the inner iteration~$i$. The weighted fit in Line 8 is at the core of the Newton step for the estimation of the conditional distribution parameters. It is important to see that the general algorithm is agnostic to which (weighted) statistical learning method is employed to regress $\Vector{z}_k^{[r,i]}$ on the design matrix $\Matrix{X}_k$ using the iterative weights $\Matrix{W}_{k,k}^{[r,i]}$ \citep[][see p. 113]{stasinopoulos2024generalized}. This has facilitated the development of multiple extensions to the batch GAMLSS, such as \texttt{gamlss.boost} \citep{mayr2012generalized, hofner2014gamboostlss}, mixtures with Neural Networks \citep{rugamer2024semi}, and regularized approaches as in \cite{groll2019lasso} and \texttt{gamlss.lasso} \citep{ziel2021gamlss}. We do not include the backfitting iteration here as we constrain ourselves to non-linear additive effects that can be represented as a combination of basis functions of~$\Matrix{X}_k$.} % %In Section \ref{sec:method}, we will exploit this agnosticism by using online methods for statistical models, which will be introduced in the following Section \ref{sec:online_regression}.

%\review{\citet[Appendix C]{rigby2005generalized} show that each outer iteration in Algorithm \ref{alg:batch_gamlss} is a Newton-step for the maximization of the global likelihood and therefore, standard results on the convergence of Newton-Raphson algorithms apply. Additionally, they state that \emph{``Step optimization is rarely needed in practice in our experience}". Nevertheless, the convergence of Newton-Raphson algorithms depends on the chosen start values and on the convexity of the log-likelihood function with respect to the distribution parameters. Therefore, we additionally implement automatic step size reduction if the proposed step in the inner iteration decreases the likelihood. By default, the step size is halved if the likelihood is not improved. In our \texttt{Python} implementation, users can also enable extensive logging to monitor and interpret the algorithm's behavior and results. %Furthermore, in the batch case, we use the exact maximum-likelihood estimates as start values, opposed to hard-coded values in the \texttt{R} implementation.}

\review{
  Lastly, let us discuss two important aspects of using iteratively reweighted least squares for the estimation of distributional regression models using a concrete example. The calculation of the score vector, working vector, and weights depends on the distributional assumption and the chosen link function. Remark \ref{remark:gaussian_distributional_regression} introduces a simple Gaussian example and gives the concrete working vectors and weights.

  \begin{remark}\label{remark:gaussian_distributional_regression}
    When considering a distributional regression model for the normal distribution with density $$
    f(x| \mu, \sigma) = %\frac{1}{
    1 / \sqrt{2 \pi \sigma^2} \exp\left(-1/2%\dfrac{
      (x-\mu)^2/
      \sigma^2
    \right), $$
    and using a parameterization of~$\theta_1 = \mu$, $\theta_2= \sigma^2$ and identity link functions~$g_1(\theta)=\theta$ and~$g_2(\theta)=\theta$, the iteratively reweighted least squares algorithm uses the following working vectors   and weights
    \begin{align}\label{eq:working_vector_remark}
      \Vector{z}_1 &= \Vector{z}_{\mu} = \Vector{y} \quad \text{and}
      & \Matrix{W}_{1,1} &= \operatorname{diag} \left((\Vector{\sigma}^2)^{-1}\right)  \\
      \Vector{z}_2 &= \Vector{z}_{\sigma^2} = (\Vector{y} - \Vector{\mu})^2
      % \end{align}
      % \begin{align} \label{eq:weights_remark_1}
      \quad \text{and}   &\Matrix{W}_{2,2} &= \operatorname{diag}\left((2\Vector{\sigma}^4)^{-1}\right).
    \end{align}
  \end{remark}

  The working vectors $\Vector{z}_1, \Vector{z}_2$ and weights $\Matrix{W}_{1,1}$ equal the working vectors used to estimate mean-variance models for conditional heteroskedasticity using least-squares estimation, where $\Matrix{W}_{2,2}$ is often omitted and set as constant $\Matrix{W}_{2,2} = 1$ \cite[see e.g.][Chapter~6]{francq2019garch}. For this approach, consistent model selection using the adaptive LASSO and asymptotic normality of the estimators have been shown by \cite{dette2013least} for independent observations and \cite{ziel2016iteratively} for time series settings. The weights~$\Matrix{W}_{2,2}$ in Remark \ref{remark:gaussian_distributional_regression} additionally correct for heteroskedasticity in the estimation of the conditional scale parameter. \cite{groll2019lasso} discuss LASSO estimation in the general setting of distributional regression in an application and simulation study; however, general results on consistent model selection are, to the best knowledge of the authors, not available.
}

The working vectors (Equation \ref{eq:working_vector_remark} in Remark \ref{remark:gaussian_distributional_regression}) also underscore an important fact in distributional modeling: \review{The models for the conditional distribution parameters depend on each other. Therefore, errors due to low-quality models for one distribution parameter can propagate through the distributional model and adversely affect the estimation of other distribution parameters. This can lead to hard-to-predict effects in the estimation, if the distribution parameters are not information-orthogonal \citep{cox1987parameter}. There is a directional component to this issue, as there is no hope in estimating the conditional scale correctly if the model for the location is misspecified, while the impact of a misspecified scale model on the location is not as severe. For models with conditional skewness and tail behavior, the issue can be even more severe, as the relationships might be highly non-linear. This issue is discussed in detail in the context of the M5 competition in \citet{ziel2022m5}.}

\section{Online Estimation of Regularized Linear Models}\label{sec:online_regression}

Let us now move to the online or streaming setting. For the sake of simplicity, we assume we are in a regular regression \review{to the mean} setting and omit the subscript $k$ for the distribution parameter used in the previous subsection. The issue at hand can be summarized as follows: given some data $\Vector{y}_N=\transpose{(y_1,\ldots, y_N)}$, $\Matrix{X}_N = (\Vector{x}_1, \ldots, \Vector{x}_N)$ and weights $\Matrix{W}_N = \operatorname{Diag}\left(w_1, \ldots, w_N \right)$ we have estimated a set of coefficients $\widehat{\Vector{\beta}}_N$. We are now interested in recovering the coefficients $\widehat{\Vector{\beta}}_{N+1}$, given a new observation for $y_{N+1}$, a new row of $\Vector{x}_{N+1}$ and a new weight $w_{N+1}$. \review{In slight abuse of notation, we employ the subscripts $N$, and $N+1$ to indicate which data is available to the forecaster by now, thereby $\Vector{y}_{N+1} = (y_1, \ldots, y_N, y_{N+1})$ and $\Matrix{X}_{N+1} = (\Vector{x}_1, \ldots, \Vector{x}_N, \Vector{x}_{N+1})$. We employ this notation exclusively around the updating step $N \rightarrow N+1$.} The following two sections review the estimation of weighted linear models using recursive least squares~(Section \ref{sec:recursive_least_squares}) and the incremental estimation of $L_1$ and $L_2$ penalized linear models using online coordinate descent~(OCD, see Section \ref{sec:online_coordinate_descent}).

\subsection{Weighted Recursive Least Squares}\label{sec:recursive_least_squares}

The following is a condensed introduction \cite[e.g.][Chapter 10-12]{haykin2014adaptive} \review{to recursive estimation for weighted least squares estimation}. \review{Nevertheless, it is useful to emphasize a few points that will be of interest for the online distributional learning algorithm introduced in Section \ref{sec:method}.} The weighted least squares regression problem estimates the coefficients $\widehat{\Vector{\beta}}_N^\WLS$ that minimize the loss:
\begin{equation}
  \widehat{\Vector{\beta}}^\WLS_N = \argmin_{\Vector{\beta}} \left\{ \left\lVert \Matrix{W}_N^\frac{1}{2} (\Vector{y}_N - \Matrix{X}_N\Vector{\beta}) \right\rVert^2_2 \right\} = (\transpose{\Matrix{X}_N}\Matrix{W}_N\Matrix{X}_N)^{-1} \transpose{\Matrix{X}_N}\Matrix{W}_N \Vector{y}_N,
\end{equation}
by noting that $\Matrix{W}_N$ is invertible as long as all weights are positive and non-zero. \review{Define the weighted Gramian $\Matrix{G}_N = \transpose{\Matrix{X}_N}\Matrix{W}_N\Matrix{X}_N$.} Under the assumptions that (1) the weights are known and (2) do not change for past observations, the weighted least-squares problem can be re-formulated as a recursive algorithm for the next observations $y_{N+1}$ and covariates. \review{These assumptions are in general not an issue if one is using \emph{sample} weights. We will return to these assumptions in Section \ref{sec:method} for the case of \emph{estimation} weights. In either case, the weights should be living on a consistent scale. By employing the Sherman-Morrison formula (see Appendix \ref{app:gram_update}), we can efficiently update the Gramian matrix by taking
  \begin{align}
    \Matrix{G}_{N+1}^{-1} = \Matrix{G}_{N}^{-1} - \dfrac{
      w_{N+1}\Matrix{G}_{N}^{-1}\transpose{\Vector{x}_{N+1}}\Vector{x}_{N+1}\Matrix{G}_{N}^{-1}
    }{
      1+ w_{N+1}\Vector{x}_{N+1}\Matrix{G}_{N}^{-1}\transpose{\Vector{x}_{N+1}}
    }\label{eq:update_weighted_inverse_gramian}
  \end{align}
  and we receive
  \begin{align}
    \what{\Vector{\beta}}^{\WLS}_{N+1}
    = (\transpose{\Matrix{X}_{N+1}}\Matrix{W}_{N+1}\Matrix{X}_{N+1})^{-1} \transpose{\Matrix{X}_{N+1}}\Matrix{W}_{N+1} \Vector{y}_{N+1}
    = \what{\Vector{\beta}}^{\WLS}_{N} + \Matrix{G}_{N+1}^{-1}  w_{N+1}\transpose{\Vector{x}_{N+1}}\underbrace{\left(y_{N+1} - \Vector{x}_{N+1}\what{\Vector{\beta}}^{\WLS}_{N} \right)}_{\text{Forecast error in }N+1} \label{eq:update_equation_wls}
  \end{align}
where the latter Equation \ref{eq:update_equation_wls} gives a convenient update equation for streaming settings.} An important special case is the recursive least squares with exponential forget, which minimizes the loss
\review{
  \begin{equation}
    \what{\Vector{\beta}}_N = \argmin_{\Vector{\beta}} \left\{ \left\lVert
      %\operatorname{Diag}\left((1 - \gamma)^{N-n}\right)
      \Matrix{\Gamma}_N
      ^\frac{1}{2}
    (\Vector{y}_N- \Matrix{X}_N\Vector{ \beta}) \right\rVert^2_2 \right\},
  \end{equation}
  where $$\Matrix{\Gamma}_N = \operatorname{Diag}\left((1-\gamma)^{N-1}, (1-\gamma)^{N-2}, \ldots,(1-\gamma)^{1}, (1-\gamma)^{0} \right)$$ and $\gamma$ denotes the forget factor.
We} note that the exponential discounting of older observations leads to
\begin{equation}
  \transpose{\Matrix{X}}_{N+1}\Matrix{\Gamma}_{N+1}\Matrix{X}_{N+1} = (1 -\gamma) (\transpose{\Matrix{X}_{N}}\Matrix{\Gamma}_{N}\Matrix{X}_N)+ \transpose{\Vector{x}}_{N+1}\Vector{x}_{N+1}
\end{equation}
and the \review{corresponding} update equation for the exponentially discounted inverted Gram matrix
\begin{equation}
  \begin{aligned}
    (\transpose{\Matrix{X}}_{N+1}\Matrix{\Gamma}_{N+1}\Matrix{X}_{N+1})^{-1}
    = \frac{1}{1 -\gamma} \left((\transpose{\Matrix{X}}_{N}\Matrix{\Gamma}_{N}\Matrix{X}_{N})^{-1} -
      \frac{
        (\transpose{\Matrix{X}}_{N}\Matrix{\Gamma}_{N}\Matrix{X}_{N})^{-1} \transpose{\Vector{x}}_{N+1}\Vector{x}_{N+1} (\transpose{\Matrix{X}}_{N}\Matrix{\Gamma}_{N}\Matrix{X}_{N})^{-1}
      }{
        (1 - \gamma) +  \Vector{X}_{N+1}(\transpose{\Matrix{X}}_{N}\Matrix{\Gamma}_{N}\Matrix{X}_{N})^{-1}\transpose{\Vector{x}}_{N+1}
    }\right)
  \end{aligned}\label{eq:update_exponentially_discounted_inverse_gramian}
\end{equation}
which allows for an efficient online update of the OLS coefficients under exponential forgetting by using Equation \ref{eq:update_equation_wls} and setting $w_{N+1} = 1$. \review{Note the subtle difference between Equation \ref{eq:update_weighted_inverse_gramian} for updating the inverse Gramian matrix with sample weights $\Matrix{W}_N$ and Equation \ref{eq:update_exponentially_discounted_inverse_gramian} using discounting weights $\Matrix{\Gamma}_N$, as the sample weights are constant, but the unique structure of exponential discounting allows the weights $(1-\gamma)^n$ to depend on the distance %$d = (N+1) - n$
to the current newest observation.}

Lastly, we combine sample weights and exponential discounting: For the exponentially discounted, weighted Gram matrix $\Matrix{G}_N = (\transpose{\Matrix{X}_N} \Matrix{\Gamma}_N \Matrix{W}_N \Matrix{X}_N)$, we have
\begin{equation} \label{eq:full_gramian_update}
  \Matrix{G}_{N+1} = (1 - \gamma) \Matrix{G}_{N} + w_{N+1}\transpose{\Vector{x}_{N+1}}\Vector{x}_{N+1}
\end{equation}
and accordingly the update of the inverted Gram matrix $\Matrix{G}_{N+1}^{-1}$ can be written as:
\begin{equation} \label{eq:full_inverted_gramian_update}
  \Matrix{G}^{-1}_{N+1} = \dfrac{1}{1-\gamma}\left( \Matrix{G}^{-1}_{N} -
    \frac{
      w_{N+1} \Matrix{G}^{-1}_{N} \transpose{\Vector{x}}_{N+1}\Vector{x}_{N+1} \Matrix{G}^{-1}_{N}
    }{
      (1 - \gamma) + w_{N+1}\left(\Vector{x}_{N+1}\Matrix{G}^{-1}_{N}\transpose{\Vector{x}}_{N+1}\right)
    }
  \right)
\end{equation}
which can be plugged into Equation \ref{eq:update_equation_wls} to update the coefficients $\widehat{\Vector{\beta}}$, as in Algorithm \ref{alg:online_ols}.

\begin{algorithm}[h]
  \caption{Recursive Weighted Exponentially Discounted Least-Squares}\label{alg:online_ols}
  \DontPrintSemicolon
  \KwIn{New observations $\Matrix{X}_{N+1}, y_{N+1}, w_{N+1}$ and stored $\Matrix{G}^{-1}_N, \what{\Vector{\beta}}_N$.}
  Update $\Matrix{G}^{-1}_N \rightarrow \Matrix{G}^{-1}_{N+1}$ according to Equation \ref{eq:update_weighted_inverse_gramian} resp. \ref{eq:update_exponentially_discounted_inverse_gramian}. \;
  \review{Update $\what{\Vector{\beta}}_N \rightarrow \what{\Vector{\beta}}_{N+1}$ according to Equation \ref{eq:update_equation_wls}.} \;% $\what{\Vector{\beta}}_{N+1} = \widehat{\Vector{\beta}}_N + \Matrix{G}^{-1}_{N+1} w_{N+1} \transpose{\Vector{x}}_{N+1} \left(y_{N+1} - \Vector{x}_{N+1} \what{\Vector{\beta}}_N\right)$\;
  %  \KwStore{Updated Gramian $\Matrix{G}^{-1}_{N+1}$.}
  \KwOut{$\what{\Vector{\beta}}_{N+1}$.}
\end{algorithm}

\subsection{Online Coordinate Descent}\label{sec:online_coordinate_descent}

\review{The recursive least squares estimator can be numerically unstable due to its reliance on the inverted Gramian matrix $\Matrix{G}_N^{-1}$. Furthermore, for regression problems with large sets of potential covariates, some regularization is commonly employed to keep a parsimonious model. The elastic net \citep{friedman2010regularization, friedman2007pathwise} minimizes a combination of $L_1$ (or LASSO) and $L_2$ (or Ridge) penalties on the coefficient vector and is commonly written as:
  \begin{equation}
    \what{\Vector{\beta}}_N
    = \argmin_{\Vector{\beta}} \left\{ \left\| \Matrix{W}_N^\frac{1}{2}  \left( \Vector{y}_N - \Matrix{X}_N \Vector{\beta} \right) \right\|_2^2
      + \alpha \lambda\left\| \Vector{\beta} \right\|_1
      +  (1 - \alpha) \lambda \left\| \Vector{\beta} \right\|_2
    \right\}.
  \end{equation}
where the ratio between $L_1$ and $L_2$ penalties is governed by $\alpha$. From the elastic net, we can recover ordinary least squares ($\lambda = 0$), the LASSO ($\alpha=1$), and the ridge regression ($\alpha=0$). In the batch setting, }
\citet{friedman2007pathwise, friedman2010regularization} introduced path-wise cyclic coordinate descent (CCD) for elastic net regression problems, \review{and due to its computational efficiency, it remains the state-of-the-art method for coefficient estimation}. For given regularization parameters \review{$\lambda_1$ and $\lambda_2$}, we repeatedly update the coefficient vector $\Vector{\beta}_N = (\beta_{N,1}, \ldots, \beta_{N,J})$ of length $J$ and update
\review{
  \begin{equation}
    \widehat{\beta}_{N,j} \xleftarrow{} \frac{S\left( \sum_{n=1}^{N} w_n x_{n,j}(y_n - \tilde{y}_n^{(j)}),  \alpha \lambda \right)}{\sum_{n=1}^{N} w_n  x_{n,j}^2 + ( 1 - \alpha) \lambda }\label{eq:lasso_update_batch} % + \lambda ( 1 - \alpha)} % LASSO only
\end{equation}}%
where $S(\beta, \lambda) = \text{sign}(\beta) \max(\lvert \beta \rvert - \lambda, 0)$ is the soft-thresholding function \review{and $y_n - \tilde{y}_n^{(j)} = y_n - \widehat{y}_n + x_{nj}\widehat{\beta}_j$ is the partial residual excluding the contribution of $x_{nj}$}. Note that, implicitly, \cite{friedman2010regularization} assume that $\sum_{n=1}^N w_n = 1$. %Commonly, $\alpha = 1$ refers to the LASSO regression and $\lambda = 1$ refers to the ridge regression.
We use a numerical convergence criterion to \review{terminate} the algorithm after convergence.

As proposed in \cite{angelosante2010online} and  \cite{messner2019online} for LASSO estimation,  we rewrite the update Equation \ref{eq:lasso_update_batch} to calculate the updated parameter from the weighted Gramian matrix $\Matrix{G}_N = \transpose{\Matrix{X}}_N\Matrix{W}_N\Matrix{X}_N$ and the \review{vector $\Vector{h}_N = \transpose{\Matrix{X}}_N\Matrix{W}_N\Vector{y}_N$} and their potentially exponentially discounted counterparts (see Section \ref{sec:online_regression} and Equation \ref{eq:full_gramian_update}).
\review{Since $\Matrix{G}_N$ and $\Vector{h}_N $ can be updated online, this allows an efficient incremental estimation on a data stream.}
%\review{Moving from LASSO to elastic net penalties in the online estimation is straightforward by including the terms $\alpha \lambda$ in the nominator and $ ( 1 - \alpha) \lambda$ in the denominator.}
\review{We observe for the elements
  $G_{N,j,j}$ and $h_{N,j}$ of $\Matrix{G}_{N}$ and $\Vector{h}_{N}$ that
  \begin{align*}
    G_{N,j,j} = \sum_{n=1}^N w_n x_{n,j}^2  && \text{and} && {h}_{N,j} =\sum_{n}^{N} w_n x_{n,j}y_n .
  \end{align*} Thus, writing the nominator inside the soft-thresholding function as
  \begin{align*}
    \sum_{n=1}^{N} w_n x_{n,j}\left(y_n - \tilde{y}_n^{(j)}\right) &= \sum_{n=1}^{N} w_n x_{n,j}\left(y_n - \widehat{y}_n + x_{nj}\widehat{\beta}_j\right)
    = \sum_{n=1}^{N} w_n x_{n,j}\left(y_n - \sum_i^Jx_{ni}\widehat{\beta}_i + x_{nj}\widehat{\beta}_j\right) \\
    &= \underbrace{{\sum_{n=1}^{N} w_n x_{n,j}y_n}}_{=h_{N,j} }
    - \underbrace{\sum_i^J \sum_n^N w_n x_{n,j} x_{n,i}\beta_i}_{=\Matrix{G}_{N,j, \middledot} \what{\Vector{\beta}}_{N}} % \sum_{n=1}^{N+1} w_n x_{n,j}\widehat{y}_n
    + \underbrace{\sum_{n=1}^{N} w_n x_{n,j}^2\widehat{\beta}_{N,j}}_{=G_{N, j, j}\widehat{\beta}_{N,j}}
  \end{align*}
and} we receive the \review{update equation for the elastic net as}
\review{
  \begin{equation}
    \widehat{\beta}_{N,j} \xleftarrow{} \frac{%
      S\left(h_{N,j} - \Matrix{G}_{N, j, \middledot} \what{\Vector{\beta}}_{N} + G_{N,j, j}\widehat{\beta}_{N,j},   \alpha  \lambda \right)
    }{%
      G_{N,j,j}+ ( 1 - \alpha) \lambda
    }. \label{eq:lasso_update_online}
\end{equation}}%
Therefore, the algorithm only needs to store the $J \times J$ {{matrix $\Matrix{G}_N$}} and the {{$J$-dimensional vector\;$\Vector{h}_N$}}. \review{Once a new pair of observations arrives, we can update $\Matrix{G}_{N+1}$ and $\Vector{h}_{N+1}$ and re-run the cyclic coordinate descent based on the previous estimated coefficients.}  \review{As in batch} coordinate descent, we can run the algorithm for a decreasing sequence of regularization strengths $\lambda$, starting with $\lambda_\text{max} = \max{}\lvert \Vector{h}_N \rvert$ as the element-wise maximum in $\Vector{h}_N$ and using an exponential grid towards $\lambda_\text{min} = \epsilon \lambda_\text{max}$ with $\epsilon_\lambda = 0.001$ as typical values \citep{friedman2010regularization, friedman2007pathwise, messner2019online}. \review{Note that, as $\Vector{h}_N$ changes in the online setting with each new pair of observations, the regularization grid varies slowly.} Algorithm \ref{alg:online_lasso} presents the algorithm schematically.

\begin{algorithm}[htb]
  \caption{\review{Online Elastic Net Regression using Online Coordinate Descent Estimation}, see \cite{angelosante2010online} and \cite{messner2019online}}\label{alg:online_lasso}
  \DontPrintSemicolon
  \KwIn{New observations $\Matrix{x}_{N+1}, y_{N+1}, w_{N+1}$ and stored $\Matrix{G}_N, \Vector{h}_N$.}
  Update $\Matrix{G}_N \rightarrow \Matrix{G}_{N+1}$ and $\Vector{h}_N \rightarrow \Vector{h}_{N+1}$ according to Equation \ref{eq:full_gramian_update}. \;
  Update $\lambda_{\text{max}} = \max{}\lvert \Vector{h}_{N+1} \rvert$ and initialize $\Vector{\lambda}$ as exponential grid. \;
  \For{$\lambda \in \Vector{\lambda}$}{
    Set starting coefficients $\widehat{\Vector{\beta}}_{N+1,\lambda} \gets \widehat{\Vector{\beta}}_{N,\lambda}$ \;
    \While{\text{\normalfont{not converged}}}{
      \ForAll{$j \in 1, \ldots, J$}{
        \text{Update $\widehat{\beta}_{N+1, \lambda, j}$ according to Equation \ref{eq:lasso_update_online}} \;
      }
      Check convergence for $\widehat{\Vector{\beta}}_{N+1, \lambda}$  and proceed to next $\lambda$ if converged. \;
    }
  }
  %  \KwStore{Updated Gramian $\Matrix{G}_{N+1}$, new vector $\Vector{h}_{N+1}$ and coefficient vector $\widehat{ \Vector{\beta}}_{N+1, \lambda}$.}
  \KwOut{ %$\widehat{\Vector{\beta}}_{N+1} = %\transpose{\left(\widehat{\beta}_{N+1, \lambda}, \ldots\right)}
  $\widehat{ \Vector{\beta}}_{N+1, \lambda}$ for all $\lambda \in \Vector{\lambda}$}
\end{algorithm}

\review{We extend the formulation to allow for the possibility to box-constrain $L_j \leq \beta_j \leq U_j$ with user-chosen bounds $L_j$ and $U_j$. As in the batch implementation in the \texttt{glmnet} \texttt{R} package \citep{friedman2010regularization, tay2023elastic}, Equation \ref{eq:lasso_update_online} is then formulated as:
  \begin{equation*}
    \widehat{\beta}_{N,j} \xleftarrow{} \operatorname{clip}\left(
      \frac{%
        S\left(h_{N,j} - \Matrix{G}_{N, j, \middledot} \what{\Vector{\beta}}_{N} + G_{N,j, j}\widehat{\beta}_{N,j},   \alpha  \lambda \right)
      }{%
        G_{N,j,j}+ ( 1 - \alpha) \lambda
    }, L_j, U_j\right).
\end{equation*}}%

\section{Online Distributional Regression} \label{sec:method}

Having established the batch IRLS algorithm for the estimation of distributional regression models and the online coordinate descent for regularized linear models, this section proceeds by putting the pieces together and presents the online linear \review{distributional regression} model. \review{Algorithm \ref{alg:online_gamlss} outlines the main procedure and we will subsequently describe the details with respect to model selection, mini-batch updates, stopping criteria, and the possible approximation error towards the batch case. We present the algorithm using a path-based, regularized estimation method in each inner iteration, and therefore include model selection; however, using RLS or OCD for unregularized estimation is possible as well.}

\subsection{Update Algorithm}

\review{We start again with the distributional regression model from the introduction given in Equation \ref{eq:y_distributed_f_theta} and \ref{eq:distreg_model}:
  \begin{equation*}
    y_n \sim \mathcal{F}(\Vector{\theta}_{n, \middledot})  \quad \text{and }\quad
    g_k(\Vector{\theta}_{\middledot, k}) = \Vector{\eta}_{\middledot, k} = \Matrix{X}_k\Vector{\beta}_k
\end{equation*}}%
After having seen $N$ observations, we have estimated coefficients $\widehat{\Vector{\beta}}_{N, k}$ for $k = 1, .., K$ distribution parameters. Given some new data $(y_{N+1}, \Matrix{x}_{N+1})$, we aim to update our coefficients towards $\widehat{\Vector{\beta}}_{N+1, k}$. On a high level, for updating the coefficients in the online distributional regression, we do the following steps in the outer cycle: we initialize the predictors ${\eta}_{N+1, k}^{[0, 0]} = g_k\left(\Vector{x}_{N+1, k} \widehat{\Vector{\beta}}_{N, k}\right)$ and this allows us to evaluate  $\Vector{u}_{N+1, k}^{[0, 0]}$, ${w}_{N+1, k,k}^{[0, 0]}$ and ${z}_{N+1, k}^{[0, 0]}$. From there, we can update the weighted, exponentially discounted Gramian matrices $\Matrix{G}_{N, k} \rightarrow \Matrix{G}_{N+1, k}^{[0, 0]}$ and the vectors $\Vector{h}_{N, k} \rightarrow \Vector{h}_{N+1, k}^{[0, 0]}$ \review{by taking
  \begin{align}
    \Matrix{G}_{N+1, k}^{[r,i]} &= \left(1-\gamma\right)\Matrix{G}_{N, k} + w_{N+1, k,k}^{[r,i]}\left( \transpose{\Vector{x}}_{N+1} \Vector{x}_{N+1}      \right) \label{eq:update_gram_gamlss}  \\
    \Vector{h}_{N+1, k}^{[r,i]} &= \left(1-\gamma\right)\Vector{h}_{N, k} + w_{N+1, k,k}^{[r,i]}\left( \transpose{\Vector{x}}_{N+1} z_{N+1, k}^{[r,i]}     \right) \label{eq:update_h_gamlss}
\end{align}}%
and subsequently use an online estimation method from Section \ref{sec:online_regression} to update the coefficients inside the iteration. As in the batch algorithm (see Algorithm \ref{alg:batch_gamlss}), we run the outer and inner cycles until convergence.

\begin{algorithm}[!ht]
  \caption{Online update for (regularized) distributional regression models.}\label{alg:online_gamlss}
  \small
  \DontPrintSemicolon
  \review{\KwIn{\text{Distribution $\mathcal{F}$, new data $({y}_{N+1}, \Matrix{x}_{N+1})$, stored $\Matrix{G}_{N, k}, \Vector{h}_{N, k}$ }}}
  Initialize the fitted values $\widehat{\theta}_{N+1, k}^{[0, 0]} = \Vector{x}_{N+1, k}\widehat{\Vector{\beta}}_{N, k}$ for $k = 1, \ldots, p$.  \;
  Evaluate the linear predictors ${\widehat{\eta}^{[0, 0]}_{N+1, k} = g_k(\widehat{\theta}_{N+1, k}^{[0, 0]})}$ for $k = 0, \ldots, p$. \;
  \For{$r = 1, \ldots, \text{\normalfont until convergence}$}{
    \For{$k = 1, \ldots, K$}{
      \For{$i = 0, 1, \ldots$}{
        Evaluate $u_{N+1,k}^{[r,i]}$, $w_{N+1,k,k}^{[r,i]}$ and $z_{N+1, k}^{[r,i]}$ (Eq. \ref{eq:gamlss_score}, \ref{eq:gamlss_weight} and \ref{eq:gamlss_working_vector}). \;
        \review{Update $\Matrix{G}_{N+1, k}^{[r,i]}$ and $\Vector{h}_{N+1, k}^{[r,i]}$ using $z_{N+1, k}^{[r,i]}$, weights $w_{N+1, k,k}^{[r,i]}$ (Eq. \ref{eq:update_gram_gamlss} and \ref{eq:update_h_gamlss})}. \;
        \review{Use OCD to update $\widehat{\Vector{\beta}}_{N+1,k, \lambda} \rightarrow  \widehat{\Vector{\beta}}_{N+1, k, \lambda}^{[r,i+1]}$ based on $\Matrix{G}_{N+1, k}^{[r,i]}$ and $\Vector{h}_{N+1, k}^{[r,i]}$.}\;
        Select the optimal $\lambda$ using IC and set $\widehat{\Vector{\beta}}_{N+1,k}^{[r,i+1]} = \widehat{\Vector{\beta}}_{ N+1, k, \lambda^\text{opt}}^{[r,i+1]}$. \;
        Calculate the updated $\widehat{\Vector{\eta}}_{N+1, k}^{[r, i+1]}$ and $\widehat{\Vector{\theta}}_{N+1, k}^{[i,r+1]}$ \;
        Evaluate the convergence.\;
      }
      \review{\textbf{break} if $| \ell(\Vector{y} \mid \Vector{\theta}^{[r, i+1]}) - \ell(\Vector{y} \mid \Vector{\theta}^{[r, i]} )|$ is sufficiently small.} \;
      Set
      $\widehat{\Vector{\beta}}_{N+1, k}^{[r + 1, 0]} \gets \widehat{\Vector{\beta}}_{N+1, k}^{[r, i]}$, set
      $\widehat{\Vector{\eta}}_{N+1, k}^{[r + 1, 0]} \gets \widehat{\Vector{\eta}}_{N+1, k}^{[r, i]}$ and set
      $\widehat{\Vector{\theta}}_{N+1, k}^{[r + 1, 0]} \gets \widehat{\Vector{\theta}}_{N+1, k}^{[r, i]}$.
    }
    \review{\textbf{break} if $| \ell(\Vector{y} \mid \Vector{\theta}^{[r+1]}) - \ell(\Vector{y} \mid \Vector{\theta}^{[r]} )|$ is sufficiently small.} \;
  }
  %  \KwStore{Updated Gramian $\Matrix{G}_{N+1, k}$ and $\Vector{h}_{N+1, k}$ and coefficient vector $\widehat{\Vector{\beta}}_{N+1, k}$, model selection information.}
  \KwOut{$\widehat{\Vector{\beta}}_{N+1, k}= \widehat{\Vector{\beta}}_{N+1, k}^{[r,i]}$, $\widehat{\Vector{\eta}}_{N+1, k} = \widehat{\Vector{\eta}}_{N+1, k}^{[r,i]}$ and $\widehat{\Vector{\theta}}_{N+1, k}=\widehat{\Vector{\theta}}_{N+1, k}^{[r,i]}$ for all $k=1,\ldots, K$}
\end{algorithm}

\review{The following paragraphs discuss some points of the algorithm in detail.} To make the discussion clearer, we call an \emph{update step} a full update for all distribution parameters $\widehat{\Vector{\theta}}_{N, k} \rightarrow \widehat{\Vector{\theta}}_{N+1, k}$ and associated regression coefficients $\widehat{\Vector{\beta}}_{N, k} \rightarrow \widehat{\Vector{\beta}}_{N+1, k}$ for $k = 1, \ldots, K$, while \emph{step} might refer to any arbitrary step in the algorithm. \review{We start with a few general points and subsequently discuss the issue of model selection (Section~\ref{sec:online_model_selection}) and the difference between batch and online estimation (Section~\ref{sec:batch_online_theory}) in detail.}
\begin{itemize}
  \item Note that in Line 8, in each inner iteration, we start at the Gramian matrix of the previous full fit, not at the previous iterations $r$ or $i$ of the algorithm, since this would imply adding the $\Vector{x}_{N+1, k}$ and $z_{N+1, k}^{[r,i]}$ multiple times to the Gramian matrices within one update step. However, we can (and should) warm-start the OCD algorithm using the coefficient path from the previous iterations within each inner iteration of the update step.
  \item We can potentially use different forget factors for $\gamma_k$ for each distribution parameter. Generally, a higher forget leads to faster adaptation of the coefficients. However, since the estimation of higher moments depends on the estimation of the first moment(s), we note that too aggressive adaptation of the coefficients of the location will lead to an \emph{underestimation} of the conditional heteroskedasticity and potentially higher moments.
  \item \review{The algorithm is able to process one-step and batch updates of size $b$. In the case of mini-batch updates, all scalar values~$u_{N+1, k}^{[r,i]}$, $w_{N+1,k,k}^{[r,i]}$ and~$z_{N+1, k}^{[r,i]}$ need to be replaced by their vector counterparts. The Gramian $\Matrix{G}_{ N+b,k}$ and $\Vector{h}_{N+b,k}$ can be updated in the same fashion as above, and the inverse $\Matrix{G}_{N+b,k}^{-1}$ can be updated using the Woodbury matrix identity. After the update of $\Matrix{G}_{N+b,k}$ and $\Vector{h}_{N+b,k}$, the algorithm remains unchanged.}
  \item \review{As a related point, for one-step updates, this can lead to identification issues: Assume we observe a new, somewhat extreme new observation $y_{N+1}$. Whether this new observation warrants an update of the scale, tail, or skewness parameter is unclear. For the presented Algorithm \ref{alg:online_gamlss}, we observe that the largest updates are generally placed on the distribution parameters that are updated first, and therefore some caution with respect to the ordering should be taken (especially with respect to the parameters for skewness and (symmetric) tail behavior).}
  \item The convergence criteria is the change of the deviance $\mathcal{d} = -2 \sum^N_{n1=} \ell(y_n \mid \hat{\Vector{\theta}}_n)$ given the current iteration's fitted values, both for the batch and online case. We track the (exponentially discounted) deviance in the online case. We employ a relative and absolute stopping criteria.
  \item \review{Algorithm \ref{alg:online_gamlss} is constant in memory use with respect to the number of observations~$N, N+1, N+2, \ldots$ seen.
    In the non-regularized case, the algorithm needs to store $\Matrix{G}_{N,k}$ of size~$J \times J$, the vector~$\Vector{h}_{N,k}$ of size~$J \times 1$ and the coefficient vectors~$\widehat{\Vector{\beta}}_{N,k}$ of size~$J \times 1$ for each distribution parameter~$k = 1, \ldots, K$, which gives a total space complexity of~$\mathcal{O}(K(J^2+2J))$. In the regularized case, we additionally store the estimation path (for warm-starting) and data for the model selection (see the following paragraph).}
\end{itemize}

\subsection{Online Model Selection}\label{sec:online_model_selection}
\review{As already hinted at in Section \ref{sec:online_coordinate_descent}, the estimation of the coefficients $\Vector{\beta}_k$ on a path of regularization parameters yields the need for model selection.} Since distributional regression is a likelihood-based approach, we propose the use of information criteria (IC) for this task. We define the generalized information criterion (GIC) as
\begin{equation}
  \operatorname{GIC}(\widehat{\mathcal{L}}, \nu) = - 2 \log(\widehat{\mathcal{L}}) + \nu_0 P + \nu_1 P \log(\tilde{N}) + \nu_2 P \log(\log(\tilde{N}))
\end{equation}
where $P$ denotes the number of estimated \review{non-zero coefficients}, $\nu = (\nu_0, \nu_1, \nu_2)$ denotes a triplet of parameters, $\tilde{N}_k = \frac{(1 - \gamma_k^N)}{1 - \gamma_l}$ is the effective training length, and $\widehat{\mathcal{L}}$ denotes the maximized value of the likelihood function of the model. The most commonly used information criteria Akaike's Information Criterion {(AIC, $\nu = (2, 0, 0)$)}, Bayesian Information Criterion {(BIC, $\nu = (0, 1, 0)$)} and the Hannan-Quinn Criterion {(HQC, $\nu = (0, 0, 2)$)} can be recovered from the GIC \citep{kim2012consistent, kock2016consistent}. \review{From here, we have two options to proceed: we can use \emph{local} model selection on the level of the distribution parameter $k$, based on the residual sum of squares (RSS) of each inner iteration (Line 8 in Algorithm \ref{alg:batch_gamlss} and Line 9 in Algorithm \ref{alg:online_gamlss}) or \emph{global} model selection, based on the log-likelihood under the full model.
  \begin{enumerate}
    \item \emph{Local Model Selection:} Under the Gaussian assumption, a local likelihood for each IRLS iteration can be formulated as a function of the residual sum of squares of the local regression of $\Vector{z_k}$ on the design matrix
      \begin{equation}\label{eq:rss_likelihood}
        \log(\widehat{\mathcal{L}}_k) = - \frac{\tilde{N}_k}{2}\log\left(\frac{\sum_{n=1}^N(1-\gamma_k)^{N-n-1}(\Vector{z}_k - \Matrix{X}_k\what{\Vector{\beta}}_k)^2}{\tilde{N}_k}\right) + C_k
      \end{equation} where $N$ is the number of observations received so far and $C_k = - \frac{\tilde{N}_k}{2} \left(1 + \log(2 \pi) \right)$ is a constant which only depends on the data and hence can be neglected if the data underlying the model selection is the same for all models. This allows an efficient online update of the information criterion for each distribution parameter $\Vector{\theta}_k$ individually.
      We then choose $$\lambda_k^\opt = \argmin_{\Vector{\lambda}_k} \operatorname{GIC}(\widehat{\mathcal{L}_k}(\lambda_k), \nu_k).$$
    \item \emph{Global Model Selection:} Here, we employ the log-likelihood of the distribution $\mathcal{F}$ directly in the GIC and select the optimal regularization parameter for each by taking: $$\lambda_k^\opt = \argmin_{\Vector{\lambda}_k} \operatorname{GIC}(\widehat{\mathcal{L}}(\lambda_k), \nu).$$
  \end{enumerate}
The local model selection is used, e.g., for the selection of regularization parameters in the backfitting step in the classic GAMLSS algorithm and in the \texttt{R} \texttt{gamlss.lasso} package \citep{ziel2021gamlss}. It has the advantage of avoiding expensive likelihood evaluations, but relies on the assumptions that the local fit's residuals are approximately Gaussian distributed. Furthermore, the local model selection allows specifying different parameterizations of the GIC for different distribution parameters.} We suggest selecting a \review{more conservative parameterization of the GIC} for higher $k$ to avoid overfitting in modeling the conditional scale, kurtosis, and skewness \review{when using \emph{local} model selection. We also note that since the coefficients and estimates for different distribution parameters live on very different scales, it can be the case that $\varepsilon_{k, \lambda}$ needs to be smaller, otherwise the lower end of the coefficient path might not approach the OLS solution.} \cite{groll2019lasso} analyze the impact of different shrinkage parameters on batch distributional regression. Also \cite{marcjasz2023distributional} see the advantages of different regularization for the distribution parameters when using distributional neural networks for time series forecasting.

\subsection{Comparison of Online and Batch Estimation}\label{sec:batch_online_theory}

An important relationship to the batch version of our proposed algorithm becomes apparent when comparing how weights are updated. In the batch setting, with repeated fits of the \review{model using all observations}, the weight matrix $\Matrix{W}_{k,k}^{[r,i]} = \operatorname{Diag}\left(w_{0,k,k},\ldots, w_{N,k,k}\right)$ is updated in every iteration $r,i$ and again, in the next batch fit for $N+1$, the weight matrix is updated for \emph{all} $w_{n,k,k}$ from $n = 1,\ldots, N+1$ in all iterations. In the online setting, $\Matrix{W}_{N,k,k}$ is the weight matrix after convergence of the update step $N-1 \rightarrow N$. In the update step $N \rightarrow N+1$, we cannot update $\Matrix{W}_{N,k,k}$ anymore (see also the Assumptions noted in Section \ref{sec:online_regression}). Therefore, we set
\begin{equation}\label{eq:online_weight_matrx}
  \Matrix{W}_{N+1, k,k}^{[r,i]} =
  \begin{pmatrix}
    (1-\gamma)\Matrix{W}_{N,k,k} & \Matrix{0} \\
    \Matrix{0} & w_{N+1, k,k}^{[r,i]}
  \end{pmatrix}
\end{equation}
and can only update $w_{N+1, k,k}^{[r,i]}$. Note that due to the update rule in Equation \ref{eq:full_gramian_update}, this effect is counter-weighted by the exponential forget $\gamma$. However, there is a delicate balance to strike to trade off the beneficial effect of the forget and increased instability in the coefficient estimation \review{due to smaller effective sample sizes}. This might lead to slower convergence compared to batch learning if the data is drawn from a stationary process. The online distributional regression is, therefore, an \emph{approximation} of the repeated batch estimation, contrary to the case for, e.g., RLS or online coordinate descent in the case of regression to the mean, where the update step leads to the equivalent results. When choosing a specific distribution assumption for~$\mathcal{F}$ in the IRLS framework, the weight matrix for the \review{distribution parameter}~$\theta_k$ in each update step can be recovered with the help of Equations \ref{eq:gamlss_vector_u} to \ref{eq:gamlss_working_vector}. The online weight matrix is retrieved by iteratively inserting the weight matrix in Equation \ref{eq:online_weight_matrx}.
For illustration, building on the standard linear-Gaussian case for the distributional regression in Remark~\ref{remark:gaussian_distributional_regression}, we can examine the difference explicitly for the mean equation:
\review{
  \begin{remark} \label{remark:weights_online}
    Assume a linear-Gaussian distributional regression with $\theta_1 = \mu$ and $\theta_2 = \sigma^2$ of Remark \ref{remark:gaussian_distributional_regression}. Assume that we have seen $n = 1, \ldots., N, N+1, \ldots, M$ observations, where $N$ denotes the initial fit and the steps $N+1, \ldots, M$ are updated online. In the batch and online setting, the weight matrices for $\theta_1=\mu$ for the step $(M-1) \rightarrow M$ are:
    \begin{small}
      \begin{align*}
        \Matrix{W}_{M,1,1}^{[r,i], \operatorname{online}} &= \operatorname{diag}\Biggl(%
          \underbrace{
            \dfrac{(1- \gamma)^{M-1}}{\widehat{\sigma}^2_{1|N}}, \ldots,
            \dfrac{(1-\gamma)^{M-N-1}}{\widehat{\sigma}^2_{N|N}}
          }_\text{\normalfont Initial batch $n=1, \ldots, N$.},
          \underbrace{
            \dfrac{(1-\gamma)^{M-N-2}}{\widehat{\sigma}^2_{N+1|N+1}} , \ldots,
            \dfrac{(1-\gamma)}{\widehat{\sigma}^2_{M-1|M-1}},
            \dfrac{1}{(\widehat{\sigma}^2_{M|M})^{[r, i]}}
          }_\text{\normalfont Online updates $n = N+1, \ldots, N+M$.}
        \Biggl) \\
        \Matrix{W}_{M,1,1}^{[r,i], \operatorname{batch}} &= \operatorname{diag}\left(%
          \dfrac{(1 - \gamma)^{M-1}}{(\widehat{\sigma}^2_{1|M})^{[r,i]}}, \ldots,
          \dfrac{(1- \gamma)^{M-N-1}}{(\widehat{\sigma}^2_{N|M})^{[r, i]}}, \ldots,
          \dfrac{(1 - \gamma)}{(\widehat{\sigma}^2_{M-1|M})^{[r, i]}} ,
          \dfrac{1}{(\widehat{\sigma}^2_{M|M})^{[r,i]}}
        \right)
      \end{align*}%
    \end{small}%
    where $\gamma$ is the exponential discounting factor, which is applied to the repeated batch estimation and the online estimation, and the subscript $n|i$ denotes the last observation $i$ available. The difference can be summarized in three terms
    \begin{equation}\label{eq:weight_difference}
      \begin{split}
        \Matrix{W}_{M,1,1}^{[r, i], \operatorname{online}} - \Matrix{W}_{M,1,1}^{[r, i], \operatorname{batch}} =
        \underbrace{\sum_{n=1}^{N} (1-\gamma)^{M-n} \left(\dfrac{1}{\widehat{\sigma}^{2,\operatorname{batch}}_{n|N}} -\dfrac{1}{(\widehat{\sigma}^{2,\operatorname{online}}_{n|N})^{[r, i]}}\right)}_\text{\normalfont Initial batch.} + \\
        \underbrace{\sum_{n=N+1}^{M-1} (1-\gamma)^{M-n} \left(\dfrac{1}{\widehat{\sigma}^{2,\operatorname{batch}}_{n|n}} -\dfrac{1}{(\widehat{\sigma}^{2,\operatorname{online}}_{n|M})^{[r,i]}}\right)}_\text{\normalfont Online fit $N+1, \ldots, M-1$} +
        \underbrace{\dfrac{1}{(\widehat{\sigma}^{2,\operatorname{online}}_{M|M})^{[r,i]}} - \dfrac{1}{(\widehat{\sigma}^{2,\operatorname{batch}}_{M|M})^{[r,i]}}}_\text{\normalfont Current step $M-1 \rightarrow M$.}
      \end{split}
    \end{equation}
    where it is important to note that even in the last term $(\widehat{\sigma}^{2,\operatorname{online}}_{M|M})^{[r,i]}$ and $(\widehat{\sigma}^{2,\operatorname{batch}}_{M|M})^{[r,i]}$ from the online and batch estimation are \underline{not} identical, since the online estimation is path-dependent (see the recursive insertion of $w_{N+1, k,k}$ in Equation \ref{eq:online_weight_matrx}).
  \end{remark}
}%

From Remark \ref{remark:weights_online} we see that the difference in the online and batch setting depends on the length of the initial batch \review{$N$, how long the algorithm is running online $M-N$, and the strength of the exponential discounting. Looking at Equation \ref{eq:weight_difference} and keeping in mind that for large sample sizes, $(1-\gamma)^{M}, \ldots, (1-\gamma)^{M-n}$ will eventually approach values close to zero, it is tempting to conclude that for sufficiently high discounting and after a sufficiently large number of online updates $M$ the first term of Equation \ref{eq:weight_difference} ceases to influence the estimation.\footnote{Precisely, when $(1-\gamma)^{(N-M)} \rightarrow 0$ the observations that are in the initial sample receive a weight of zero.} However, due to the path-dependency of the online algorithm, where each update step's weight depends on the previous update step, this argument does not hold. For different combinations of the distribution~$\mathcal{F}$, link functions $g_k(\middledot)$, and distribution parameter $k$ the exact form of the above weights might be significantly more complex. This result shows a main difference to the regression to the mean} setting, where the batch and online results coincide \review{for OCD and RLS algorithms}.

\FloatBarrier
\section{Simulation Study}\label{sec:simulation}

\begin{figure}[htbp]
  \Description[Difference between the estimated and true coefficients for the distributional regression model in the first simulation study.]{
    Difference between the estimated and true coefficients for the distributional regression model in the first simulation study. We compare estimation with OLS and OCD and see that the error for OCD on the location parameter is somewhat similar, but is lower for OCD on small sample sizes for the scale parameter.
  }
  \centering
  \includegraphics[width=\linewidth]{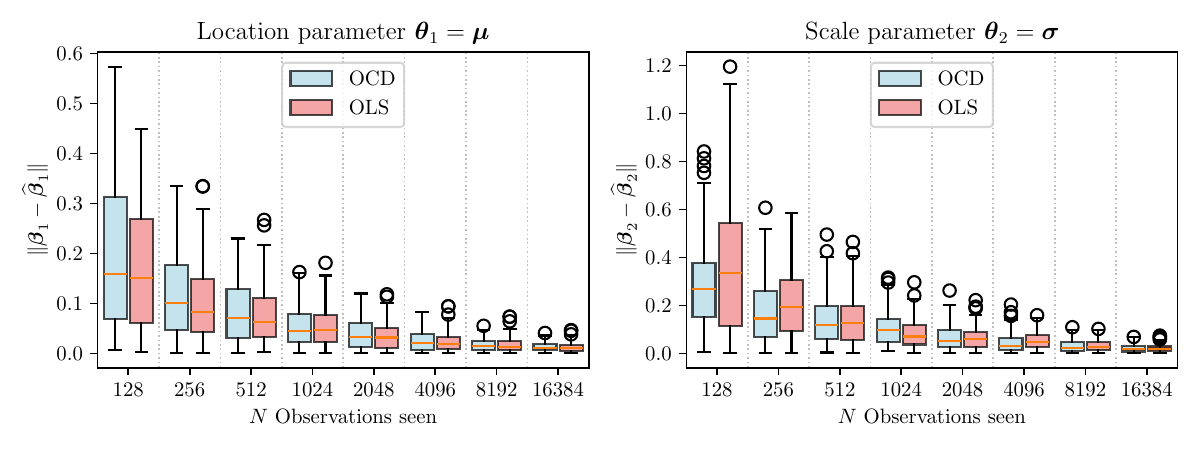}
  \caption{The difference $\| \Vector{\beta}_k - \widehat{\Vector{\beta}}_k \|_1$ relative to the selected initial batch size $n_0$. The main panel shows box-plots of all $M=10$ simulations. The top plot gives results for $\widehat{\Vector{\beta}}_0$, the bottom plot gives results for $\widehat{\Vector{\beta}}_1$.}
  \label{fig:simulation_study_results_init_batch}
\end{figure}

\begin{figure}[htbp]
  \Description[Difference in computation times between OLS and OCD based estimation.]{
    Difference in computation times between OLS and OCD based estimation. We see that the estimation time increases faster for OLS than for OCD due to the quadratic complexity in the IRLS update. For a very small number of features, OLS is faster.
  }
  \centering
  \includegraphics[width=\linewidth]{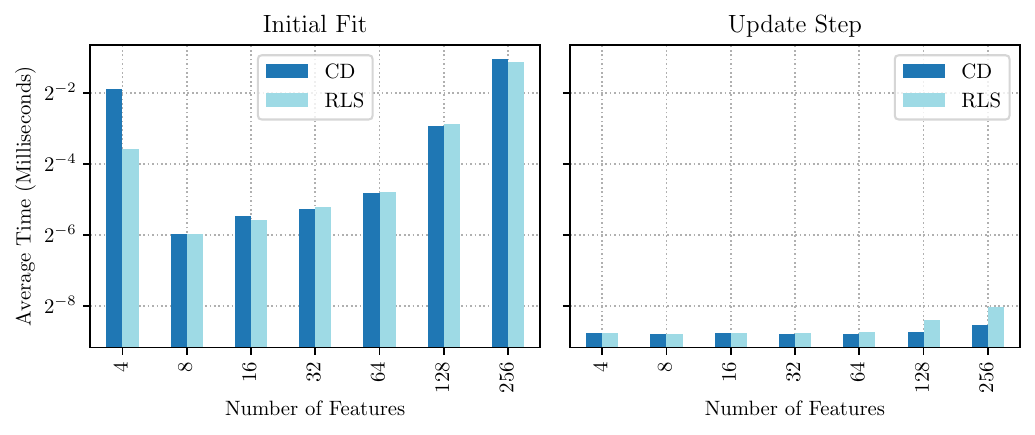}
  \caption{Average estimation time in Milliseconds, conditional on the number of features, using RLS and OCD.}
  \label{fig:simulation_study_results_timing}
\end{figure}

\begin{figure}[htb]
  \Description[
    Various scoring rules, averaged over all runs for different values of the forget in the simulation study with randomly changing structural breaks.
  ]{
    Various scoring rules, averaged over all runs for different values of the forget in the simulation study with randomly changing structural breaks. We see the lowest scores around a forget of 2^8, which corresponds to an effective sample size of 256 observations. The computation time increases with increasing forget.
  }
  \centering
  \includegraphics[width=\linewidth]{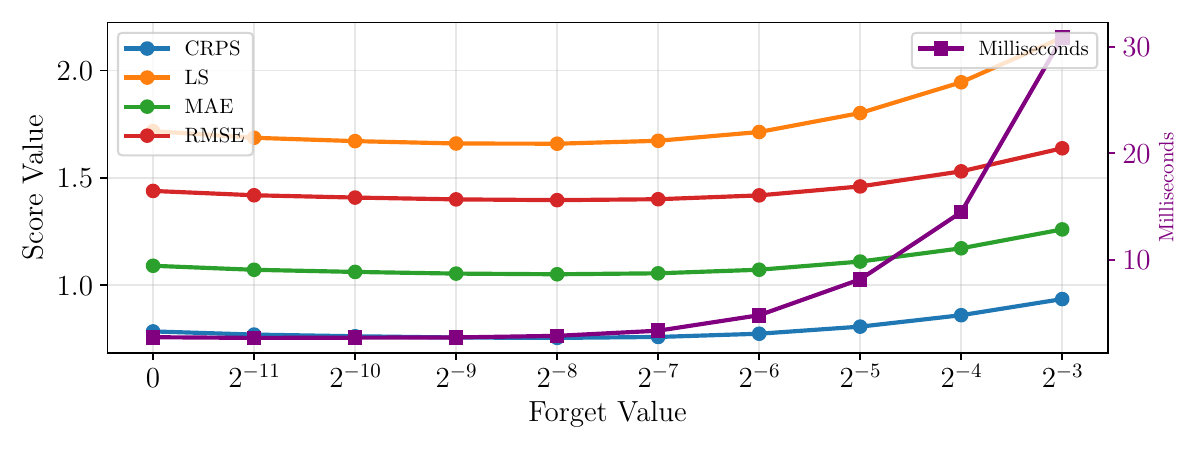}
  \caption{Average performance scores (left axis) and estimation time (right axis) across all 10 runs, for different forget values.}
  \label{fig:results_chaning_betas}
\end{figure}

\review{
  This section presents three distinct simulation studies on different aspects of the online distributional regression model. We analyze timings, the impact of the length of the initial training set, and the forgetting factor. Here, the data-generating process (DGP) is similar for all three studies. We consider a setting where $\Vector{y}$ is normally distributed. The location and scale parameters depend on the $N \times J$ covariate matrix $\Matrix{X} =\Matrix{X}_{1} = \Matrix{X}_{2}$. We simulate the entries of $\Matrix{X}$ by a $J$-dimensional normal distribution ${\Matrix{X}_{i} \sim \mathcal{N}( 0_J, I_J/2)}$ and define
  \begin{equation}
    y_{n}
    \sim \mathcal{N}\left( \Matrix{X}_{n}\Vector{\beta}_1  , \;g^{-1}(\Vector{X}_{n}\Vector{\beta}_2)\right)
  \end{equation}
  with $g(x, \alpha) = \log\left(1 + e^{x}\right) + \alpha$, a shifted softplus link function. Accordingly, we use a shifted softplus link function for estimating the scale parameter of the normal distribution. We set $\alpha=0.5$ to ensure sufficiently large variances. This greatly improves the stability of the estimations. The coefficient vectors $\Vector{\beta}_k$ of length $J$ are drawn from the uniform distribution:
  \begin{align}
    \Vector{\beta}_k \sim \mathcal{U}(-1, 1).
  \end{align}
  Unless mentioned otherwise, we set $J=10$, use $N=10000$ observations, use $2000$ observations for the initial fit, and execute 10 simulation runs.
  \begin{enumerate}
    \item The first simulation examines the convergence of the parameters to their true values in a batch setting. For that, we compute the $\ell_1$-norm $\| \middledot \|_1$ of the difference between $\Vector{\beta}_k$ and $\Vector{\widehat{\beta}}_k$ for different initial sample sizes. Thereby, we consider sample sizes of an exponential grid running from $2^7$ to $2^{13}$. We consider model estimations using RLS and OCD. Figure~\ref{fig:simulation_study_results_init_batch} illustrates the results. The results are as expected. Both parameters converge to their true values as $N$ grows. There are no notable differences between RLS and OCD in this setting.
    \item The second study compares the estimation times between RLS and OCD. Thereby, we distinguish between the initial fit and the subsequent update steps. Furthermore, we vary the number of features $J$. We consider an exponential grid ranging from $J=2^2$ to $J=2^8$. RLS has a per-iteration time complexity of $\mathcal{O}(JN^2)$. Online coordinate descent has a complexity of $\mathcal{O}(JN)$, which means we expect RLS computation times to be lower for smaller feature sets. On the other hand, we expect OCD to be faster if the feature set is sufficiently large. Our measurements are depicted in Figure~\ref{fig:simulation_study_results_timing}. Surprisingly, RLS and OCD need more time for the initial fit of the smallest model ($J=4$) compared to the larger ones. We barely observe differences in the timings for smaller data sets. For more than $J=2^6=64$ features, online updates using OCD are faster compared to RLS.
    \item Additionally, we conduct a third study. This demonstrates how forgetting affects the speed of parameter adjustment. For this, we slightly adjust the data-generating process. We define change points at which $\theta_{j}$ changes. These change points are drawn from a geometric distribution with a mean of $1000$. Then, we estimate distributional models using LASSO, and we consider a grid of forget values $\gamma = \{0, 2^{-11}, 2^{-10}, \ldots, 2^{-3}\}$. Figure~\ref{fig:results_chaning_betas} presents the results. In this setting, a positive forget is beneficial. More precisely, the forget value of $2^{-8}$ produces the best results across all considered scoring rules. This forget corresponds to an effective sample size of $2^8 = 256$. Furthermore, we observe an increase in estimation time as the forgetting increases. This is expected. As the forgetting increases, the effective sample size decreases, and the algorithm becomes less stable. This means more iterations are needed for the algorithm to converge. Figure~\ref{fig:simulation_study_params_forget} in Appendix~\ref{app:simulation_study_additional_results} presents the parameter estimates for selected forget values for all 20 parameters over time.
\end{enumerate}}%

\FloatBarrier
\section{Forecasting Study for Electricity Prices}\label{sec:forecasting}

\subsection{Data, Study Design and Models}

The following section presents an application in electricity price forecasting (EPF). We employ the same setting as \cite{marcjasz2023distributional} and \cite{brusaferri2024line} in forecasting day-ahead electricity prices for the German short-term electricity market. We directly compare the forecasting performance of online estimation vs repeated batch estimation directly for models using the Gaussian, Student-$t$, and Johnson $S_U$ (JSU) distribution and for OLS/RLS and LASSO estimation (see Table \ref{tab:results_epf_batch_vs_online}). Furthermore, we compare our approach to established benchmark models from the literature and conduct several ablation studies (see Appendix \ref{app:forecasting_study_additional_results}, Tables \ref{tab:results_epf_benchmark} and \ref{tab:results_epf_ablation}). The following paragraphs describe the German day-ahead electricity market, data set, and models in detail.
\begin{figure}[htb]
  \Description[
    Time Series plots of electricity prices, residual load and fundamental fuel prices.
  ]{
    Time Series plots of electricity prices, residual load and fundamental fuel prices. Electricity prices are characterized by their high volatility, while the residual load is somewhat more seasonal effects and slightly skewed. The fuel price time series behave more stochastic and show less seasonal pattern.
  }
  \centering
  \includegraphics[width=\linewidth]{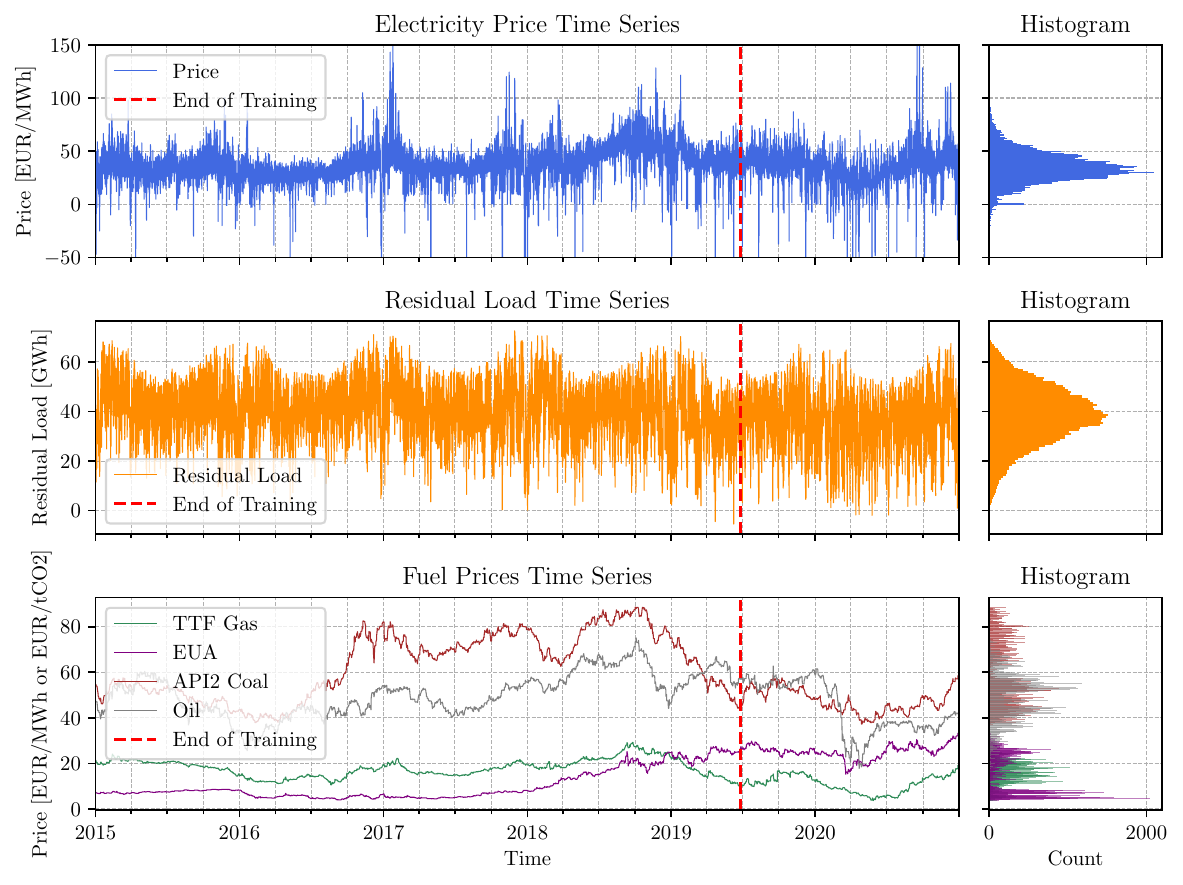}
  \caption{Overview of the time series data for the electricity price forecasting application. The top shows the EPEX day-ahead spot price, the middle panel shows the residual load, i.e., the system load minus renewable energy production, and the bottom panel shows the fundamental fuel prices.}
  \label{fig:epf_time_series_overview}
\end{figure}%

%The German electricity market consists of four major markets: The long-term futures market, the day-ahead spot market, the intraday market, and the balancing market. Since our forecasting study is concerned with the day-ahead or spot market only, our description focuses on this market only. The spot market is the major reference price for the futures market, and its trading volume is a multiple of the intraday and balancing market.
Let $h \in \{0, 1, \ldots, H\}$ and $H=23$ denote the 24 delivery hours and $d$ denote the delivery day. The day-ahead electricity market serves as the main trading venue for electricity and as reference price for the long-term futures market. It is organized as a daily auction at $d-1$, 12:00 hours for all 24 delivery hours of the following day, \review{where electricity generators, consumers, and traders can place price-volume bids to buy and sell electricity in each delivery hour. The market is cleared in a pay-as-cleared auction, taking into account the available transfer capacity between the European countries} \citep[further information can be found in, e.g.][]{viehmann2017state, marcjasz2023distributional, ziel2018day, lago2018forecasting, epftoolbox}. \review{Figure \ref{fig:epf_time_series_overview} gives an overview of the electricity price time series, the residual load (system load minus renewable energy production), and fuel prices for conventional generation assets. The residual load is the amount of electricity production that has to be covered by conventional generation sources (e.g., gas, coal, and oil). From an economic point of view, the renewable generation is often assumed to operate at zero or even negative marginal costs, while conventional power plants need to cover fuel and emission certificate costs, and therefore, the residual load is an important factor in the price formation.} The data is retrieved from \cite{marcjasz2023distributional}\footnote{See \url{https://github.com/gmarcjasz/distributionalnn}. \cite{marcjasz2023distributional} provide data from the European Network of Transmission System Operators (ENTSO-E) and Refinitiv.} The initial training set comprises 2015-01-15 to 2019-06-26 (1631 days), and the test set comprises 2019-06-26 to 2020-12-31 (534 days). This is in line with the test and training split of \cite{brusaferri2024line} and \cite{marcjasz2023distributional}.

We employ a so-called expert-type model as it is common in electricity price forecasting \citep{lago2018forecasting, ziel2018day, marcjasz2023distributional}. The model consists \review{of 43 linear terms and of 5 non-linear additive effects modeled by B-Spline bases. The covariates capture} the autoregressive price effects, seasonal effects, the fundamental effects of renewable generation, and the influence of fuel prices. We employ the \review{Gaussian}, Student-$t$, and Johnson's $S_U$ (JSU) distributions. \review{The Gaussian distribution serves as a benchmark, and the latter two} have been used for electricity price forecasting and other financial applications. In the distributional framework, we model each distribution parameter $k = 1, \ldots, p$ for each delivery hour $h = \{0, 1, \ldots, H\}$ individually:
\begin{align}\label{eq:model}
  g_k(\theta_{k,d,h})
  &= \beta_{k, 0, h} + \sum_{s \in \{0, 1, \ldots, H\}} \beta_{k, s, h} P_{d-1, s} + \sum_{l=2}^{L=7}  \beta_{k, 24+l, h} P_{d-l, h}\\ \nonumber
  &+ \beta_{k,31, h}\operatorname{min}(\Vector{P}_{d-1})
  + \beta_{k,32,h}\max(\Vector{P}_{d-1})
  + \beta_{k,33,h}\mathcal{Q}_{10}(\Vector{P}_{d-1})
  + \beta_{k,34,h}\mathcal{Q}_{90}(\Vector{P}_{d-1}) \\ \nonumber
  &+ \beta_{k,35,h}\widehat{\operatorname{ResLoad}}_{d}
  + \sum_{\operatorname{WD} \in \text{\{Mon,Tue,Thu,Fri,Sat,Sun,Hol\}}} \beta_{k, 35+w, h} \operatorname{WD}_{d} \\ \nonumber
  &+ b_{k,1,h}(\widehat{\operatorname{ResLoad}}_{d, h})
  + b_{k,2}(\operatorname{EUA}_{d-2})
  + b_{k,3}(\operatorname{Gas}_{d-2})
  + b_{k,4}(\operatorname{Coal}_{d-2})
  + b_{k,5}(\operatorname{Oil}_{d-2})
\end{align}%
where $\beta_{k, 0, h}$ is the intercept or bias, $\beta_{k, 1, h}$ to $\beta_{k, 30, h}$ capture autoregressive effects for the previous day $d-1$, all delivery hours, and for the last 7 days of the same delivery hour~$h$. $\beta_{k, 31, h}$ to $\beta_{k, 34, h}$ capture the influence of very large and low prices for the previous day. $\beta_{k, 35, h}$ models the influence of the daily residual load. $\beta_{k, 36, h}$ to $\beta_{k, 42, h}$ capture the effects of weekly seasonality and holidays.\footnote{German public holidays are taken from the \texttt{Python} \texttt{holidays} package. On a public holiday, we set all other weekday dummies to zero. The holiday days include New Year's, Good Friday, Easter Monday, Labor Day, Ascension Day, Whit Monday, German Unity Day, and Christmas.} Lastly, the terms $b_{k,1,h}(\middledot)$ to $b_{k,5}$ capture the non-linear effects of hourly residual load, European Emission Allowances (EUAs), natural gas, coal and oil prices by B-Splines of degree 2 and using 4 knots, placed according to the quantiles of the covariates in the training set \citep[see][for more information on B-Splines]{eilers1996flexible}.

We benchmark forecasts using established, strictly proper probabilistic scoring rules \citep{gneiting2008probabilistic, gneiting2011making, gneiting2014probabilistic, nowotarski2018recent}. For the mean prediction, we employ the Root Mean Squared Error (RMSE), and \review{for the median prediction, we use the} Mean Absolute Error (MAE). We evaluate the Coverage (COV) and the Interval Score \citep[IS, also known as Winkler Score, see ][]{bracher2021evaluating} for the \review{50\%{} and 80\%{}} prediction intervals (PI). For the full predictive distribution, we evaluate the Log Score (LS) and the continuous ranked probability score (CRPS) using the approximation via the Pinball Score (PS) on a dense grid of quantiles $\mathcal{Q} = \{0.01, 0.02, 0.03, \ldots, 0.99\}$. The implementation of the scoring rules is provided by the \texttt{scoringrules} package \citep{zanetta_scoringrules_2024}. We evaluate the statistical significance of the difference in predictive accuracy using the Diebold-Mariano test \citep{diebold2002comparing, diebold2015comparing}. \review{For the coverage, we evaluate the Kupiec-test (see \cite{kupiec1995techniques}) per hour $h$. Detailed definitions of the scoring rules and the statistical tests can be found in Appendix~\ref{app:scoring_rules}}. Lastly, we evaluate the computation time for all forecasting studies.

\subsection{Results and Discussion}\label{sec:results}

\begin{figure}[!ht]
  \Description[
    Illustrative forecasts and prediction intervals for the week around Christmas 2020. The prediction intervals are time-varying and adjust to the price level and uncertainty.
  ]{
    Illustrative forecasts and prediction intervals for the week around Christmas 2020. The prediction intervals are time-varying and adjust to the price level and uncertainty. On 24th, 26th prediction intervals are wider. On 28th, prediction intervals are very wide and prices turn out negative.
  }
  \centering
  \includegraphics[width=0.75\textwidth]{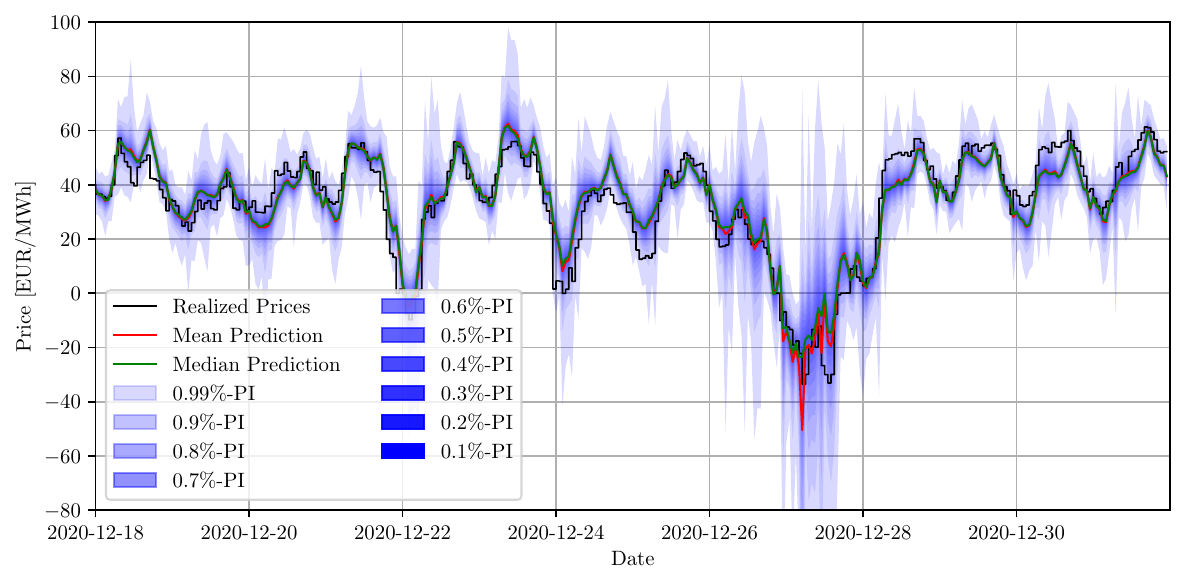}
  \caption{Illustrative Forecast. Prediction Intervals derived from the online probabilistic model. Here, we assume the power prices to follow Johnson's $S_U$ distribution and model all moments conditionally on Equation \ref{eq:model}. The prediction intervals correspond to the $\{0.005, 0.05, 0.1, \ldots, 0.95, 0.995\}$ quantiles of the predictive distribution. Note that the period with extremely low prices corresponds to Christmas.}
  \label{fig:illustrative_forecast}
\end{figure}

\begin{figure}
  \Description[
    Non-linear effects of the residual load on the location and scale parameter for the $t$-distribution.
  ]{
    Non-linear effects of the residual load on the location and scale parameter for the $t$-distribution. The scale parameter (i.e., the price volatility) increases if the residual load is very high or very low, leading to a $u$-shaped non-linear effect, while the strength of the effect changes through the out-of-sample set.
  }
  \centering
  \includegraphics[width=\linewidth]{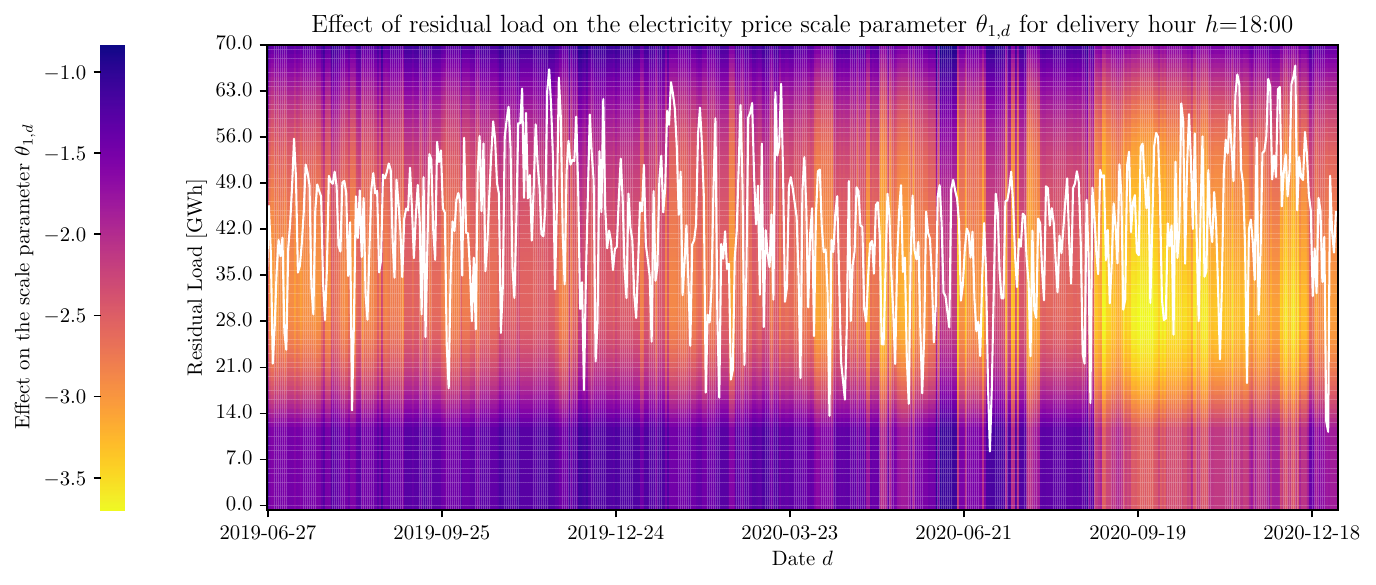}
  \caption{Non-linear effects of the residual load on the scale (volatility) parameter for the $t$-distribution. The background coloring gives the non-linear effect for the value on the $x$-axis, and the $y$-axis denotes the update steps on the test set. The white line corresponds to the actual value. For the location, the effect on the location parameter is steeper towards the lower and upper end of the variable's domain.  The scale parameter (i.e., the price volatility) increases if the residual load isS very high or very low, leading to a $u$-shaped non-linear effect, while the strength of the effect changes through the out-of-sample set.}
  \label{fig:results_epf_nonlinear_effect_scale}
\end{figure}

The following section presents and discusses the results of the forecasting study. Figure~\ref{fig:illustrative_forecast} gives an exemplary forecast. The time-varying width of the prediction intervals illustrates the importance of modelling the conditional volatility. Figure~\ref{fig:results_epf_nonlinear_effect_scale} shows the non-linear effects of residual load on the scale parameter (A similar Figure for the location $\mu$ can be found in Appendix \ref{app:forecasting_study_additional_results}). The effect is $u$-shaped, indicating that the price volatility increases for very high and very low residual load levels. This is in line with economic theory, as extreme residual load levels indicate scarcity or oversupply situations, which lead to more volatile prices.

\begin{table}[!ht]
  \resizebox{\textwidth}{!}{%
  \input{tables/electricity_price/01_epf_batch_vs_online}}
  \centering\includegraphics[width=0.75\textwidth]{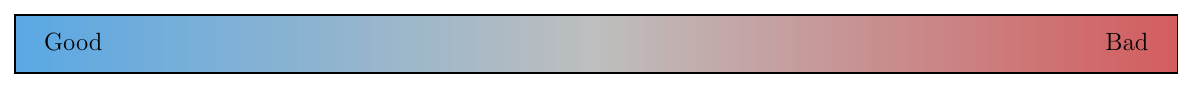}
  \caption{%
    Results for the main experiment: Repeated batch estimation vs online estimation. Distribution refers to the assumed parametric form of the response distribution. Setting refers to whether models are estimated incrementally or retrained on the increasing window training set. The best value in each column is marked \textbf{bold}. We conduct the pairwise DM-test (Online, vs. Batch) for the RMSE, MAE, IS, CRPS, and LS and mark results that cross the $p<0.05$ significance threshold \textit{italic}. For the coverage of the 50\% and 80\% prediction intervals, the background color corresponds to the number of hours where the \cite{kupiec1995techniques}-test is passed at the 5\%-level. Note that the timing corresponds to a full forecasting study, i.e., the estimation of $534 \times 24 = 12.816$ models.
  }
  \label{tab:results_epf_batch_vs_online}
\end{table}

Table \ref{tab:results_epf_batch_vs_online} presents our main results, the direct comparison of repeated batch and online estimation for the distributional regression model. We make the following observations about the predictive performance in our forecasting study:
\begin{itemize}
  \item Comparing the distributional regression models in Table \ref{tab:results_epf_batch_vs_online} like-for-like, the predictive performance of the repeated batch estimation and online estimation is very close and, for various models and scoring rules, not statistically significantly different. At the same time, the computational effort of the online estimation is 2-3 orders of magnitude lower than that of the repeated batch estimation. This result underscores the robustness and efficiency of the proposed algorithm.
  \item We see a general improvement in forecasting performance when using regularized estimation approaches compared to ordinary least squares. Especially for the JSU distribution, the classic OLS estimation struggles with the heavy-tailed and skewed electricity price data in the batch and online estimation. The issue is more pronounced in the online case, but can be successfully alleviated by employing regularized estimation methods such as the LASSO or elastic net (see also Table~\ref{tab:results_epf_ablation}, Panel~A). The LASSO-estimated JSU model is the strongest in our study.
\end{itemize}
Furthermore, we make the following observations when comparing to established benchmarks and conducting ablation studies:
\begin{itemize}
  \item In the comparison to established time-series benchmarks and state-of-the-art, neural-network-based EPF approaches, we see the competitive performance of the online distributional regression models. We see that the AutoARIMA generally struggles to adapt to the highly volatile data, and our investigation showed that stepwise model selection seems not robust enough.
  \item Compared with state-of-the art results of \cite{brusaferri2024line}, our approach delivers slightly better performance in terms of the CRPS, but slightly worse coverage. The DDNN approach of \cite{marcjasz2023distributional} yields better results than the distributional regression approaches and the conformalized DDNNs of \cite{brusaferri2024line}. However, the DDNN approaches necessitate specialized computing infrastructure, and even without hyperparameter tuning, need more than 100 times the computing time of the online distributional regression on a standard desktop PC.
  \item In the comparison between the linear online distributional regression and the model with B-Spline bases for the fundamental covariates, we see that the linear model yields worse performance for the Normal and Student-$t$ distribution, but slightly improved performance for the JSU distribution. We attribute this to an increased risk of overfitting and the issue of identifiability in distributional regression, especially for the skewness and tail behavior in the non-linear model, as already discussed in Section \ref{sec:method}.
  \item Table \ref{tab:results_epf_ablation}, Panel B gives results for the estimation using different forget factors, using an effective sample size of 1 to 4 years in the exponential discounting. The intuitive rationale behind exponential forgetting, that newer observations contribute more to forecasts, is confirmed in the results. For all three distributional assumptions, the forecasting can be improved using a small forget. However, we also note that increasing $\gamma$ yields higher computational costs to a small extent.
  \item Further discussion of the ablation studies can be found in Appendix~\ref{app:forecasting_study_additional_results}. In summary, we note that the local model selection approach is preferable due to its computational efficiency, and that batch updates can be used to speed up the estimation further, at the cost of some predictive accuracy.
\end{itemize}
Wrapping up, our study demonstrates that online distributional regression models achieve competitive predictive performance compared to both traditional time-series benchmarks and advanced neural network-based approaches. Moreover, they offer significantly lower computational costs - often by two to three orders of magnitude - making them highly efficient for real-time forecasting.

\section{Implementation in the Python Package \texttt{ondil}}\label{sec:python}

We provide an open-source implementation of our algorithm in the \texttt{Python} package \texttt{ondil}. To the best of the authors' knowledge, this package is the first native \texttt{Python} implementation of IRLS-based distributional regression and can, therefore, provide a basis for future extensions.\footnote{The package \texttt{pyNM} implements GAMLSS for \texttt{Python} as binding to the \texttt{R} library \url{https://github.com/ppsp-team/PyNM}.} We provide the following features:
\begin{itemize}
  \item Our package is written in an online-first fashion and provides an API compatible with \texttt{scikit-learn} \cite{pedregosa2011scikit}, the major Python machine learning package. Beyond the \texttt{estimator.fit(X,~y)} and  \texttt{estimator.predict(X,~y)} methods, our estimator classes provide the \texttt{estimator.update(X,~y)} method to allow for incremental updating. % We employ automatic mean-variance scaling for all covariates with an incremental calculation of the exponentially discounted mean and variance using Welford's algorithm \citep{welford1962note}.
  \item The implementation relies on few dependencies \citep[only \texttt{numpy}, \texttt{numba}, and \texttt{scipy}, see][]{2020NumPy-Array, 2020SciPy-NMeth, lam2015numba}. We employ just-in-time compilation using \texttt{numba} to achieve a high-performance implementation and employ various computational tricks such as active-set coordinate descent, different warm-starting options (using the previous fit or a mixture of previous fit and previous $\beta_\lambda$ on the same coefficient path), and allow for random selection during coordinate descent to speed up the convergence of the coordinate descent \citep[see e.g.][]{shi2016primer, wright2015coordinate}.
  \item We implement standard distributions and facilitate extensions based on \texttt{scipy}-distributions.
\end{itemize}
The code for our implementation is open source \review{on \texttt{GitHub}} and the package is available at the Python Package Index.\footnote{See: \url{https://pypi.org/project/ondil} and \url{https://github.com/simon-hirsch/ondil}.} \review{All studies in this paper have been run using \texttt{ondil} version 0.3.2.}

\section{Discussion and Conclusion}\label{sec:conclusion}

\review{This paper presents an efficient, scalable approach to distributional regression by the implementation of an online estimation algorithm} for the well-known, parametric GAMLSS \citep[for batch settings see, e.g.][]{rigby2005generalized, klein2024distributional}. \review{We discuss regularized estimation using ridge, LASSO, and the elastic net, and the associated issue of online model selection. Furthermore, we formally analyze the approximation quality of online estimation and batch estimation.} Lastly, we provide an open-source Python implementation in the \texttt{ondil} package. We validate our approach in a forecasting study for the German day-ahead electricity prices, a highly volatile data set. The online distributional regression delivers competitive forecasting accuracy in the CRPS compared to repeated batch estimation and other distributional forecasting approaches while reducing the computation time by some 2-3 orders of magnitude. Furthermore, we provide simulation results on the impact of the initial training size and the exponential discounting on the approximation quality of the online models compared to repeated batch fitting. Our simulation results are in line with the aforementioned theoretical results.

Our research opens up multiple avenues for future research. First, while our empirical results are promising, further theoretical results on the error bounds of the online algorithm would further increase the trust in the proposed method.  Secondly, we present a regularized method for the online estimation, but the (online) model selection in distributional regression is a relatively untapped field, and advances here will directly benefit practical applications. \review{While our approach is capable of the estimation of non-linear models using linear basis functions, the implementation of regularized splines and other additive effects is a practically relevant addition. Lastly, online learning for distributional models is naturally confronted with issues of identifiability, and our approach is no exception. We note that regularization plays a critical role in controlling the issue in a practical study; a rigorous formal analysis seems like a promising avenue for further research, benefiting all research on online distributional models.}

\bibliographystyle{ACM-Reference-Format}
\bibliography{references}

\begin{acks}
  Simon Hirsch is employed as an industrial PhD student by Statkraft Trading GmbH and gratefully acknowledges the support and funding received. Simon Hirsch is grateful to Daniel Gruhlke for many helpful discussions. This work contains the author's opinions and does not necessarily reflect Statkraft position. The authors declare no conflict of interest.
\end{acks}

\newpage

\appendix
\section{Abbreviations}
\begin{table}[htb]
  \centering
  \begin{tabular}{r|l}
    AIC                 &   Akaike's Information Criterion \\
    BIC                 &   Bayesian Information Criterion \\
    DDNN               &   Deep Distributional Neural Network \\
    DGP                &   Data generating process \\
    EPF                &   Electricity Price Forecasting \\
    GAMLSS             &   Generalized Additive Model for Location, Scale and Shape \\
    GIC                &   Generalized Information Criterion \\
    HQC                &   Hannan-Quinn Criterion \\
    OCD                &   Online Coordinate Descent \\
    OLS                &   Ordinary Least Squares \\
    CCD                &   Cyclic Coordinate Descent \\
    LASSO              &   Least Absolute Shrinkage and Selection Operator \\
    PB                 &   Pinball Scores \\
    PI                 &   Prediction Interval \\
    CRPS               &   Continuous Ranked Probability Scores \\
    RMSE               &   Root Mean Square Absolute Error \\
    MAE                &   Mean Absolute Error \\
    IS                 &   Interval Score (also known as Winkler Score) \\
    IRLS               &   Iteratively Reweighted Least Squares \\
    COV                 &   Coverage \\
    LS                 &   Log-Score \\
    ARIMA              &   AutoRegressive Integrated Moving Average \\
    EUA                &   European Emission Allowance \\
    RLS                &   Recursive Least Squares \\
    TTF                &   Title Transfer Facility (trading hub for natural gas contracts) \\
  \end{tabular}
  \caption{Abbreviations}
  \label{tab:abbrevations}
\end{table}

\FloatBarrier

\section{Working Vectors and Weights in Remark \ref{remark:gaussian_distributional_regression}}
In the following, we prove the result given in Remark \ref{remark:gaussian_distributional_regression}. For the parameters $\theta_1 = \mu$ and $\theta_2 = \sigma^2$, the derivatives of the log-likelihood of the normal distribution are given by the following equations. Note that we evaluate the deriatives element-wise, hence $\partial \Vector{\ell} / \partial \Vector{\theta} = (\partial \ell_1 / \partial \theta_1, ..., \partial \ell_N / \partial \theta_N)$. We have:
\begin{align*}
  \dfrac{\partial \Vector{\ell}}{\partial \Vector{\mu}}                &= (\Vector{\sigma}^2)^{-1} \hdmul (\Vector{y} - \Vector{\mu})    &
  \dfrac{\partial \Vector{\ell}}{\partial (\Vector{\sigma}^2)}        &= (\Vector{\sigma}^4)^{-1} \hdmul \dfrac{1}{2}  ((\Vector{y} - \Vector{\mu})^2 - \Vector{\sigma}^2)\\
  \mathbb{E}\left[\dfrac{\partial^2 \ell}{\partial \Vector{\mu}^2}\right]            &= - \left(\Vector{\sigma}^2\right)^{-1}            &
  \mathbb{E}\left[\dfrac{\partial^2 \ell}{\partial (\Vector{\sigma}^2)^2}\right]     &= -\dfrac{1}{2}\left({\Vector{\sigma}^4}\right)^{-1}
\end{align*}
we therefore have the weights $\Matrix{W}_{1,1} = \diag((\Vector{\sigma}^2)^{-1})$ and $\Matrix{W}_{1,1} = \diag(2(\Vector{\sigma}^4)^{-1})$ since $\partial \Vector{\eta}_k/\partial\Vector{\theta}_k = 1$ for the identity link function. The working vector is given by Equation \ref{eq:gamlss_working_vector} and can be written as:
\begin{align*}
  \Vector{z}_1 &= \Vector{\mu} +  \diag\left(\left(\Vector{\sigma}^2\right)^{-1}\right)^{-1} \left({\Vector{\sigma}^2}\right)^{-1} \hdmul (\Vector{y} - \Vector{\mu}) = \Vector{y} \\
  \Vector{z}_2 &= \Vector{\sigma}^2 + \diag\left(\left(2\Vector{\sigma}^4\right)^{-1}\right)^{-1} \left(\Vector\sigma^4\right)^{-1} \hdmul  \dfrac{1}{2} \left({(\Vector{y} - \Vector{\mu})^2 - \Vector{\sigma}^2}\right)= (\Vector{y} - \Vector{\mu})^2
\end{align*}
where we exploit that $\diag(\Vector{a})\Vector{b} = \Vector{a} \hdmul \Vector{b}$ and hence returns the working vectors given in Remark \ref{remark:gaussian_distributional_regression} $\blacksquare$

\section{Details on the Update of the weighted Gramian Matrix}\label{app:gram_update}
The following paragraphs outline the update of the weighted Gramian matrix {$\Matrix{G}_N=\transpose{\Matrix{X}_N}\Matrix{W}_N\Matrix{X}_N$} using some new observations $\Vector{x}_{N+1}$ and new weights $w_{N+1}$ given to the forecaster. We desire to recover $\Matrix{G}_{N+1} = \transpose{\Matrix{X}_{N+1}}\Matrix{W}_{N+1}\Matrix{X}_{N+1}$ where the subscript $N$ and $N+1$ denote that the matrices contain all information received until $N$ respectively $N+1$.

\begin{definition}[Sherman-Morrison Formula]
  The Sherman-Morrison Formula states that for an invertible matrix $\Matrix{A}$ and vectors $\Vector{u}$ and $\Vector{v}$ it holds
  $$(\Matrix{A} + \Vector{u}\transpose{\Vector{v}})^{-1} = \Matrix{A}^{-1} - \frac{\Matrix{A}^{-1} \Vector{u} \transpose{\Vector{v}} \Matrix{A}^{-1}}{1+\transpose{\Vector{v}}\Matrix{A}^{-1} \Vector{u}}.
  $$
\end{definition}
For convenience, we define the matrices and vectors $\Matrix{M}_{n} = \Matrix{W}^\frac{1}{2}_N\Matrix{X}_N$, $\Matrix{M}_{N+1} = \Matrix{W}^\frac{1}{2}_{N+1}\Matrix{X}_{N+1}$, and $\Vector{m}_{N+1} = \sqrt{w_{N+1}} \Vector{x}_{N+1}$. Since we have $$\transpose{\Matrix{M}}_{N+1}\Matrix{M}_{N+1} = \transpose{\Matrix{M}}_{N}\Matrix{M}_{N} + \transpose{\Vector{m}}_{N+1}\Vector{m}_{N+1}$$ we can rewrite the inverted weighted Gram Matrix $$\Matrix{G}_{N+1}^{-1} = (\transpose{\Matrix{X}_{N+1}}\Matrix{W}_{N+1}\Matrix{X}_{N+1})^{-1} = (\transpose{\Matrix{M}_{N+1}}\Matrix{M}_{N+1})^{-1}$$ for the next observation $N+1$ as
\begin{align*}
  (\transpose{\Matrix{M}}_{N+1}\Matrix{M}_{N+1})^{-1}
  &= (\transpose{\Matrix{M}}_{N}\Matrix{M}_{N})^{-1} - \frac{
    (\transpose{\Matrix{M}}_{N}\Matrix{M}_{N})^{-1}  \transpose{\Vector{m}}_{N+1}\Vector{m}_{N+1} (\transpose{\Matrix{M}}_{n}\Matrix{M}_{n})^{-1}
  }{
  1 + \Vector{m}_{N+1}(\transpose{\Matrix{M}_{N}}\Matrix{M}_{N})^{-1} \transpose{\Vector{m}_{N+1}}  } \\
  &= (\transpose{\Matrix{X}_{N}}\Matrix{W}_{N}\Matrix{X}_{N})^{-1} - \frac{
    w_{N+1} (\transpose{\Matrix{X}_{N}}\Matrix{W}_{N}\Matrix{X}_{N})^{-1}\transpose{\Vector{x}}_{N+1}\Vector{x}_{N+1} (\transpose{\Matrix{X}_{N}}\Matrix{W}_{N}\Matrix{X}_{N})^{-1}
  }{
  1 + w_{N+1} \Vector{x}_{N+1} (\transpose{\Matrix{X}_{N}}\Matrix{W}_{N}\Matrix{X}_{N})^{-1} \transpose{\Vector{x}_{N+1}}}
\end{align*}
which gives the update Equation \ref{eq:update_weighted_inverse_gramian}. A similar argument can be made for Equation \ref{eq:update_exponentially_discounted_inverse_gramian}.

\section{Forecast Evaluation and Scoring Rules}\label{app:scoring_rules}

\review{In the following, we briefly describe the scoring rules used. We use a rather generic notation here.} For the mean forecast~$\widehat{\mu}_{n}$, the predictive distribution~$\widehat{\mathcal{D}}_n$, \review{shorthand for~$\mathcal{D}(\what{\Vector{\theta}}_{n})$}, and the realized \review{value} $y_n$, the scoring rules are defined as follows:
\begin{align}
  \operatorname{RMSE} &= \sqrt{\frac{1}{N} \sum_n^N \left( \widehat{\mu}_{n}\ - y_n \right)^2} \\
  \operatorname{MAE} &= \frac{1}{N} \sum_n^N \left| \widehat{\mathcal{D}}_n^{-1}(0.5) - y_n \right|
\end{align}
where $\widehat{\mathcal{D}}_n^{-1}(p)$ is the quantile or percentage point function. For a $(1 - \alpha) \times 100\%{}$-PI defined by the lower and upper bounds $\widehat{L}_n = \widehat{\mathcal{D}}_n^{-1}(\alpha / 2)$ and $\widehat{U}_n = \widehat{\mathcal{D}}_n^{-1}(1 - \alpha / 2)$, the COV and IS are defined as
\begin{align}
  \operatorname{COV_\alpha} &= \frac{1}{N} \sum_n^N \boldsymbol{1}_{\widehat{L}_{n} \leq y_{n} \leq \widehat{U}_{n}} \\
  \operatorname{IS_\alpha} &= \frac{1}{N} \sum_n^N
  \begin{cases}
    (\widehat{U}_n - \widehat{L}_n) + \frac{\alpha}{2}(\widehat{L}_n - y_{n}) &\quad \text{if $\widehat{L}_{n} > y_{n}$,} \\
    (\widehat{U}_n - \widehat{L}_n) + \frac{\alpha}{2}(y_n - \widehat{U}_{n}) &\quad \text{if $\widehat{U}_{n} < y_{n}$,} \\
    (\widehat{U}_n - \widehat{L}_n) &\quad \text{else.}
  \end{cases}
\end{align}
\review{Furthermore, we evaluate the coverage by employing the Kupiec test. The test evaluates whether a prediction lies inside the $(1 - \alpha) \times 100\%{}$-PI, under the assumption that all violations are independent. Let $C_1$ and  $C_0$ be the count of ``hits" and ``misses",
  \begin{align*}
    C_1 = \sum_{n=1}^N \boldsymbol{1}_{\widehat{L}_{n} \leq y_{n} \leq \widehat{U}_{n}} \quad \text{and} \quad C_0 = N - C_1
  \end{align*} and $\pi = C_1 / N$ is the percentage of ``hits". The likelihood ratio statistic for the test is given as:
  \begin{equation*}
    \operatorname{LR}_\text{Kupiec} = 2 \log \left( \frac{\alpha^{C_0}(1-\alpha)^{C_1}}{(1-\pi)^{C_0}\pi^{C_1}}\right)
\end{equation*} and is distributed asymptotically $\chi^2_1$ \citep{kupiec1995techniques, nowotarski2018recent}.}

Let us note that for a skewed distribution and the central prediction intervals, the Interval Score might be misleading since there can be non-central prediction intervals with the same coverage but smaller width (and hence lower IS). The CRPS is approximated by the PS on a dense grid of quantiles
\begin{align}
  \operatorname{PS}_{\alpha, d, s} &=
  \begin{cases}
    \alpha \left( y_n - \widehat{\mathcal{D}}_{{n}}^{-1}(\alpha) \right)            & \quad \text{if $y_n \geq \widehat{\mathcal{D}}_n^{-1}(\alpha)$,} \\
    (1 - \alpha) \left( \widehat{\mathcal{D}}_{{n}}^{-1}(\alpha) - y_n \right)      & \quad \text{else.}
  \end{cases} \\
  \operatorname{CRPS} &= \dfrac{2}{| \mathcal{Q} |} \frac{1}{N} \sum_{\alpha \in \mathcal{Q}}\sum_n^N   \operatorname{PS}_{\alpha,n}
\end{align}
Note that within the forecasting community, the CRPS is sometimes reported using $1 / | \mathcal{Q} |$ as the first fraction instead of $2 / | \mathcal{Q} |$. This formulation corresponds to $0.5 \times \operatorname{CRPS}$ and has initially been used this way in the GEFcom 2014 as the Average Pinball Score (APS) \citep{hong2016probabilistic, marcjasz2023distributional, nowotarski2018recent}.
Furthermore, we report the Log-Score (LS) as the negative log-likelihood of the true value under the predictive distribution. The LS is defined as:
\begin{equation}
  \text{LS} = \dfrac{1}{N} \sum_N^N - \operatorname{log}\left(\widehat{\boldsymbol{d}}_n(y_n)\right)
\end{equation}
where $\widehat{\boldsymbol{d}}_n(y_n)$ is the probability density function of the predictive distribution $\widehat{\mathcal{D}}_n$ and observation $y_n$. Let us note that the RMSE, MAE, IS, LS, and the CRPS are strictly proper scoring rules. For all of them, lower scores correspond to a better forecast. The COV is a measure of calibration for probabilistic forecasts only.

We evaluate the statistical significance of the differences in predictive accuracy using the Diebold-Mariano test \citep{diebold2002comparing, diebold2015comparing}. For two models $A$ and $B$, and the 24-dimensional vectors of scores $\Matrix{S}_d^A = \transpose{(s_{d, 0}^A, s_{d, 1}^A, \ldots,  s_{d, H}^A)}$ and $\Matrix{S}_d^B = \transpose{(s_{d, 0}^B, s_{d, 1}^B, \ldots,  s_{d, H}^B)}$ the DM-test employs the loss differential $$\boldsymbol{\Delta}^{A,B}_d = \| \Matrix{S}_{d, h}^A  \|_1 -  \| \Matrix{S}_{d, h}^B \|_1,$$ where $\| \middledot \|_1$ represents the $L_1$-norm \citep[see e.g.][]{nowotarski2018recent, lago2018forecasting, epftoolbox}. \review{We have the $H_0$ that $$\mathbb{E}\left[ \boldsymbol{\Delta}^{A,B}_d \right] = 0,$$ i.e. difference between the score vectors is zero. Rejecting the $H_0$, therefore, implies that the forecasts of one model are significantly better than the forecasts of the other model.}

\section{Additional Results for the Forecasting Study}\label{app:forecasting_study_additional_results}

This section presents additional results for the forecasting study on the German day-ahead electricity prices. We present two additional experiment sets and some further discussion of the in-sample results of the non-linear effects. We compare the full distributional regression models (modeling all distribution parameters) from the online and repeated batch estimation to a set of established benchmark models, such as the persistence and AutoARIMA(X), as well as the results of \cite{marcjasz2023distributional} and \cite{brusaferri2024line}. Furthermore we run several ablation studies on the online-LASSO models to analyze the impact of different hyper and algorithmic parameters. This includes changing the forget~$\gamma$, the online batch size, the model selection (see Section~\ref{sec:online_model_selection}), and additionally using the elastic net and unregularized online coordinate descent estimation.

In detail, the benchmark models are as follows:
\begin{itemize}
  \item \textbf{Linear Distributional Regression:} We also use an online distributional regression model where the terms $b_{k,1,h}(\middledot)$ to $b_{k,5}$ in Equation \ref{eq:model} are replaced by linear terms to test the ability of the model to capture non-linear effects.
  \item \textbf{Persistence} and \textbf{AutoARIMAX:} We include the persistence forecast and the Auto-ARIMA(X) as well-established benchmark models. The AutoARIMA(X) is estimated using the \texttt{statsforecast}-library by \citet{garza2022statsforecast}. We use the AutoARIMA as a pure time series model. For the AutoARIMAX we use the same set of covariates as in Equation \ref{eq:model}.
  \item \textbf{Distributional Neural Networks:} We report the results of \cite{marcjasz2023distributional}, who employ an ensemble of 4 deep distributional neural networks (DDNN), and we report the results of \cite{brusaferri2024line}, who employ a similar DDNN, respectively, a deep quantile regression neural network combined with online conformalization following the method of \cite{angelopoulos2023conformal}.
\end{itemize}%

In the ablation studies, we run the forecasting studies with the following changes to the main online LASSO model for all three distributional assumptions (Normal, Student-$t$, and JSU):
\begin{itemize}
  \item \textbf{Panel A:} We compare the LASSO regularization to the elastic net and unregularized online coordinate descent estimation.
  \item \textbf{Panel B:} We vary the forget factor $\gamma$ to correspond to an effective sample size of 1 year ($\gamma = 1/365$), 2 years ($\gamma = 1/730$), 3 years ($\gamma = 1/1095$), and 4 years ($\gamma = 1/1460$).
  \item \textbf{Panel C:} We compare the daily update scheme (1 observation per update) to a weekly update scheme (7 observations per update).
  \item \textbf{Panel D:} We compare the local model selection approach (see Section \ref{sec:online_model_selection}) to the global model selection approach.
\end{itemize}

\begin{figure}[!htb]
  \Description[
    Non-linear effects of the residual load on the location and scale parameter for the $t$-distribution.
  ]{
    Non-linear effects of the residual load on the location and scale parameter for the $t$-distribution. For the location parameter, we see a non-linear decrease in the residual load, i.e. for higher residual load, the price is lower. The scale parameter (i.e., the price volatility) increases if the residual load is very high or very low, leading to a $u$-shaped non-linear effect, while the strength of the effect changes through the out-of-sample set.
  }
  \centering
  \includegraphics[width=\linewidth]{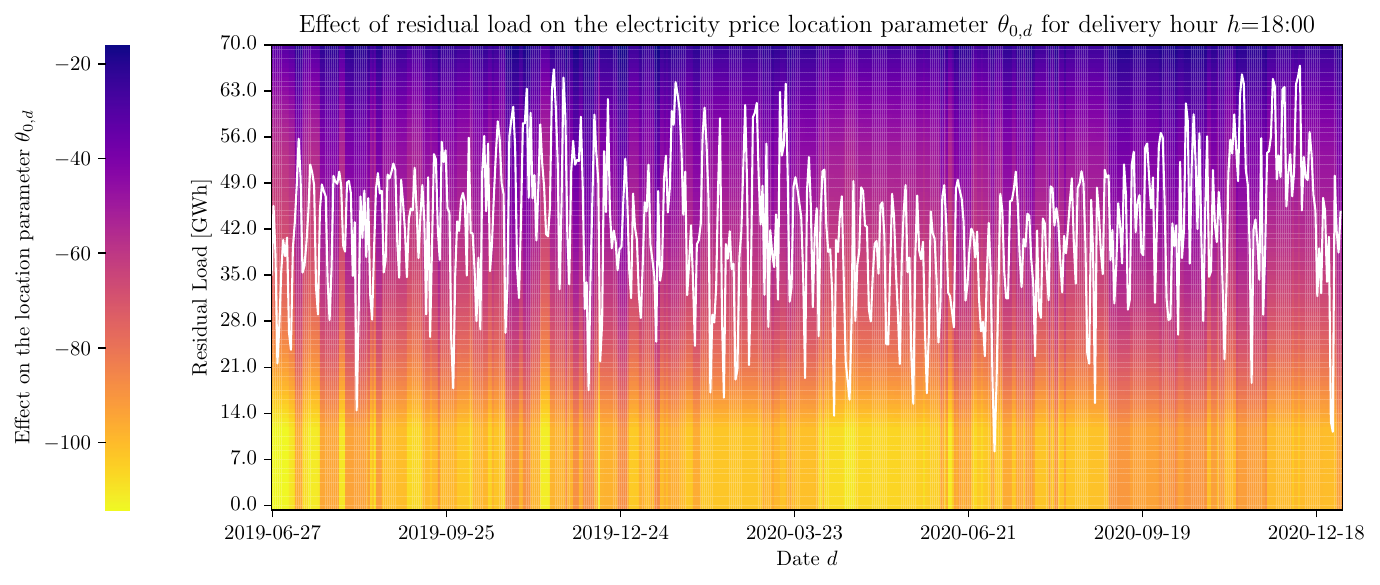}
  \caption{Non-linear effects of the residual load on the location and scale parameter for the $t$-distribution. The background coloring gives the non-linear effect for the value on the $x$-axis, and the $y$-axis denotes the update steps on the test set. The white line corresponds to the actual value. For the location, the effect on the location parameter is steeper towards the lower and upper end of the variable's domain. Low residual load corresponds to high renewable energy production, which leads to lower prices and vice versa.}
  \label{fig:results_epf_nonlinear_effect}
\end{figure}

As in Section \ref{sec:results}, we start our discussion of these additional results with a discussion of the in-sample results, before moving to the predictive performance. The non-linear effects in Figure \ref{fig:results_epf_nonlinear_effect} show the effect of the residual load on the location and shape parameters for the Student-$t$ distribution. For the location, we see a non-linearly increasing effect on the price level, i.e., the higher the residual load, the higher the electricity price. For the scale parameter, we see a $u$-shaped effect: Low residual load and very high residual load both contribute to large estimated scale parameters, i.e., high price volatility. Economically, this can be explained as low residual load corresponds to high renewable energy production, which inherently uncertain production volumes translate to price uncertainty, while very high residual load corresponds to a ``tight" supply situation, where the majority of power plants are already running, and therefore further changes in supply or demand will naturally induce more volatility.

\begin{table}[!htb]
  \resizebox{\textwidth}{!}{%
    \input{tables/electricity_price/01_epf_benchmarks}%
  }
  \newline
  \centering\includegraphics[width=0.75\textwidth]{tables/legend.pdf}
  \caption{%
    Results for the full, LASSO estimated batch and online distributional regression models and benchmark results from \cite{brusaferri2024line}, \cite{marcjasz2023distributional}, and traditional time series benchmarks. Note that \cite{brusaferri2024line} report only deciles~$\mathcal{Q} = \{0.1, \ldots, 0.9\}$, which are used for the calculation of the CRPS. Computation times for AutoARIMA(X) are based on parallelization on eight cores. Computation times for the studies of \cite{brusaferri2024line} and \cite{marcjasz2023distributional} have been provided by the authors in private discussions and are approximate and \emph{exclude} the hyperparameter tuning necessary for neural networks. For the coverage of the 50\% and 80\% prediction intervals, the background color corresponds to the number of hours where the \cite{kupiec1995techniques}-test is passed at the 5\%-level.
  }
  \label{tab:results_epf_benchmark}
\end{table}

We note some additional points regarding the predictive performance from Tables \ref{tab:results_epf_benchmark} and \ref{tab:results_epf_ablation}:
\begin{itemize}
  \item Note that some fields in Table \ref{tab:results_epf_benchmark} are empty. This is due to the fact that not all benchmark models deliver full distributional forecasts and might only provide quantile forecasts or only forecasts for the location parameter. For the AutoARIMA(X), we assume Gaussian residuals.
  \item In the ablation studies on the model selection, we see that the global and local model selection deliver very similar results, while the global model selection increases the computational cost by a factor of roughly 5-10 due to the high number of likelihood evaluations necessary. This result is somewhat surprising, as the assumption of Gaussian residuals for the local regressions, $\Vector{z}_k - \Matrix{X}_k\Vector{\beta}_k$, is somewhat unlikely to hold. On the other hand, both the batch LASSO implementation of \cite{ziel2021gamlss}, as well as the smooth-effect estimation of \cite{rigby2005generalized} use a local model selection approach for the selection of regularization parameters.
  \item Further improvements in terms of computational times can be achieved using batch updates. In Table \ref{tab:results_epf_ablation}, Panel C, we employ a weekly updating scheme (7 observations per update) compared to the daily update scheme. However, clearly, the predictive accuracy is traded off for the speed improvements, as the forecast horizon increases as well. Let us note that the computation time decreases only by about one-half due to the influence of the initial batch.
  \item The difference in model performance between RLS and OCD estimation is rather small, with OCD performing slightly better in most cases. This is expected, as unregularized estimation using OCD should converge to the least squares solution. Interestingly, the elastic net performs much worse than the LASSO in most cases. This might be due to the fact that the elastic net does not shrink to exactly zero, and hence the model complexity is higher than for the LASSO and the model selection cannot exploit sparsity.
\end{itemize}
Overall, these additional results further corroborate the findings of Section \ref{sec:results} and highlight the robust and computationally efficient performance of the online distributional regression approach. The variety of modelling and algorithmic choices available allows for a flexible adaptation to the needs of the forecaster.

\begin{table}
  \begin{center}
    \small Panel A: Changed estimation method.
    \resizebox{0.85\textwidth}{!}{\input{tables/electricity_price/02_epf_ablation_method}}%
    \\ [1ex]
    \small Panel B: Changed forget factor $\gamma$.
    \resizebox{0.85\textwidth}{!}{\input{tables/electricity_price/03_epf_ablation_forget}}%
    \\ [1ex]
    \small Panel C: Changed batch size: Daily vs. weekly online updating.
    \resizebox{0.85\textwidth}{!}{\input{tables/electricity_price/04_epf_ablation_batch}}%
    \\ [1ex]
    \small Panel D: Changed model selection from local RSS to global log-likelihood.
    \resizebox{0.85\textwidth}{!}{\input{tables/electricity_price/05_epf_ablation_model_selection}}%
    \\
    \includegraphics[width=0.75\textwidth]{tables/legend.pdf}
    \caption{%
      Results for the ablation studies: Changed estimation methods (a), changed forget $\gamma$ (b), changed mini-batch size (c) and changed model selection (d).
      The top three rows of each table present the base case, estimated as in Table \ref{tab:results_epf_batch_vs_online}. For each ablation study, we change one parameter.
      The best value in each column is marked \textbf{bold}.
      For the coverage of the 50\% and 80\% prediction intervals, the background color corresponds to the number of hours where the \cite{kupiec1995techniques}-test is passed at the 5\%-level.
    }\label{tab:results_epf_ablation}
  \end{center}
\end{table}

\FloatBarrier
\section{Additional Results for the Simulation Study}\label{app:simulation_study_additional_results}

\begin{figure}[!htb]
  \Description[
    True and Estimated Coefficients for all features, both parameters and selected (No, best, high) forget values of the third simulation study. The initial window is indicated by the shaded area.
  ]{
    True and Estimated Coefficients for all features, both parameters and selected (No, best, high) forget values of the third simulation study. The initial window is indicated by the shaded area. The estimated parameters follow the random changes in the true coefficients.
  }
  \centering
  \resizebox{0.98\textwidth}{!}{%
    \includegraphics[width=\linewidth]{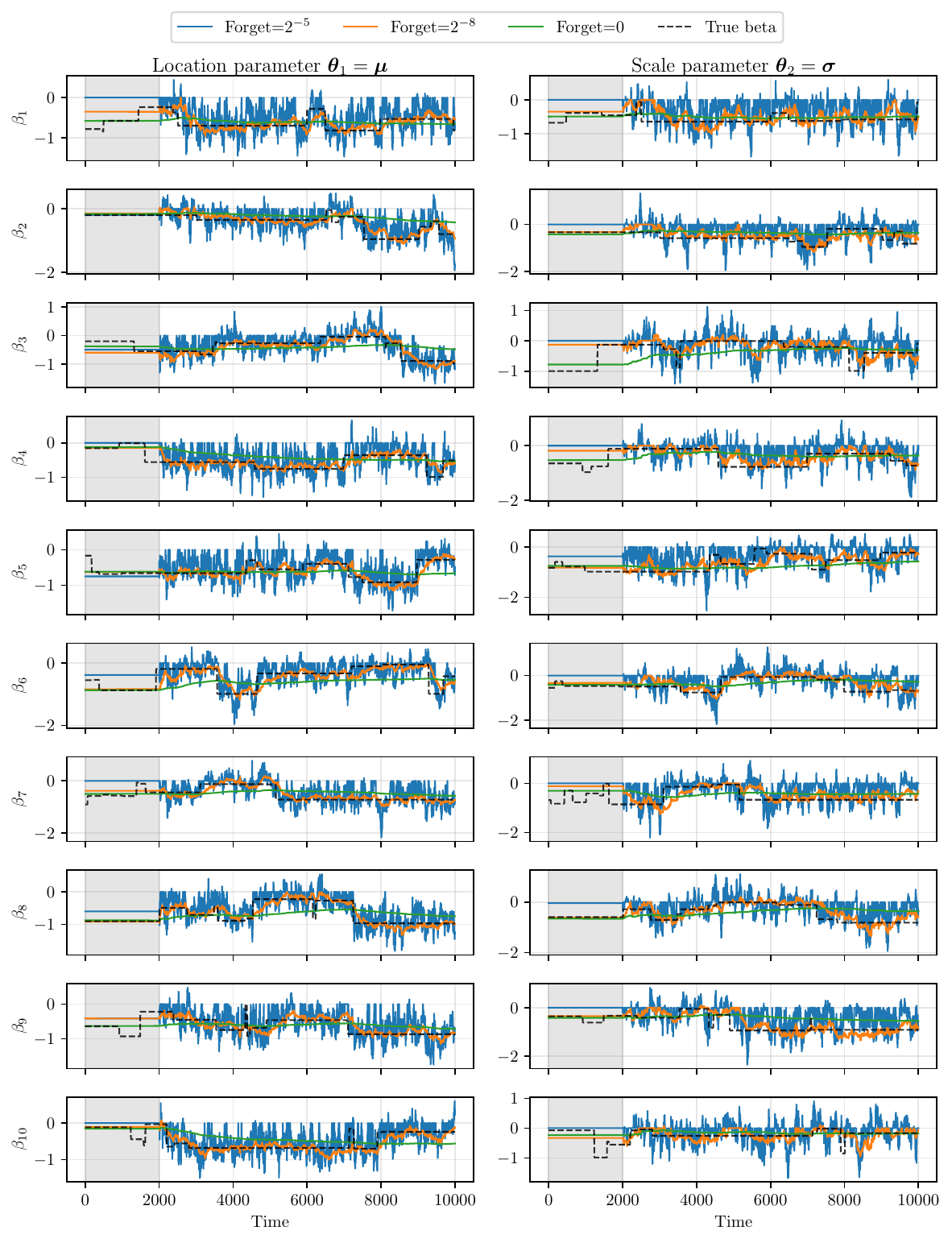}
  }
  \caption{True and Estimated Coefficients for all features, both parameters and selected (No, best, high) forget values of the third simulation study. The initial window is indicated by the shaded area.}
  \label{fig:simulation_study_params_forget}
\end{figure}
\FloatBarrier

\end{document}

%% file: tables/electricity_price/01_epf_batch_vs_online.tex
\begin{tabular}{llllrrrrrrrrr}
\toprule
Setting & Distribution & Method & K & MAE & RMSE & COV50 & COV80 & IS50 & IS80 & CRPS & LS & Time (min) \\
\midrule
\color{black} Batch & \color{black} Normal & \color{black} OLS & \color{black} 1 & {\cellcolor[HTML]{9DB7CA}} \color[HTML]{000000} \color{black} 4.402 & {\cellcolor[HTML]{74AEDA}} \color[HTML]{F1F1F1} \color{black} 6.314 & {\cellcolor[HTML]{8CB3D1}} \color[HTML]{000000} \color{black} 0.525 & {\cellcolor[HTML]{8CB3D1}} \color[HTML]{000000} \color{black} 0.816 & {\cellcolor[HTML]{9FB7CA}} \color[HTML]{000000} \color{black} 25.109 & {\cellcolor[HTML]{AABAC6}} \color[HTML]{000000} \color{black} 21.663 & {\cellcolor[HTML]{A0B7C9}} \color[HTML]{000000} \color{black} 3.252 & {\cellcolor[HTML]{D35D5F}} \color[HTML]{F1F1F1} \color{black} 3.256 & {\cellcolor[HTML]{9EB7CA}} \color[HTML]{000000} \color{black} 5.6 \\
\color{black} Batch & \color{black} Normal & \color{black} OLS & \color{black} 2 & {\cellcolor[HTML]{67ABDF}} \color[HTML]{F1F1F1} \color{black} \itshape 3.948 & {\cellcolor[HTML]{62AAE1}} \color[HTML]{F1F1F1} \color{black} \itshape 6.083 & {\cellcolor[HTML]{8CB3D1}} \color[HTML]{000000} \color{black} 0.474 & {\cellcolor[HTML]{C2AEAE}} \color[HTML]{000000} \color{black} 0.762 & {\cellcolor[HTML]{63AAE1}} \color[HTML]{F1F1F1} \color{black} 22.231 & {\cellcolor[HTML]{71ADDB}} \color[HTML]{F1F1F1} \color{black} \itshape 19.297 & {\cellcolor[HTML]{6DACDD}} \color[HTML]{F1F1F1} \color{black} \itshape 2.934 & {\cellcolor[HTML]{87B2D3}} \color[HTML]{000000} \color{black} 3.009 & {\cellcolor[HTML]{ACBAC5}} \color[HTML]{000000} \color{black} 9.0 \\
\color{black} Batch & \color{black} Normal & \color{black} LASSO & \color{black} 1 & {\cellcolor[HTML]{97B5CD}} \color[HTML]{000000} \color{black} 4.345 & {\cellcolor[HTML]{73AEDA}} \color[HTML]{F1F1F1} \color{black} 6.302 & {\cellcolor[HTML]{A5B9C7}} \color[HTML]{000000} \color{black} 0.539 & {\cellcolor[HTML]{9DB7CB}} \color[HTML]{000000} \color{black} 0.821 & {\cellcolor[HTML]{9AB6CC}} \color[HTML]{000000} \color{black} 24.845 & {\cellcolor[HTML]{ACBAC5}} \color[HTML]{000000} \color{black} 21.736 & {\cellcolor[HTML]{9DB7CB}} \color[HTML]{000000} \color{black} 3.230 & {\cellcolor[HTML]{D26263}} \color[HTML]{F1F1F1} \color{black} 3.248 & {\cellcolor[HTML]{AFBBC4}} \color[HTML]{000000} \color{black} 10.3 \\
\color{black} Batch & \color{black} Normal & \color{black} LASSO & \color{black} 2 & {\cellcolor[HTML]{65AAE0}} \color[HTML]{F1F1F1} \color{black} 3.928 & {\cellcolor[HTML]{63AAE0}} \color[HTML]{F1F1F1} \color{black} \itshape 6.101 & {\cellcolor[HTML]{8CB3D1}} \color[HTML]{000000} \color{black} 0.473 & {\cellcolor[HTML]{A5B9C7}} \color[HTML]{000000} \color{black} 0.767 & {\cellcolor[HTML]{64AAE0}} \color[HTML]{F1F1F1} \color{black} 22.281 & {\cellcolor[HTML]{66ABE0}} \color[HTML]{F1F1F1} \color{black} 18.867 & {\cellcolor[HTML]{67ABDF}} \color[HTML]{F1F1F1} \color{black} 2.898 & {\cellcolor[HTML]{78AFD9}} \color[HTML]{F1F1F1} \color{black} 2.985 & {\cellcolor[HTML]{C88E8F}} \color[HTML]{F1F1F1} \color{black} 95.7 \\
\color{black} Batch & \color{black} T & \color{black} OLS & \color{black} 1 & {\cellcolor[HTML]{6DACDD}} \color[HTML]{F1F1F1} \color{black} \itshape 3.991 & {\cellcolor[HTML]{5BA8E4}} \color[HTML]{F1F1F1} \color{black} 5.999 & {\cellcolor[HTML]{CE7576}} \color[HTML]{F1F1F1} \color{black} 0.421 & {\cellcolor[HTML]{CA8586}} \color[HTML]{F1F1F1} \color{black} 0.750 & {\cellcolor[HTML]{A8B9C6}} \color[HTML]{000000} \color{black} \itshape 25.553 & {\cellcolor[HTML]{84B1D4}} \color[HTML]{000000} \color{black} \itshape 20.094 & {\cellcolor[HTML]{74AEDA}} \color[HTML]{F1F1F1} \color{black} \itshape 2.975 & {\cellcolor[HTML]{A6B9C7}} \color[HTML]{000000} \color{black} \itshape 3.057 & {\cellcolor[HTML]{C59D9E}} \color[HTML]{F1F1F1} \color{black} 55.5 \\
\color{black} Batch & \color{black} T & \color{black} OLS & \color{black} 2 & {\cellcolor[HTML]{63AAE0}} \color[HTML]{F1F1F1} \color{black} \itshape 3.913 & {\cellcolor[HTML]{60A9E2}} \color[HTML]{F1F1F1} \color{black} \itshape 6.062 & {\cellcolor[HTML]{C0B6B6}} \color[HTML]{000000} \color{black} 0.453 & {\cellcolor[HTML]{C79596}} \color[HTML]{F1F1F1} \color{black} 0.751 & {\cellcolor[HTML]{6AABDE}} \color[HTML]{F1F1F1} \color{black} 22.527 & {\cellcolor[HTML]{66ABE0}} \color[HTML]{F1F1F1} \color{black} \itshape 18.877 & {\cellcolor[HTML]{65AAE0}} \color[HTML]{F1F1F1} \color{black} \itshape 2.888 & {\cellcolor[HTML]{6CACDD}} \color[HTML]{F1F1F1} \color{black} 2.967 & {\cellcolor[HTML]{C59C9D}} \color[HTML]{F1F1F1} \color{black} 56.7 \\
\color{black} Batch & \color{black} T & \color{black} OLS & \color{black} 3 & {\cellcolor[HTML]{62AAE1}} \color[HTML]{F1F1F1} \color{black} \itshape 3.902 & {\cellcolor[HTML]{5FA9E2}} \color[HTML]{F1F1F1} \color{black} 6.058 & {\cellcolor[HTML]{C3A5A6}} \color[HTML]{000000} \color{black} 0.451 & {\cellcolor[HTML]{C79596}} \color[HTML]{F1F1F1} \color{black} 0.749 & {\cellcolor[HTML]{6AACDE}} \color[HTML]{F1F1F1} \color{black} 22.593 & {\cellcolor[HTML]{63AAE1}} \color[HTML]{F1F1F1} \color{black} \itshape 18.751 & {\cellcolor[HTML]{62AAE1}} \color[HTML]{F1F1F1} \color{black} \itshape 2.871 & {\cellcolor[HTML]{79AFD8}} \color[HTML]{F1F1F1} \color{black} \itshape 2.988 & {\cellcolor[HTML]{C69B9B}} \color[HTML]{F1F1F1} \color{black} 59.9 \\
\color{black} Batch & \color{black} T & \color{black} LASSO & \color{black} 1 & {\cellcolor[HTML]{6CACDD}} \color[HTML]{F1F1F1} \color{black} \itshape 3.983 & {\cellcolor[HTML]{5CA8E3}} \color[HTML]{F1F1F1} \color{black} 6.007 & {\cellcolor[HTML]{CE7576}} \color[HTML]{F1F1F1} \color{black} 0.431 & {\cellcolor[HTML]{C3A5A6}} \color[HTML]{000000} \color{black} 0.757 & {\cellcolor[HTML]{A3B8C8}} \color[HTML]{000000} \color{black} 25.322 & {\cellcolor[HTML]{84B1D4}} \color[HTML]{000000} \color{black} 20.065 & {\cellcolor[HTML]{73AEDA}} \color[HTML]{F1F1F1} \color{black} 2.969 & {\cellcolor[HTML]{A3B8C8}} \color[HTML]{000000} \color{black} 3.051 & {\cellcolor[HTML]{C79394}} \color[HTML]{F1F1F1} \color{black} 78.9 \\
\color{black} Batch & \color{black} T & \color{black} LASSO & \color{black} 2 & {\cellcolor[HTML]{61AAE1}} \color[HTML]{F1F1F1} \color{black} \itshape 3.900 & {\cellcolor[HTML]{61AAE1}} \color[HTML]{F1F1F1} \color{black} \itshape 6.079 & {\cellcolor[HTML]{C59E9E}} \color[HTML]{F1F1F1} \color{black} 0.448 & {\cellcolor[HTML]{C98D8E}} \color[HTML]{F1F1F1} \color{black} 0.750 & {\cellcolor[HTML]{6EACDD}} \color[HTML]{F1F1F1} \color{black} 22.727 & {\cellcolor[HTML]{5FA9E2}} \color[HTML]{F1F1F1} \color{black} \itshape 18.642 & {\cellcolor[HTML]{62AAE1}} \color[HTML]{F1F1F1} \color{black} \itshape 2.869 & {\cellcolor[HTML]{63AAE0}} \color[HTML]{F1F1F1} \color{black} 2.955 & {\cellcolor[HTML]{CB8284}} \color[HTML]{F1F1F1} \color{black} 145.3 \\
\color{black} Batch & \color{black} T & \color{black} LASSO & \color{black} 3 & {\cellcolor[HTML]{61AAE1}} \color[HTML]{F1F1F1} \color{black} \itshape 3.897 & {\cellcolor[HTML]{61AAE1}} \color[HTML]{F1F1F1} \color{black} \itshape 6.081 & {\cellcolor[HTML]{C3A5A6}} \color[HTML]{000000} \color{black} 0.449 & {\cellcolor[HTML]{C98D8E}} \color[HTML]{F1F1F1} \color{black} 0.749 & {\cellcolor[HTML]{6EACDD}} \color[HTML]{F1F1F1} \color{black} 22.717 & {\cellcolor[HTML]{5EA9E3}} \color[HTML]{F1F1F1} \color{black} \itshape 18.581 & {\cellcolor[HTML]{61AAE1}} \color[HTML]{F1F1F1} \color{black} \itshape 2.863 & {\cellcolor[HTML]{63AAE0}} \color[HTML]{F1F1F1} \color{black} 2.954 & {\cellcolor[HTML]{CD7879}} \color[HTML]{F1F1F1} \color{black} 207.8 \\
\color{black} Batch & \color{black} JSU & \color{black} OLS & \color{black} 1 & {\cellcolor[HTML]{72ADDB}} \color[HTML]{F1F1F1} \color{black} \itshape 4.032 & {\cellcolor[HTML]{5FA9E2}} \color[HTML]{F1F1F1} \color{black} 6.047 & {\cellcolor[HTML]{BEBEBE}} \color[HTML]{000000} \color{black} 0.461 & {\cellcolor[HTML]{A5B9C7}} \color[HTML]{000000} \color{black} 0.770 & {\cellcolor[HTML]{9AB6CC}} \color[HTML]{000000} \color{black} \itshape 24.867 & {\cellcolor[HTML]{87B2D3}} \color[HTML]{000000} \color{black} \itshape 20.163 & {\cellcolor[HTML]{78AFD9}} \color[HTML]{F1F1F1} \color{black} \itshape 2.998 & {\cellcolor[HTML]{AFBBC4}} \color[HTML]{000000} \color{black} \itshape 3.072 & {\cellcolor[HTML]{C4A3A4}} \color[HTML]{F1F1F1} \color{black} 44.3 \\
\color{black} Batch & \color{black} JSU & \color{black} OLS & \color{black} 2 & {\cellcolor[HTML]{67ABDF}} \color[HTML]{F1F1F1} \color{black} \itshape 3.946 & {\cellcolor[HTML]{AEBAC4}} \color[HTML]{000000} \color{black} 7.099 & {\cellcolor[HTML]{C79596}} \color[HTML]{F1F1F1} \color{black} 0.435 & {\cellcolor[HTML]{D06D6F}} \color[HTML]{F1F1F1} \color{black} 0.738 & {\cellcolor[HTML]{76AED9}} \color[HTML]{F1F1F1} \color{black} \itshape 23.115 & {\cellcolor[HTML]{61AAE1}} \color[HTML]{F1F1F1} \color{black} \itshape 18.708 & {\cellcolor[HTML]{65AAE0}} \color[HTML]{F1F1F1} \color{black} \itshape 2.885 & {\cellcolor[HTML]{6AACDE}} \color[HTML]{F1F1F1} \color{black} \itshape 2.966 & {\cellcolor[HTML]{C4A1A1}} \color[HTML]{F1F1F1} \color{black} 47.9 \\
\color{black} Batch & \color{black} JSU & \color{black} OLS & \color{black} 3 & {\cellcolor[HTML]{7BAFD7}} \color[HTML]{F1F1F1} \color{black} \itshape 4.108 & {\cellcolor[HTML]{D35D5F}} \color[HTML]{F1F1F1} \color{black} 14.976 & {\cellcolor[HTML]{D16566}} \color[HTML]{F1F1F1} \color{black} 0.419 & {\cellcolor[HTML]{D35D5F}} \color[HTML]{F1F1F1} \color{black} 0.704 & {\cellcolor[HTML]{9FB7CA}} \color[HTML]{000000} \color{black} \itshape 25.094 & {\cellcolor[HTML]{80B0D5}} \color[HTML]{F1F1F1} \color{black} \itshape 19.915 & {\cellcolor[HTML]{7CAFD7}} \color[HTML]{F1F1F1} \color{black} \itshape 3.024 & {\cellcolor[HTML]{70ADDC}} \color[HTML]{F1F1F1} \color{black} \itshape 2.973 & {\cellcolor[HTML]{C69898}} \color[HTML]{F1F1F1} \color{black} 67.0 \\
\color{black} Batch & \color{black} JSU & \color{black} OLS & \color{black} 4 & {\cellcolor[HTML]{C4A1A1}} \color[HTML]{F1F1F1} \color{black} 4.999 & {\cellcolor[HTML]{D35D5F}} \color[HTML]{F1F1F1} \color{black} 8584.146 & {\cellcolor[HTML]{D16566}} \color[HTML]{F1F1F1} \color{black} 0.419 & {\cellcolor[HTML]{D35D5F}} \color[HTML]{F1F1F1} \color{black} 0.705 & {\cellcolor[HTML]{CE7475}} \color[HTML]{F1F1F1} \color{black} 31.168 & {\cellcolor[HTML]{D35D5F}} \color[HTML]{F1F1F1} \color{black} 33.018 & {\cellcolor[HTML]{D35F60}} \color[HTML]{F1F1F1} \color{black} 4.217 & {\cellcolor[HTML]{74AEDA}} \color[HTML]{F1F1F1} \color{black} \itshape 2.979 & {\cellcolor[HTML]{C69898}} \color[HTML]{F1F1F1} \color{black} 67.1 \\
\color{black} Batch & \color{black} JSU & \color{black} LASSO & \color{black} 1 & {\cellcolor[HTML]{6BACDD}} \color[HTML]{F1F1F1} \color{black} \itshape 3.978 & {\cellcolor[HTML]{5CA9E3}} \color[HTML]{F1F1F1} \color{black} 6.021 & {\cellcolor[HTML]{C2AEAE}} \color[HTML]{000000} \color{black} 0.469 & {\cellcolor[HTML]{8CB3D1}} \color[HTML]{000000} \color{black} 0.780 & {\cellcolor[HTML]{91B4CF}} \color[HTML]{000000} \color{black} \itshape 24.418 & {\cellcolor[HTML]{80B0D6}} \color[HTML]{F1F1F1} \color{black} \itshape 19.890 & {\cellcolor[HTML]{71ADDB}} \color[HTML]{F1F1F1} \color{black} \itshape 2.958 & {\cellcolor[HTML]{A2B8C9}} \color[HTML]{000000} \color{black} \itshape 3.050 & {\cellcolor[HTML]{CA8889}} \color[HTML]{F1F1F1} \color{black} 116.1 \\
\color{black} Batch & \color{black} JSU & \color{black} LASSO & \color{black} 2 & {\cellcolor[HTML]{60A9E2}} \color[HTML]{F1F1F1} \color{black} \itshape 3.888 & {\cellcolor[HTML]{7AAFD8}} \color[HTML]{F1F1F1} \color{black} 6.397 & {\cellcolor[HTML]{CE7576}} \color[HTML]{F1F1F1} \color{black} 0.432 & {\cellcolor[HTML]{D06D6F}} \color[HTML]{F1F1F1} \color{black} 0.741 & {\cellcolor[HTML]{7AAFD8}} \color[HTML]{F1F1F1} \color{black} \itshape 23.304 & {\cellcolor[HTML]{5EA9E3}} \color[HTML]{F1F1F1} \color{black} \itshape 18.589 & {\cellcolor[HTML]{5FA9E2}} \color[HTML]{F1F1F1} \color{black} \itshape 2.848 & {\cellcolor[HTML]{61AAE1}} \color[HTML]{F1F1F1} \color{black} \itshape 2.951 & {\cellcolor[HTML]{CE7577}} \color[HTML]{F1F1F1} \color{black} 224.8 \\
\color{black} Batch & \color{black} JSU & \color{black} LASSO & \color{black} 3 & {\cellcolor[HTML]{5BA8E4}} \color[HTML]{F1F1F1} \color{black} \itshape 3.851 & {\cellcolor[HTML]{A6B9C7}} \color[HTML]{000000} \color{black} 6.982 & {\cellcolor[HTML]{D06D6F}} \color[HTML]{F1F1F1} \color{black} 0.431 & {\cellcolor[HTML]{D35D5F}} \color[HTML]{F1F1F1} \color{black} 0.728 & {\cellcolor[HTML]{75AEDA}} \color[HTML]{F1F1F1} \color{black} \itshape 23.038 & {\cellcolor[HTML]{5AA8E4}} \color[HTML]{F1F1F1} \bfseries \color{black} \itshape 18.420 & {\cellcolor[HTML]{5AA8E4}} \color[HTML]{F1F1F1} \color{black} \itshape 2.822 & {\cellcolor[HTML]{5BA8E4}} \color[HTML]{F1F1F1} \color{black} \itshape 2.942 & {\cellcolor[HTML]{D16668}} \color[HTML]{F1F1F1} \color{black} 389.2 \\
\color{black} Batch & \color{black} JSU & \color{black} LASSO & \color{black} 4 & {\cellcolor[HTML]{5AA8E4}} \color[HTML]{F1F1F1} \bfseries \color{black} \itshape 3.839 & {\cellcolor[HTML]{A5B8C8}} \color[HTML]{000000} \color{black} 6.968 & {\cellcolor[HTML]{CC7D7E}} \color[HTML]{F1F1F1} \color{black} 0.431 & {\cellcolor[HTML]{D16566}} \color[HTML]{F1F1F1} \color{black} 0.733 & {\cellcolor[HTML]{72ADDB}} \color[HTML]{F1F1F1} \color{black} \itshape 22.950 & {\cellcolor[HTML]{5CA9E3}} \color[HTML]{F1F1F1} \color{black} \itshape 18.513 & {\cellcolor[HTML]{5AA8E4}} \color[HTML]{F1F1F1} \bfseries \color{black} \itshape 2.821 & {\cellcolor[HTML]{5AA8E4}} \color[HTML]{F1F1F1} \bfseries \color{black} \itshape 2.940 & {\cellcolor[HTML]{D35D5F}} \color[HTML]{F1F1F1} \color{black} 551.6 \\
\midrule
\color{black} Online & \color{black} Normal & \color{black} OLS & \color{black} 1 & {\cellcolor[HTML]{98B6CC}} \color[HTML]{000000} \color{black} \itshape 4.355 & {\cellcolor[HTML]{72ADDB}} \color[HTML]{F1F1F1} \color{black} \itshape 6.285 & {\cellcolor[HTML]{ADBAC4}} \color[HTML]{000000} \color{black} 0.532 & {\cellcolor[HTML]{94B5CE}} \color[HTML]{000000} \color{black} 0.815 & {\cellcolor[HTML]{9BB6CB}} \color[HTML]{000000} \color{black} \itshape 24.895 & {\cellcolor[HTML]{A7B9C7}} \color[HTML]{000000} \color{black} \itshape 21.530 & {\cellcolor[HTML]{9BB6CB}} \color[HTML]{000000} \color{black} \itshape 3.222 & {\cellcolor[HTML]{D35F60}} \color[HTML]{F1F1F1} \color{black} 3.254 & {\cellcolor[HTML]{5BA8E4}} \color[HTML]{F1F1F1} \color{black} \itshape 0.5 \\
\color{black} Online & \color{black} Normal & \color{black} OLS & \color{black} 2 & {\cellcolor[HTML]{6DACDD}} \color[HTML]{F1F1F1} \color{black} 3.990 & {\cellcolor[HTML]{65AAE0}} \color[HTML]{F1F1F1} \color{black} 6.125 & {\cellcolor[HTML]{73AEDA}} \color[HTML]{F1F1F1} \bfseries \color{black} 0.498 & {\cellcolor[HTML]{94B5CE}} \color[HTML]{000000} \color{black} 0.785 & {\cellcolor[HTML]{5CA9E3}} \color[HTML]{F1F1F1} \color{black} \itshape 21.937 & {\cellcolor[HTML]{79AFD8}} \color[HTML]{F1F1F1} \color{black} 19.618 & {\cellcolor[HTML]{73AEDA}} \color[HTML]{F1F1F1} \color{black} 2.972 & {\cellcolor[HTML]{87B2D3}} \color[HTML]{000000} \color{black} 3.008 & {\cellcolor[HTML]{5AA8E4}} \color[HTML]{F1F1F1} \color{black} \itshape 0.5 \\
\color{black} Online & \color{black} Normal & \color{black} LASSO & \color{black} 1 & {\cellcolor[HTML]{98B6CC}} \color[HTML]{000000} \color{black} 4.355 & {\cellcolor[HTML]{73AEDA}} \color[HTML]{F1F1F1} \color{black} 6.305 & {\cellcolor[HTML]{A5B9C7}} \color[HTML]{000000} \color{black} 0.536 & {\cellcolor[HTML]{9DB7CB}} \color[HTML]{000000} \color{black} 0.819 & {\cellcolor[HTML]{9CB6CB}} \color[HTML]{000000} \color{black} 24.947 & {\cellcolor[HTML]{ACBAC5}} \color[HTML]{000000} \color{black} 21.730 & {\cellcolor[HTML]{9DB7CA}} \color[HTML]{000000} \color{black} 3.234 & {\cellcolor[HTML]{D26062}} \color[HTML]{F1F1F1} \color{black} 3.251 & {\cellcolor[HTML]{64AAE0}} \color[HTML]{F1F1F1} \color{black} \itshape 0.8 \\
\color{black} Online & \color{black} Normal & \color{black} LASSO & \color{black} 2 & {\cellcolor[HTML]{67ABDF}} \color[HTML]{F1F1F1} \color{black} 3.941 & {\cellcolor[HTML]{67ABDF}} \color[HTML]{F1F1F1} \color{black} 6.145 & {\cellcolor[HTML]{6AACDE}} \color[HTML]{F1F1F1} \color{black} 0.497 & {\cellcolor[HTML]{7BAFD7}} \color[HTML]{F1F1F1} \color{black} 0.788 & {\cellcolor[HTML]{5AA8E4}} \color[HTML]{F1F1F1} \bfseries \color{black} \itshape 21.822 & {\cellcolor[HTML]{65AAE0}} \color[HTML]{F1F1F1} \color{black} 18.842 & {\cellcolor[HTML]{67ABDF}} \color[HTML]{F1F1F1} \color{black} 2.898 & {\cellcolor[HTML]{75AEDA}} \color[HTML]{F1F1F1} \color{black} 2.981 & {\cellcolor[HTML]{75AEDA}} \color[HTML]{F1F1F1} \color{black} \itshape 1.4 \\
\color{black} Online & \color{black} T & \color{black} OLS & \color{black} 1 & {\cellcolor[HTML]{74AEDA}} \color[HTML]{F1F1F1} \color{black} 4.048 & {\cellcolor[HTML]{5CA8E3}} \color[HTML]{F1F1F1} \color{black} 6.008 & {\cellcolor[HTML]{CE7576}} \color[HTML]{F1F1F1} \color{black} 0.413 & {\cellcolor[HTML]{CE7576}} \color[HTML]{F1F1F1} \color{black} 0.743 & {\cellcolor[HTML]{B2BBC3}} \color[HTML]{000000} \color{black} 26.077 & {\cellcolor[HTML]{8CB3D1}} \color[HTML]{000000} \color{black} 20.401 & {\cellcolor[HTML]{7BAFD7}} \color[HTML]{F1F1F1} \color{black} 3.018 & {\cellcolor[HTML]{B0BBC3}} \color[HTML]{000000} \color{black} 3.073 & {\cellcolor[HTML]{66ABE0}} \color[HTML]{F1F1F1} \color{black} \itshape 0.8 \\
\color{black} Online & \color{black} T & \color{black} OLS & \color{black} 2 & {\cellcolor[HTML]{67ABDF}} \color[HTML]{F1F1F1} \color{black} 3.950 & {\cellcolor[HTML]{62AAE1}} \color[HTML]{F1F1F1} \color{black} 6.081 & {\cellcolor[HTML]{8CB3D1}} \color[HTML]{000000} \color{black} 0.477 & {\cellcolor[HTML]{ADBAC4}} \color[HTML]{000000} \color{black} 0.774 & {\cellcolor[HTML]{62AAE1}} \color[HTML]{F1F1F1} \color{black} \itshape 22.193 & {\cellcolor[HTML]{6EACDD}} \color[HTML]{F1F1F1} \color{black} 19.185 & {\cellcolor[HTML]{6BACDD}} \color[HTML]{F1F1F1} \color{black} 2.925 & {\cellcolor[HTML]{6EACDC}} \color[HTML]{F1F1F1} \color{black} 2.971 & {\cellcolor[HTML]{67ABDF}} \color[HTML]{F1F1F1} \color{black} \itshape 0.8 \\
\color{black} Online & \color{black} T & \color{black} OLS & \color{black} 3 & {\cellcolor[HTML]{69ABDE}} \color[HTML]{F1F1F1} \color{black} 3.959 & {\cellcolor[HTML]{62AAE1}} \color[HTML]{F1F1F1} \color{black} 6.081 & {\cellcolor[HTML]{73AEDA}} \color[HTML]{F1F1F1} \color{black} 0.486 & {\cellcolor[HTML]{8CB3D1}} \color[HTML]{000000} \bfseries \color{black} 0.791 & {\cellcolor[HTML]{61AAE1}} \color[HTML]{F1F1F1} \color{black} \itshape 22.141 & {\cellcolor[HTML]{6EACDC}} \color[HTML]{F1F1F1} \color{black} 19.206 & {\cellcolor[HTML]{6BACDD}} \color[HTML]{F1F1F1} \color{black} 2.922 & {\cellcolor[HTML]{84B1D4}} \color[HTML]{000000} \color{black} 3.003 & {\cellcolor[HTML]{67ABDF}} \color[HTML]{F1F1F1} \color{black} \itshape 0.8 \\
\color{black} Online & \color{black} T & \color{black} LASSO & \color{black} 1 & {\cellcolor[HTML]{6FADDC}} \color[HTML]{F1F1F1} \color{black} 4.011 & {\cellcolor[HTML]{5AA8E4}} \color[HTML]{F1F1F1} \bfseries \color{black} 5.985 & {\cellcolor[HTML]{C98D8E}} \color[HTML]{F1F1F1} \color{black} 0.431 & {\cellcolor[HTML]{C3A5A6}} \color[HTML]{000000} \color{black} 0.757 & {\cellcolor[HTML]{A6B9C7}} \color[HTML]{000000} \color{black} 25.447 & {\cellcolor[HTML]{86B2D3}} \color[HTML]{000000} \color{black} 20.154 & {\cellcolor[HTML]{75AEDA}} \color[HTML]{F1F1F1} \color{black} 2.986 & {\cellcolor[HTML]{A6B9C7}} \color[HTML]{000000} \color{black} 3.057 & {\cellcolor[HTML]{6EACDC}} \color[HTML]{F1F1F1} \color{black} \itshape 1.1 \\
\color{black} Online & \color{black} T & \color{black} LASSO & \color{black} 2 & {\cellcolor[HTML]{64AAE0}} \color[HTML]{F1F1F1} \color{black} 3.922 & {\cellcolor[HTML]{65AAE0}} \color[HTML]{F1F1F1} \color{black} 6.127 & {\cellcolor[HTML]{9DB7CB}} \color[HTML]{000000} \color{black} 0.472 & {\cellcolor[HTML]{94B5CE}} \color[HTML]{000000} \color{black} 0.776 & {\cellcolor[HTML]{64AAE0}} \color[HTML]{F1F1F1} \color{black} \itshape 22.310 & {\cellcolor[HTML]{66ABE0}} \color[HTML]{F1F1F1} \color{black} 18.891 & {\cellcolor[HTML]{66ABE0}} \color[HTML]{F1F1F1} \color{black} 2.892 & {\cellcolor[HTML]{63AAE1}} \color[HTML]{F1F1F1} \color{black} 2.954 & {\cellcolor[HTML]{7CAFD7}} \color[HTML]{F1F1F1} \color{black} \itshape 1.7 \\
\color{black} Online & \color{black} T & \color{black} LASSO & \color{black} 3 & {\cellcolor[HTML]{63AAE0}} \color[HTML]{F1F1F1} \color{black} 3.918 & {\cellcolor[HTML]{63AAE0}} \color[HTML]{F1F1F1} \color{black} 6.109 & {\cellcolor[HTML]{8CB3D1}} \color[HTML]{000000} \color{black} 0.473 & {\cellcolor[HTML]{94B5CE}} \color[HTML]{000000} \color{black} 0.776 & {\cellcolor[HTML]{65AAE0}} \color[HTML]{F1F1F1} \color{black} \itshape 22.339 & {\cellcolor[HTML]{67ABDF}} \color[HTML]{F1F1F1} \color{black} 18.923 & {\cellcolor[HTML]{66ABE0}} \color[HTML]{F1F1F1} \color{black} 2.893 & {\cellcolor[HTML]{62AAE1}} \color[HTML]{F1F1F1} \color{black} 2.953 & {\cellcolor[HTML]{8DB3D1}} \color[HTML]{000000} \color{black} \itshape 3.1 \\
\color{black} Online & \color{black} JSU & \color{black} OLS & \color{black} 1 & {\cellcolor[HTML]{79AFD8}} \color[HTML]{F1F1F1} \color{black} 4.095 & {\cellcolor[HTML]{5FA9E2}} \color[HTML]{F1F1F1} \color{black} 6.053 & {\cellcolor[HTML]{C0B6B6}} \color[HTML]{000000} \color{black} 0.454 & {\cellcolor[HTML]{BEBEBE}} \color[HTML]{000000} \color{black} 0.763 & {\cellcolor[HTML]{A5B8C8}} \color[HTML]{000000} \color{black} 25.396 & {\cellcolor[HTML]{8FB4D0}} \color[HTML]{000000} \color{black} 20.503 & {\cellcolor[HTML]{80B0D6}} \color[HTML]{F1F1F1} \color{black} 3.045 & {\cellcolor[HTML]{BEBDBD}} \color[HTML]{000000} \color{black} 3.096 & {\cellcolor[HTML]{6BACDD}} \color[HTML]{F1F1F1} \color{black} \itshape 1.0 \\
\color{black} Online & \color{black} JSU & \color{black} OLS & \color{black} 2 & {\cellcolor[HTML]{71ADDB}} \color[HTML]{F1F1F1} \color{black} 4.022 & {\cellcolor[HTML]{90B4CF}} \color[HTML]{000000} \color{black} 6.683 & {\cellcolor[HTML]{D16566}} \color[HTML]{F1F1F1} \color{black} 0.428 & {\cellcolor[HTML]{CE7576}} \color[HTML]{F1F1F1} \color{black} 0.734 & {\cellcolor[HTML]{84B1D4}} \color[HTML]{000000} \color{black} 23.736 & {\cellcolor[HTML]{6EACDD}} \color[HTML]{F1F1F1} \color{black} 19.180 & {\cellcolor[HTML]{6FADDC}} \color[HTML]{F1F1F1} \color{black} 2.947 & {\cellcolor[HTML]{8BB3D2}} \color[HTML]{000000} \color{black} 3.014 & {\cellcolor[HTML]{6CACDD}} \color[HTML]{F1F1F1} \color{black} \itshape 1.0 \\
\color{black} Online & \color{black} JSU & \color{black} OLS & \color{black} 3 & {\cellcolor[HTML]{C59E9E}} \color[HTML]{F1F1F1} \color{black} 5.033 & {\cellcolor[HTML]{D35D5F}} \color[HTML]{F1F1F1} \color{black} 14.133 & {\cellcolor[HTML]{D16566}} \color[HTML]{F1F1F1} \color{black} 0.401 & {\cellcolor[HTML]{D06D6F}} \color[HTML]{F1F1F1} \color{black} 0.681 & {\cellcolor[HTML]{D06D6F}} \color[HTML]{F1F1F1} \color{black} 31.619 & {\cellcolor[HTML]{C98D8E}} \color[HTML]{F1F1F1} \color{black} 24.982 & {\cellcolor[HTML]{C69899}} \color[HTML]{F1F1F1} \color{black} 3.735 & {\cellcolor[HTML]{C4A1A1}} \color[HTML]{F1F1F1} \color{black} 3.142 & {\cellcolor[HTML]{6DACDD}} \color[HTML]{F1F1F1} \color{black} \itshape 1.0 \\
\color{black} Online & \color{black} JSU & \color{black} OLS & \color{black} 4 & {\cellcolor[HTML]{D35D5F}} \color[HTML]{F1F1F1} \color{black} $>10^4$ & {\cellcolor[HTML]{D35D5F}} \color[HTML]{F1F1F1} \color{black} $>10^4$ & {\cellcolor[HTML]{D35D5F}} \color[HTML]{F1F1F1} \color{black} 0.396 & {\cellcolor[HTML]{D35D5F}} \color[HTML]{F1F1F1} \color{black} 0.674 & {\cellcolor[HTML]{D35D5F}} \color[HTML]{F1F1F1} \color{black} $>10^4$ & {\cellcolor[HTML]{D35D5F}} \color[HTML]{F1F1F1} \color{black} $>10^4$ & {\cellcolor[HTML]{D35D5F}} \color[HTML]{F1F1F1} \color{black} $>10^4$ & {\cellcolor[HTML]{CA8889}} \color[HTML]{F1F1F1} \color{black} 3.184 & {\cellcolor[HTML]{6EACDD}} \color[HTML]{F1F1F1} \color{black} \itshape 1.0 \\
\color{black} Online & \color{black} JSU & \color{black} LASSO & \color{black} 1 & {\cellcolor[HTML]{73AEDA}} \color[HTML]{F1F1F1} \color{black} 4.041 & {\cellcolor[HTML]{5CA8E3}} \color[HTML]{F1F1F1} \color{black} 6.010 & {\cellcolor[HTML]{C2AEAE}} \color[HTML]{000000} \color{black} 0.458 & {\cellcolor[HTML]{C0B6B6}} \color[HTML]{000000} \color{black} 0.768 & {\cellcolor[HTML]{9DB7CB}} \color[HTML]{000000} \color{black} 24.987 & {\cellcolor[HTML]{87B2D3}} \color[HTML]{000000} \color{black} 20.208 & {\cellcolor[HTML]{79AFD8}} \color[HTML]{F1F1F1} \color{black} 3.003 & {\cellcolor[HTML]{ADBAC4}} \color[HTML]{000000} \color{black} 3.068 & {\cellcolor[HTML]{75AEDA}} \color[HTML]{F1F1F1} \color{black} \itshape 1.4 \\
\color{black} Online & \color{black} JSU & \color{black} LASSO & \color{black} 2 & {\cellcolor[HTML]{68ABDF}} \color[HTML]{F1F1F1} \color{black} 3.954 & {\cellcolor[HTML]{72ADDB}} \color[HTML]{F1F1F1} \color{black} 6.295 & {\cellcolor[HTML]{D06D6F}} \color[HTML]{F1F1F1} \color{black} 0.430 & {\cellcolor[HTML]{CC7D7E}} \color[HTML]{F1F1F1} \color{black} 0.738 & {\cellcolor[HTML]{83B1D5}} \color[HTML]{000000} \color{black} 23.710 & {\cellcolor[HTML]{67ABDF}} \color[HTML]{F1F1F1} \color{black} 18.920 & {\cellcolor[HTML]{67ABDF}} \color[HTML]{F1F1F1} \color{black} 2.900 & {\cellcolor[HTML]{79AFD8}} \color[HTML]{F1F1F1} \color{black} 2.986 & {\cellcolor[HTML]{82B1D5}} \color[HTML]{000000} \color{black} \itshape 2.1 \\
\color{black} Online & \color{black} JSU & \color{black} LASSO & \color{black} 3 & {\cellcolor[HTML]{66ABE0}} \color[HTML]{F1F1F1} \color{black} 3.932 & {\cellcolor[HTML]{7DB0D7}} \color[HTML]{F1F1F1} \color{black} 6.422 & {\cellcolor[HTML]{D06D6F}} \color[HTML]{F1F1F1} \color{black} 0.421 & {\cellcolor[HTML]{D06D6F}} \color[HTML]{F1F1F1} \color{black} 0.718 & {\cellcolor[HTML]{86B2D3}} \color[HTML]{000000} \color{black} 23.848 & {\cellcolor[HTML]{67ABDF}} \color[HTML]{F1F1F1} \color{black} 18.938 & {\cellcolor[HTML]{66ABE0}} \color[HTML]{F1F1F1} \color{black} 2.891 & {\cellcolor[HTML]{78AFD9}} \color[HTML]{F1F1F1} \color{black} 2.986 & {\cellcolor[HTML]{91B4CF}} \color[HTML]{000000} \color{black} \itshape 3.5 \\
\color{black} Online & \color{black} JSU & \color{black} LASSO & \color{black} 4 & {\cellcolor[HTML]{5FA9E2}} \color[HTML]{F1F1F1} \color{black} 3.877 & {\cellcolor[HTML]{D35D5F}} \color[HTML]{F1F1F1} \color{black} 8.985 & {\cellcolor[HTML]{D06D6F}} \color[HTML]{F1F1F1} \color{black} 0.419 & {\cellcolor[HTML]{D16566}} \color[HTML]{F1F1F1} \color{black} 0.716 & {\cellcolor[HTML]{82B1D5}} \color[HTML]{000000} \color{black} 23.673 & {\cellcolor[HTML]{63AAE1}} \color[HTML]{F1F1F1} \color{black} 18.764 & {\cellcolor[HTML]{5FA9E2}} \color[HTML]{F1F1F1} \color{black} 2.855 & {\cellcolor[HTML]{6EACDD}} \color[HTML]{F1F1F1} \color{black} 2.971 & {\cellcolor[HTML]{9DB7CB}} \color[HTML]{000000} \color{black} \itshape 5.4 \\
\bottomrule
\end{tabular}

%% file: tables/electricity_price/01_epf_benchmarks.tex
\begin{tabular}{lllrrrrrrrrr}
\toprule
Setting & Distribution & Method & MAE & RMSE & COV50 & COV80 & IS50 & IS80 & CRPS & LS & Time (min) \\
\midrule
\color{black} Online & \color{black} Normal & \color{black} LASSO & {\cellcolor[HTML]{8BB3D1}} \color[HTML]{000000} \color{black} 3.941 & {\cellcolor[HTML]{5FA9E2}} \color[HTML]{F1F1F1} \color{black} 6.145 & {\cellcolor[HTML]{6AACDE}} \color[HTML]{F1F1F1} \bfseries \color{black} 0.497 & {\cellcolor[HTML]{7BAFD7}} \color[HTML]{F1F1F1} \color{black} 0.788 & {\cellcolor[HTML]{D35D5F}} \color[HTML]{F1F1F1} \color{black} 21.822 & {\cellcolor[HTML]{92B4CF}} \color[HTML]{000000} \color{black} 18.842 & {\cellcolor[HTML]{8FB4D0}} \color[HTML]{000000} \color{black} 2.898 & {\cellcolor[HTML]{63AAE1}} \color[HTML]{F1F1F1} \color{black} 2.981 & {\cellcolor[HTML]{78AFD9}} \color[HTML]{F1F1F1} \color{black} 1.4 \\
\color{black} Online & \color{black} T & \color{black} LASSO & {\cellcolor[HTML]{88B2D2}} \color[HTML]{000000} \color{black} 3.918 & {\cellcolor[HTML]{5CA8E3}} \color[HTML]{F1F1F1} \color{black} 6.109 & {\cellcolor[HTML]{8CB3D1}} \color[HTML]{000000} \color{black} 0.473 & {\cellcolor[HTML]{94B5CE}} \color[HTML]{000000} \color{black} 0.776 & {\cellcolor[HTML]{D35D5F}} \color[HTML]{F1F1F1} \color{black} 22.339 & {\cellcolor[HTML]{95B5CE}} \color[HTML]{000000} \color{black} 18.923 & {\cellcolor[HTML]{8FB4D0}} \color[HTML]{000000} \color{black} 2.893 & {\cellcolor[HTML]{5EA9E3}} \color[HTML]{F1F1F1} \color{black} 2.953 & {\cellcolor[HTML]{86B2D3}} \color[HTML]{000000} \color{black} 3.1 \\
\color{black} Online & \color{black} JSU & \color{black} LASSO & {\cellcolor[HTML]{84B1D4}} \color[HTML]{000000} \color{black} 3.877 & {\cellcolor[HTML]{D26466}} \color[HTML]{F1F1F1} \color{black} 8.985 & {\cellcolor[HTML]{D06D6F}} \color[HTML]{F1F1F1} \color{black} 0.419 & {\cellcolor[HTML]{D16566}} \color[HTML]{F1F1F1} \color{black} 0.716 & {\cellcolor[HTML]{D35D5F}} \color[HTML]{F1F1F1} \color{black} 23.673 & {\cellcolor[HTML]{90B4CF}} \color[HTML]{000000} \color{black} 18.764 & {\cellcolor[HTML]{88B2D2}} \color[HTML]{000000} \color{black} 2.855 & {\cellcolor[HTML]{61AAE1}} \color[HTML]{F1F1F1} \color{black} 2.971 & {\cellcolor[HTML]{8FB4D0}} \color[HTML]{000000} \color{black} 5.4 \\
\midrule
\color{black} Batch & \color{black} Normal & \color{black} LASSO & {\cellcolor[HTML]{8AB3D2}} \color[HTML]{000000} \color{black} 3.928 & {\cellcolor[HTML]{5CA8E3}} \color[HTML]{F1F1F1} \color{black} 6.101 & {\cellcolor[HTML]{8CB3D1}} \color[HTML]{000000} \color{black} 0.473 & {\cellcolor[HTML]{A5B9C7}} \color[HTML]{000000} \color{black} 0.767 & {\cellcolor[HTML]{D35D5F}} \color[HTML]{F1F1F1} \color{black} 22.281 & {\cellcolor[HTML]{93B5CE}} \color[HTML]{000000} \color{black} 18.867 & {\cellcolor[HTML]{90B4CF}} \color[HTML]{000000} \color{black} 2.898 & {\cellcolor[HTML]{63AAE0}} \color[HTML]{F1F1F1} \color{black} 2.985 & {\cellcolor[HTML]{BEBDBD}} \color[HTML]{000000} \color{black} 95.7 \\
\color{black} Batch & \color{black} T & \color{black} LASSO & {\cellcolor[HTML]{86B2D3}} \color[HTML]{000000} \color{black} 3.897 & {\cellcolor[HTML]{5AA8E4}} \color[HTML]{F1F1F1} \bfseries \color{black} 6.081 & {\cellcolor[HTML]{C3A5A6}} \color[HTML]{000000} \color{black} 0.449 & {\cellcolor[HTML]{C98D8E}} \color[HTML]{F1F1F1} \color{black} 0.749 & {\cellcolor[HTML]{D35D5F}} \color[HTML]{F1F1F1} \color{black} 22.717 & {\cellcolor[HTML]{8BB3D1}} \color[HTML]{000000} \color{black} 18.581 & {\cellcolor[HTML]{8AB3D2}} \color[HTML]{000000} \color{black} 2.863 & {\cellcolor[HTML]{5FA9E2}} \color[HTML]{F1F1F1} \color{black} 2.954 & {\cellcolor[HTML]{C1B1B1}} \color[HTML]{000000} \color{black} 207.8 \\
\color{black} Batch & \color{black} JSU & \color{black} LASSO & {\cellcolor[HTML]{7EB0D6}} \color[HTML]{F1F1F1} \color{black} 3.839 & {\cellcolor[HTML]{9DB7CB}} \color[HTML]{000000} \color{black} 6.968 & {\cellcolor[HTML]{CC7D7E}} \color[HTML]{F1F1F1} \color{black} 0.431 & {\cellcolor[HTML]{D16566}} \color[HTML]{F1F1F1} \color{black} 0.733 & {\cellcolor[HTML]{D35D5F}} \color[HTML]{F1F1F1} \color{black} 22.950 & {\cellcolor[HTML]{8AB3D2}} \color[HTML]{000000} \color{black} 18.513 & {\cellcolor[HTML]{83B1D5}} \color[HTML]{000000} \color{black} 2.821 & {\cellcolor[HTML]{5CA9E3}} \color[HTML]{F1F1F1} \color{black} 2.940 & {\cellcolor[HTML]{C4A1A1}} \color[HTML]{F1F1F1} \color{black} 551.6 \\
\midrule
\color{black} Linear & \color{black} Normal & \color{black} LASSO & {\cellcolor[HTML]{A1B8C9}} \color[HTML]{000000} \color{black} 4.110 & {\cellcolor[HTML]{9EB7CA}} \color[HTML]{000000} \color{black} 6.987 & {\cellcolor[HTML]{B6BCC1}} \color[HTML]{000000} \color{black} 0.547 & {\cellcolor[HTML]{94B5CE}} \color[HTML]{000000} \color{black} 0.817 & {\cellcolor[HTML]{D35D5F}} \color[HTML]{F1F1F1} \color{black} 23.694 & {\cellcolor[HTML]{C2ADAD}} \color[HTML]{000000} \color{black} 21.306 & {\cellcolor[HTML]{B5BCC1}} \color[HTML]{000000} \color{black} 3.123 & {\cellcolor[HTML]{74AEDA}} \color[HTML]{F1F1F1} \color{black} 3.084 & {\cellcolor[HTML]{74AEDA}} \color[HTML]{F1F1F1} \color{black} 1.1 \\
\color{black} Linear & \color{black} T & \color{black} LASSO & {\cellcolor[HTML]{9FB7CA}} \color[HTML]{000000} \color{black} 4.099 & {\cellcolor[HTML]{A6B9C7}} \color[HTML]{000000} \color{black} 7.100 & {\cellcolor[HTML]{73AEDA}} \color[HTML]{F1F1F1} \color{black} 0.485 & {\cellcolor[HTML]{9DB7CB}} \color[HTML]{000000} \color{black} 0.783 & {\cellcolor[HTML]{D35D5F}} \color[HTML]{F1F1F1} \color{black} 25.331 & {\cellcolor[HTML]{C2ADAD}} \color[HTML]{000000} \color{black} 21.304 & {\cellcolor[HTML]{B0BBC3}} \color[HTML]{000000} \color{black} 3.097 & {\cellcolor[HTML]{65AAE0}} \color[HTML]{F1F1F1} \color{black} 2.992 & {\cellcolor[HTML]{80B0D6}} \color[HTML]{F1F1F1} \color{black} 2.2 \\
\color{black} Linear & \color{black} JSU & \color{black} LASSO & {\cellcolor[HTML]{82B1D5}} \color[HTML]{000000} \color{black} 3.867 & {\cellcolor[HTML]{87B2D3}} \color[HTML]{000000} \color{black} 6.662 & {\cellcolor[HTML]{C0B6B6}} \color[HTML]{000000} \color{black} 0.459 & {\cellcolor[HTML]{C0B6B6}} \color[HTML]{000000} \color{black} 0.760 & {\cellcolor[HTML]{D35D5F}} \color[HTML]{F1F1F1} \color{black} 22.979 & {\cellcolor[HTML]{87B2D3}} \color[HTML]{000000} \color{black} 18.432 & {\cellcolor[HTML]{84B1D4}} \color[HTML]{000000} \color{black} 2.828 & {\cellcolor[HTML]{5AA8E4}} \color[HTML]{F1F1F1} \bfseries \color{black} 2.926 & {\cellcolor[HTML]{85B1D4}} \color[HTML]{000000} \color{black} 3.0 \\
\midrule
\color{black} Persistence & \color{black} Normal & \color{black} - & {\cellcolor[HTML]{D35D5F}} \color[HTML]{F1F1F1} \color{black} 12.244 & {\cellcolor[HTML]{D35D5F}} \color[HTML]{F1F1F1} \color{black} 17.487 & {\cellcolor[HTML]{CE7576}} \color[HTML]{F1F1F1} \color{black} 0.519 & {\cellcolor[HTML]{C0B6B6}} \color[HTML]{000000} \color{black} 0.769 & {\cellcolor[HTML]{D35D5F}} \color[HTML]{F1F1F1} \color{black} 75.498 & {\cellcolor[HTML]{D35D5F}} \color[HTML]{F1F1F1} \color{black} 64.138 & {\cellcolor[HTML]{D35D5F}} \color[HTML]{F1F1F1} \color{black} 9.275 & {\cellcolor[HTML]{D26263}} \color[HTML]{F1F1F1} \color{black} 4.345 & {\cellcolor[HTML]{5AA8E4}} \color[HTML]{F1F1F1} \color{black} 0.2 \\
\color{black} AutoARIMA & \color{black} Normal & \color{black} - & {\cellcolor[HTML]{D35D5F}} \color[HTML]{F1F1F1} \color{black} 11.277 & {\cellcolor[HTML]{D35D5F}} \color[HTML]{F1F1F1} \color{black} 15.755 & {\cellcolor[HTML]{CC7D7E}} \color[HTML]{F1F1F1} \color{black} 0.458 & {\cellcolor[HTML]{CA8586}} \color[HTML]{F1F1F1} \color{black} 0.695 & {\cellcolor[HTML]{D35D5F}} \color[HTML]{F1F1F1} \color{black} 71.669 & {\cellcolor[HTML]{D35D5F}} \color[HTML]{F1F1F1} \color{black} 57.638 & {\cellcolor[HTML]{D35D5F}} \color[HTML]{F1F1F1} \color{black} 8.467 & {\cellcolor[HTML]{D16567}} \color[HTML]{F1F1F1} \color{black} 4.311 & {\cellcolor[HTML]{C98C8D}} \color[HTML]{F1F1F1} \color{black} 1980.6 \\
\color{black} AutoARIMAX & \color{black} Normal & \color{black} - & {\cellcolor[HTML]{D35D5F}} \color[HTML]{F1F1F1} \color{black} 10.012 & {\cellcolor[HTML]{D35D5F}} \color[HTML]{F1F1F1} \color{black} 14.376 & {\cellcolor[HTML]{D16566}} \color[HTML]{F1F1F1} \color{black} 0.269 & {\cellcolor[HTML]{D35D5F}} \color[HTML]{F1F1F1} \color{black} 0.482 & {\cellcolor[HTML]{D35D5F}} \color[HTML]{F1F1F1} \color{black} 76.575 & {\cellcolor[HTML]{D35D5F}} \color[HTML]{F1F1F1} \color{black} 62.946 & {\cellcolor[HTML]{D35D5F}} \color[HTML]{F1F1F1} \color{black} 8.088 & {\cellcolor[HTML]{D35D5F}} \color[HTML]{F1F1F1} \color{black} 6.029 & {\cellcolor[HTML]{D35D5F}} \color[HTML]{F1F1F1} \color{black} 36832.6 \\
\midrule
\color{black} \citeauthor{brusaferri2024line}, \cite{brusaferri2024line} & \color{black} Normal & \color{black} DDNN-OCQ & {\cellcolor[HTML]{72ADDB}} \color[HTML]{F1F1F1} \color{black} 3.745 & {\cellcolor[HTML]{FFFFFF}} \color[HTML]{000000} \color{black}  & {\cellcolor[HTML]{FFFFFF}} \color[HTML]{000000} \color{black}  & {\cellcolor[HTML]{5AA8E4}} \color[HTML]{F1F1F1} \color{black} 0.808 & {\cellcolor[HTML]{FFFFFF}} \color[HTML]{000000} \color{black}  & {\cellcolor[HTML]{7BAFD7}} \color[HTML]{F1F1F1} \color{black} 17.976 & {\cellcolor[HTML]{9AB6CC}} \color[HTML]{000000} \color{black} 2.959 & {\cellcolor[HTML]{FFFFFF}} \color[HTML]{000000} \color{black}  & {\cellcolor[HTML]{C98C8D}} \color[HTML]{F1F1F1} \color{black} 1939.0 \\
\color{black} \citeauthor{brusaferri2024line}, \cite{brusaferri2024line} & \color{black} T & \color{black} DDNN-OCQ & {\cellcolor[HTML]{6DACDD}} \color[HTML]{F1F1F1} \color{black} 3.704 & {\cellcolor[HTML]{FFFFFF}} \color[HTML]{000000} \color{black}  & {\cellcolor[HTML]{FFFFFF}} \color[HTML]{000000} \color{black}  & {\cellcolor[HTML]{5AA8E4}} \color[HTML]{F1F1F1} \color{black} 0.803 & {\cellcolor[HTML]{FFFFFF}} \color[HTML]{000000} \color{black}  & {\cellcolor[HTML]{6CACDD}} \color[HTML]{F1F1F1} \color{black} 17.422 & {\cellcolor[HTML]{92B4CF}} \color[HTML]{000000} \color{black} 2.913 & {\cellcolor[HTML]{FFFFFF}} \color[HTML]{000000} \color{black}  & {\cellcolor[HTML]{C98C8D}} \color[HTML]{F1F1F1} \color{black} 1939.0 \\
\color{black} \citeauthor{brusaferri2024line}, \cite{brusaferri2024line} & \color{black} JSU & \color{black} DDNN-OCQ & {\cellcolor[HTML]{6CACDD}} \color[HTML]{F1F1F1} \color{black} 3.700 & {\cellcolor[HTML]{FFFFFF}} \color[HTML]{000000} \color{black}  & {\cellcolor[HTML]{FFFFFF}} \color[HTML]{000000} \color{black}  & {\cellcolor[HTML]{5AA8E4}} \color[HTML]{F1F1F1} \bfseries \color{black} 0.799 & {\cellcolor[HTML]{FFFFFF}} \color[HTML]{000000} \color{black}  & {\cellcolor[HTML]{65AAE0}} \color[HTML]{F1F1F1} \color{black} 17.178 & {\cellcolor[HTML]{90B4CF}} \color[HTML]{000000} \color{black} 2.899 & {\cellcolor[HTML]{FFFFFF}} \color[HTML]{000000} \color{black}  & {\cellcolor[HTML]{C98C8D}} \color[HTML]{F1F1F1} \color{black} 1939.0 \\
\color{black} \citeauthor{brusaferri2024line}, \cite{brusaferri2024line} & \color{black} - & \color{black} DQNN-OCQ & {\cellcolor[HTML]{6DACDD}} \color[HTML]{F1F1F1} \color{black} 3.704 & {\cellcolor[HTML]{FFFFFF}} \color[HTML]{000000} \color{black}  & {\cellcolor[HTML]{FFFFFF}} \color[HTML]{000000} \color{black}  & {\cellcolor[HTML]{5AA8E4}} \color[HTML]{F1F1F1} \color{black} 0.798 & {\cellcolor[HTML]{FFFFFF}} \color[HTML]{000000} \color{black}  & {\cellcolor[HTML]{6DACDD}} \color[HTML]{F1F1F1} \color{black} 17.450 & {\cellcolor[HTML]{93B5CE}} \color[HTML]{000000} \color{black} 2.918 & {\cellcolor[HTML]{FFFFFF}} \color[HTML]{000000} \color{black}  & {\cellcolor[HTML]{C98C8D}} \color[HTML]{F1F1F1} \color{black} 1939.0 \\
\midrule
\color{black} \citeauthor{marcjasz2023distributional}, \cite{marcjasz2023distributional} & \color{black} Normal & \color{black} DDNN-qEns & {\cellcolor[HTML]{5AA8E4}} \color[HTML]{F1F1F1} \bfseries \color{black} 3.564 & {\cellcolor[HTML]{FFFFFF}} \color[HTML]{000000} \color{black}  & {\cellcolor[HTML]{ADBAC4}} \color[HTML]{000000} \color{black} 0.542 & {\cellcolor[HTML]{C2AEAE}} \color[HTML]{000000} \color{black} 0.845 & {\cellcolor[HTML]{5AA8E4}} \color[HTML]{F1F1F1} \bfseries \color{black} 11.556 & {\cellcolor[HTML]{5AA8E4}} \color[HTML]{F1F1F1} \bfseries \color{black} 16.797 & {\cellcolor[HTML]{5AA8E4}} \color[HTML]{F1F1F1} \bfseries \color{black} 2.598 & {\cellcolor[HTML]{FFFFFF}} \color[HTML]{000000} \color{black}  & {\cellcolor[HTML]{C0B3B3}} \color[HTML]{000000} \color{black} 180.0 \\
\color{black} \citeauthor{marcjasz2023distributional}, \cite{marcjasz2023distributional} & \color{black} JSU & \color{black} DDNN-qEns & {\cellcolor[HTML]{68ABDF}} \color[HTML]{F1F1F1} \color{black} 3.670 & {\cellcolor[HTML]{FFFFFF}} \color[HTML]{000000} \color{black}  & {\cellcolor[HTML]{B6BCC1}} \color[HTML]{000000} \color{black} 0.543 & {\cellcolor[HTML]{B6BCC1}} \color[HTML]{000000} \color{black} 0.833 & {\cellcolor[HTML]{6CACDD}} \color[HTML]{F1F1F1} \color{black} 11.999 & {\cellcolor[HTML]{71ADDB}} \color[HTML]{F1F1F1} \color{black} 17.594 & {\cellcolor[HTML]{6CACDD}} \color[HTML]{F1F1F1} \color{black} 2.695 & {\cellcolor[HTML]{FFFFFF}} \color[HTML]{000000} \color{black}  & {\cellcolor[HTML]{C0B3B3}} \color[HTML]{000000} \color{black} 180.0 \\
\bottomrule
\end{tabular}

%% file: tables/electricity_price/02_epf_ablation_method.tex
\begin{tabular}{llrrrrrrrrr}
\toprule
Distribution & Method & MAE & RMSE & COV50 & COV80 & IS50 & IS80 & CRPS & LS & Time (min) \\
\midrule
\color{black} Normal & \color{black} LASSO & {\cellcolor[HTML]{62AAE1}} \color[HTML]{F1F1F1} \color{black} 3.941 & {\cellcolor[HTML]{5FA9E2}} \color[HTML]{F1F1F1} \color{black} 6.145 & {\cellcolor[HTML]{6AACDE}} \color[HTML]{F1F1F1} \color{black} 0.497 & {\cellcolor[HTML]{7BAFD7}} \color[HTML]{F1F1F1} \color{black} 0.788 & {\cellcolor[HTML]{5AA8E4}} \color[HTML]{F1F1F1} \bfseries \color{black} 21.822 & {\cellcolor[HTML]{5CA8E3}} \color[HTML]{F1F1F1} \color{black} 18.842 & {\cellcolor[HTML]{61AAE1}} \color[HTML]{F1F1F1} \color{black} 2.898 & {\cellcolor[HTML]{73AEDA}} \color[HTML]{F1F1F1} \color{black} 2.981 & {\cellcolor[HTML]{ACBAC5}} \color[HTML]{000000} \color{black} 1.4 \\
\color{black} T & \color{black} LASSO & {\cellcolor[HTML]{5FA9E2}} \color[HTML]{F1F1F1} \color{black} 3.918 & {\cellcolor[HTML]{5CA8E3}} \color[HTML]{F1F1F1} \color{black} 6.109 & {\cellcolor[HTML]{8CB3D1}} \color[HTML]{000000} \color{black} 0.473 & {\cellcolor[HTML]{94B5CE}} \color[HTML]{000000} \color{black} 0.776 & {\cellcolor[HTML]{65AAE0}} \color[HTML]{F1F1F1} \color{black} 22.339 & {\cellcolor[HTML]{5EA9E3}} \color[HTML]{F1F1F1} \color{black} 18.923 & {\cellcolor[HTML]{60A9E2}} \color[HTML]{F1F1F1} \color{black} 2.893 & {\cellcolor[HTML]{5AA8E4}} \color[HTML]{F1F1F1} \bfseries \color{black} 2.953 & {\cellcolor[HTML]{C98A8B}} \color[HTML]{F1F1F1} \color{black} 3.1 \\
\color{black} JSU & \color{black} LASSO & {\cellcolor[HTML]{5AA8E4}} \color[HTML]{F1F1F1} \bfseries \color{black} 3.877 & {\cellcolor[HTML]{D26466}} \color[HTML]{F1F1F1} \color{black} 8.985 & {\cellcolor[HTML]{D06D6F}} \color[HTML]{F1F1F1} \color{black} 0.419 & {\cellcolor[HTML]{D16566}} \color[HTML]{F1F1F1} \color{black} 0.716 & {\cellcolor[HTML]{82B1D5}} \color[HTML]{000000} \color{black} 23.673 & {\cellcolor[HTML]{5AA8E4}} \color[HTML]{F1F1F1} \bfseries \color{black} 18.764 & {\cellcolor[HTML]{5AA8E4}} \color[HTML]{F1F1F1} \bfseries \color{black} 2.855 & {\cellcolor[HTML]{6AABDE}} \color[HTML]{F1F1F1} \color{black} 2.971 & {\cellcolor[HTML]{D35D5F}} \color[HTML]{F1F1F1} \color{black} 5.4 \\
\midrule
\color{black} Normal & \color{black} OLS & {\cellcolor[HTML]{68ABDF}} \color[HTML]{F1F1F1} \color{black} 3.990 & {\cellcolor[HTML]{5DA9E3}} \color[HTML]{F1F1F1} \color{black} 6.125 & {\cellcolor[HTML]{73AEDA}} \color[HTML]{F1F1F1} \bfseries \color{black} 0.498 & {\cellcolor[HTML]{94B5CE}} \color[HTML]{000000} \color{black} 0.785 & {\cellcolor[HTML]{5CA9E3}} \color[HTML]{F1F1F1} \color{black} 21.937 & {\cellcolor[HTML]{70ADDC}} \color[HTML]{F1F1F1} \color{black} 19.618 & {\cellcolor[HTML]{6EACDD}} \color[HTML]{F1F1F1} \color{black} 2.972 & {\cellcolor[HTML]{8BB3D1}} \color[HTML]{000000} \color{black} 3.008 & {\cellcolor[HTML]{5AA8E4}} \color[HTML]{F1F1F1} \color{black} 0.5 \\
\color{black} T & \color{black} OLS & {\cellcolor[HTML]{64AAE0}} \color[HTML]{F1F1F1} \color{black} 3.959 & {\cellcolor[HTML]{5AA8E4}} \color[HTML]{F1F1F1} \bfseries \color{black} 6.081 & {\cellcolor[HTML]{73AEDA}} \color[HTML]{F1F1F1} \color{black} 0.486 & {\cellcolor[HTML]{8CB3D1}} \color[HTML]{000000} \color{black} 0.791 & {\cellcolor[HTML]{61AAE1}} \color[HTML]{F1F1F1} \color{black} 22.141 & {\cellcolor[HTML]{65AAE0}} \color[HTML]{F1F1F1} \color{black} 19.206 & {\cellcolor[HTML]{65AAE0}} \color[HTML]{F1F1F1} \color{black} 2.922 & {\cellcolor[HTML]{87B2D3}} \color[HTML]{000000} \color{black} 3.003 & {\cellcolor[HTML]{83B1D5}} \color[HTML]{000000} \color{black} 0.8 \\
\color{black} JSU & \color{black} OLS & {\cellcolor[HTML]{D35D5F}} \color[HTML]{F1F1F1} \color{black} $>10^4$ & {\cellcolor[HTML]{D35D5F}} \color[HTML]{F1F1F1} \color{black} $>10^4$ & {\cellcolor[HTML]{D35D5F}} \color[HTML]{F1F1F1} \color{black} 0.396 & {\cellcolor[HTML]{D35D5F}} \color[HTML]{F1F1F1} \color{black} 0.674 & {\cellcolor[HTML]{D35D5F}} \color[HTML]{F1F1F1} \color{black} $>10^4$ & {\cellcolor[HTML]{D35D5F}} \color[HTML]{F1F1F1} \color{black} $>10^4$ & {\cellcolor[HTML]{D35D5F}} \color[HTML]{F1F1F1} \color{black} $>10^4$ & {\cellcolor[HTML]{D35D5F}} \color[HTML]{F1F1F1} \color{black} 3.184 & {\cellcolor[HTML]{95B5CE}} \color[HTML]{000000} \color{black} 1.0 \\
\midrule
\color{black} Normal & \color{black} Elastic Net & {\cellcolor[HTML]{93B5CE}} \color[HTML]{000000} \color{black} 4.354 & {\cellcolor[HTML]{C4A2A3}} \color[HTML]{F1F1F1} \color{black} 7.897 & {\cellcolor[HTML]{73AEDA}} \color[HTML]{F1F1F1} \color{black} 0.509 & {\cellcolor[HTML]{7BAFD7}} \color[HTML]{F1F1F1} \color{black} 0.793 & {\cellcolor[HTML]{79AFD8}} \color[HTML]{F1F1F1} \color{black} 23.233 & {\cellcolor[HTML]{9AB6CC}} \color[HTML]{000000} \color{black} 21.333 & {\cellcolor[HTML]{95B5CE}} \color[HTML]{000000} \color{black} 3.216 & {\cellcolor[HTML]{87B2D3}} \color[HTML]{000000} \color{black} 3.004 & {\cellcolor[HTML]{B3BCC2}} \color[HTML]{000000} \color{black} 1.5 \\
\color{black} T & \color{black} Elastic Net & {\cellcolor[HTML]{8EB3D0}} \color[HTML]{000000} \color{black} 4.310 & {\cellcolor[HTML]{C3A7A7}} \color[HTML]{000000} \color{black} 7.815 & {\cellcolor[HTML]{84B1D4}} \color[HTML]{000000} \color{black} 0.486 & {\cellcolor[HTML]{94B5CE}} \color[HTML]{000000} \color{black} 0.781 & {\cellcolor[HTML]{7EB0D6}} \color[HTML]{F1F1F1} \color{black} 23.495 & {\cellcolor[HTML]{91B4CF}} \color[HTML]{000000} \color{black} 20.991 & {\cellcolor[HTML]{8DB3D1}} \color[HTML]{000000} \color{black} 3.169 & {\cellcolor[HTML]{6DACDD}} \color[HTML]{F1F1F1} \color{black} 2.974 & {\cellcolor[HTML]{C9898A}} \color[HTML]{F1F1F1} \color{black} 3.1 \\
\color{black} JSU & \color{black} Elastic Net & {\cellcolor[HTML]{8BB3D1}} \color[HTML]{000000} \color{black} 4.284 & {\cellcolor[HTML]{CE7576}} \color[HTML]{F1F1F1} \color{black} 8.677 & {\cellcolor[HTML]{D06D6F}} \color[HTML]{F1F1F1} \color{black} 0.426 & {\cellcolor[HTML]{D35D5F}} \color[HTML]{F1F1F1} \color{black} 0.712 & {\cellcolor[HTML]{B9BDC0}} \color[HTML]{000000} \color{black} 26.449 & {\cellcolor[HTML]{92B4CF}} \color[HTML]{000000} \color{black} 21.057 & {\cellcolor[HTML]{8EB3D0}} \color[HTML]{000000} \color{black} 3.171 & {\cellcolor[HTML]{92B4CF}} \color[HTML]{000000} \color{black} 3.016 & {\cellcolor[HTML]{D35D5F}} \color[HTML]{F1F1F1} \color{black} 5.4 \\
\midrule
\color{black} Normal & \color{black} OCD & {\cellcolor[HTML]{6DACDD}} \color[HTML]{F1F1F1} \color{black} 4.028 & {\cellcolor[HTML]{60A9E2}} \color[HTML]{F1F1F1} \color{black} 6.164 & {\cellcolor[HTML]{6AACDE}} \color[HTML]{F1F1F1} \color{black} 0.504 & {\cellcolor[HTML]{7BAFD7}} \color[HTML]{F1F1F1} \color{black} 0.791 & {\cellcolor[HTML]{5EA9E3}} \color[HTML]{F1F1F1} \color{black} 22.025 & {\cellcolor[HTML]{74AEDA}} \color[HTML]{F1F1F1} \color{black} 19.789 & {\cellcolor[HTML]{72ADDB}} \color[HTML]{F1F1F1} \color{black} 3.000 & {\cellcolor[HTML]{90B4CF}} \color[HTML]{000000} \color{black} 3.014 & {\cellcolor[HTML]{61AAE1}} \color[HTML]{F1F1F1} \color{black} 0.6 \\
\color{black} T & \color{black} OCD & {\cellcolor[HTML]{69ABDE}} \color[HTML]{F1F1F1} \color{black} 3.997 & {\cellcolor[HTML]{5CA9E3}} \color[HTML]{F1F1F1} \color{black} 6.116 & {\cellcolor[HTML]{84B1D4}} \color[HTML]{000000} \color{black} 0.491 & {\cellcolor[HTML]{94B5CE}} \color[HTML]{000000} \bfseries \color{black} 0.796 & {\cellcolor[HTML]{63AAE1}} \color[HTML]{F1F1F1} \color{black} 22.205 & {\cellcolor[HTML]{69ABDE}} \color[HTML]{F1F1F1} \color{black} 19.366 & {\cellcolor[HTML]{6AABDE}} \color[HTML]{F1F1F1} \color{black} 2.948 & {\cellcolor[HTML]{8EB3D0}} \color[HTML]{000000} \color{black} 3.011 & {\cellcolor[HTML]{82B1D5}} \color[HTML]{000000} \color{black} 0.8 \\
\color{black} JSU & \color{black} OCD & {\cellcolor[HTML]{D35D5F}} \color[HTML]{F1F1F1} \color{black} $>10^4$ & {\cellcolor[HTML]{D35D5F}} \color[HTML]{F1F1F1} \color{black} $>10^4$ & {\cellcolor[HTML]{D16566}} \color[HTML]{F1F1F1} \color{black} 0.403 & {\cellcolor[HTML]{D35D5F}} \color[HTML]{F1F1F1} \color{black} 0.680 & {\cellcolor[HTML]{D35D5F}} \color[HTML]{F1F1F1} \color{black} $>10^4$ & {\cellcolor[HTML]{D35D5F}} \color[HTML]{F1F1F1} \color{black} $>10^4$ & {\cellcolor[HTML]{D35D5F}} \color[HTML]{F1F1F1} \color{black} $>10^4$ & {\cellcolor[HTML]{D26062}} \color[HTML]{F1F1F1} \color{black} 3.180 & {\cellcolor[HTML]{94B5CE}} \color[HTML]{000000} \color{black} 1.0 \\
\bottomrule
\end{tabular}

%% file: tables/electricity_price/03_epf_ablation_forget.tex
\begin{tabular}{llrrrrrrrrr}
\toprule
Distribution & Forget & MAE & RMSE & COV50 & COV80 & IS50 & IS80 & CRPS & LS & Time (min) \\
\midrule
\color{black} JSU & \color{black} 0 & {\cellcolor[HTML]{B3BCC2}} \color[HTML]{000000} \color{black} 3.877 & {\cellcolor[HTML]{D35D5F}} \color[HTML]{F1F1F1} \color{black} 8.985 & {\cellcolor[HTML]{D06D6F}} \color[HTML]{F1F1F1} \color{black} 0.419 & {\cellcolor[HTML]{D16566}} \color[HTML]{F1F1F1} \color{black} 0.716 & {\cellcolor[HTML]{CA8788}} \color[HTML]{F1F1F1} \color{black} 23.673 & {\cellcolor[HTML]{C1B1B1}} \color[HTML]{000000} \color{black} 18.764 & {\cellcolor[HTML]{AABAC6}} \color[HTML]{000000} \color{black} 2.855 & {\cellcolor[HTML]{A6B9C7}} \color[HTML]{000000} \color{black} 2.971 & {\cellcolor[HTML]{C2AAAA}} \color[HTML]{000000} \color{black} 5.4 \\
\color{black} Normal & \color{black} 0 & {\cellcolor[HTML]{D35D5F}} \color[HTML]{F1F1F1} \color{black} 3.941 & {\cellcolor[HTML]{67ABDF}} \color[HTML]{F1F1F1} \color{black} 6.145 & {\cellcolor[HTML]{6AACDE}} \color[HTML]{F1F1F1} \bfseries \color{black} 0.497 & {\cellcolor[HTML]{7BAFD7}} \color[HTML]{F1F1F1} \bfseries \color{black} 0.788 & {\cellcolor[HTML]{5AA8E4}} \color[HTML]{F1F1F1} \bfseries \color{black} 21.822 & {\cellcolor[HTML]{C79596}} \color[HTML]{F1F1F1} \color{black} 18.842 & {\cellcolor[HTML]{CE7475}} \color[HTML]{F1F1F1} \color{black} 2.898 & {\cellcolor[HTML]{BFB8B8}} \color[HTML]{000000} \color{black} 2.981 & {\cellcolor[HTML]{5AA8E4}} \color[HTML]{F1F1F1} \color{black} 1.4 \\
\color{black} T & \color{black} 0 & {\cellcolor[HTML]{CB8284}} \color[HTML]{F1F1F1} \color{black} 3.918 & {\cellcolor[HTML]{64AAE0}} \color[HTML]{F1F1F1} \color{black} 6.109 & {\cellcolor[HTML]{8CB3D1}} \color[HTML]{000000} \color{black} 0.473 & {\cellcolor[HTML]{94B5CE}} \color[HTML]{000000} \color{black} 0.776 & {\cellcolor[HTML]{87B2D3}} \color[HTML]{000000} \color{black} 22.339 & {\cellcolor[HTML]{CD787A}} \color[HTML]{F1F1F1} \color{black} 18.923 & {\cellcolor[HTML]{CC7E80}} \color[HTML]{F1F1F1} \color{black} 2.893 & {\cellcolor[HTML]{75AEDA}} \color[HTML]{F1F1F1} \color{black} 2.953 & {\cellcolor[HTML]{A3B8C8}} \color[HTML]{000000} \color{black} 3.1 \\
\midrule
\color{black} JSU & \color{black} 1/1460 & {\cellcolor[HTML]{91B4CF}} \color[HTML]{000000} \color{black} 3.857 & {\cellcolor[HTML]{79AFD8}} \color[HTML]{F1F1F1} \color{black} 6.369 & {\cellcolor[HTML]{C98D8E}} \color[HTML]{F1F1F1} \color{black} 0.437 & {\cellcolor[HTML]{CA8586}} \color[HTML]{F1F1F1} \color{black} 0.740 & {\cellcolor[HTML]{C89091}} \color[HTML]{F1F1F1} \color{black} 23.562 & {\cellcolor[HTML]{9AB6CC}} \color[HTML]{000000} \color{black} 18.632 & {\cellcolor[HTML]{84B1D4}} \color[HTML]{000000} \color{black} 2.839 & {\cellcolor[HTML]{6DACDD}} \color[HTML]{F1F1F1} \color{black} 2.950 & {\cellcolor[HTML]{C88F90}} \color[HTML]{F1F1F1} \color{black} 7.4 \\
\color{black} Normal & \color{black} 1/1460 & {\cellcolor[HTML]{CA8687}} \color[HTML]{F1F1F1} \color{black} 3.916 & {\cellcolor[HTML]{5FA9E2}} \color[HTML]{F1F1F1} \color{black} 6.059 & {\cellcolor[HTML]{73AEDA}} \color[HTML]{F1F1F1} \color{black} 0.485 & {\cellcolor[HTML]{94B5CE}} \color[HTML]{000000} \color{black} 0.773 & {\cellcolor[HTML]{92B4CF}} \color[HTML]{000000} \color{black} 22.473 & {\cellcolor[HTML]{D35D5F}} \color[HTML]{F1F1F1} \color{black} 19.002 & {\cellcolor[HTML]{D35D5F}} \color[HTML]{F1F1F1} \color{black} 2.908 & {\cellcolor[HTML]{B6BCC1}} \color[HTML]{000000} \color{black} 2.976 & {\cellcolor[HTML]{74AEDA}} \color[HTML]{F1F1F1} \color{black} 1.8 \\
\color{black} T & \color{black} 1/1460 & {\cellcolor[HTML]{BABDBF}} \color[HTML]{000000} \color{black} 3.881 & {\cellcolor[HTML]{5DA9E3}} \color[HTML]{F1F1F1} \color{black} 6.028 & {\cellcolor[HTML]{B6BCC1}} \color[HTML]{000000} \color{black} 0.461 & {\cellcolor[HTML]{BEBEBE}} \color[HTML]{000000} \color{black} 0.761 & {\cellcolor[HTML]{ABBAC5}} \color[HTML]{000000} \color{black} 22.758 & {\cellcolor[HTML]{C0B4B4}} \color[HTML]{000000} \color{black} 18.756 & {\cellcolor[HTML]{BFB8B8}} \color[HTML]{000000} \color{black} 2.867 & {\cellcolor[HTML]{6AABDE}} \color[HTML]{F1F1F1} \color{black} 2.948 & {\cellcolor[HTML]{B8BDC0}} \color[HTML]{000000} \color{black} 4.0 \\
\color{black} JSU & \color{black} 1/1095 & {\cellcolor[HTML]{9BB6CB}} \color[HTML]{000000} \color{black} 3.863 & {\cellcolor[HTML]{72ADDB}} \color[HTML]{F1F1F1} \color{black} 6.279 & {\cellcolor[HTML]{C79596}} \color[HTML]{F1F1F1} \color{black} 0.439 & {\cellcolor[HTML]{CC7D7E}} \color[HTML]{F1F1F1} \color{black} 0.742 & {\cellcolor[HTML]{C79394}} \color[HTML]{F1F1F1} \color{black} 23.522 & {\cellcolor[HTML]{9CB6CB}} \color[HTML]{000000} \color{black} 18.637 & {\cellcolor[HTML]{8BB3D2}} \color[HTML]{000000} \color{black} 2.842 & {\cellcolor[HTML]{6AACDE}} \color[HTML]{F1F1F1} \color{black} 2.949 & {\cellcolor[HTML]{C98D8E}} \color[HTML]{F1F1F1} \color{black} 7.5 \\
\color{black} Normal & \color{black} 1/1095 & {\cellcolor[HTML]{C89091}} \color[HTML]{F1F1F1} \color{black} 3.910 & {\cellcolor[HTML]{5CA8E3}} \color[HTML]{F1F1F1} \color{black} 6.011 & {\cellcolor[HTML]{8CB3D1}} \color[HTML]{000000} \color{black} 0.484 & {\cellcolor[HTML]{94B5CE}} \color[HTML]{000000} \color{black} 0.774 & {\cellcolor[HTML]{97B5CD}} \color[HTML]{000000} \color{black} 22.530 & {\cellcolor[HTML]{C69798}} \color[HTML]{F1F1F1} \color{black} 18.838 & {\cellcolor[HTML]{CB8485}} \color[HTML]{F1F1F1} \color{black} 2.890 & {\cellcolor[HTML]{C1B2B2}} \color[HTML]{000000} \color{black} 2.984 & {\cellcolor[HTML]{7DB0D7}} \color[HTML]{F1F1F1} \color{black} 2.0 \\
\color{black} T & \color{black} 1/1095 & {\cellcolor[HTML]{AABAC6}} \color[HTML]{000000} \color{black} 3.872 & {\cellcolor[HTML]{5CA9E3}} \color[HTML]{F1F1F1} \color{black} 6.019 & {\cellcolor[HTML]{B6BCC1}} \color[HTML]{000000} \color{black} 0.460 & {\cellcolor[HTML]{C0B6B6}} \color[HTML]{000000} \color{black} 0.761 & {\cellcolor[HTML]{B3BCC2}} \color[HTML]{000000} \color{black} 22.865 & {\cellcolor[HTML]{C4A2A3}} \color[HTML]{F1F1F1} \color{black} 18.807 & {\cellcolor[HTML]{BFB9B9}} \color[HTML]{000000} \color{black} 2.866 & {\cellcolor[HTML]{65AAE0}} \color[HTML]{F1F1F1} \color{black} 2.947 & {\cellcolor[HTML]{BEBCBC}} \color[HTML]{000000} \color{black} 4.3 \\
\color{black} JSU & \color{black} 1/730 & {\cellcolor[HTML]{88B2D2}} \color[HTML]{000000} \color{black} 3.852 & {\cellcolor[HTML]{64AAE0}} \color[HTML]{F1F1F1} \color{black} 6.115 & {\cellcolor[HTML]{C59E9E}} \color[HTML]{F1F1F1} \color{black} 0.442 & {\cellcolor[HTML]{C98D8E}} \color[HTML]{F1F1F1} \color{black} 0.744 & {\cellcolor[HTML]{C98D8E}} \color[HTML]{F1F1F1} \color{black} 23.601 & {\cellcolor[HTML]{BBBDBF}} \color[HTML]{000000} \color{black} 18.720 & {\cellcolor[HTML]{87B2D3}} \color[HTML]{000000} \color{black} 2.840 & {\cellcolor[HTML]{74AEDA}} \color[HTML]{F1F1F1} \color{black} 2.952 & {\cellcolor[HTML]{CC7E7F}} \color[HTML]{F1F1F1} \color{black} 9.1 \\
\color{black} Normal & \color{black} 1/730 & {\cellcolor[HTML]{C79595}} \color[HTML]{F1F1F1} \color{black} 3.907 & {\cellcolor[HTML]{5BA8E4}} \color[HTML]{F1F1F1} \color{black} 6.002 & {\cellcolor[HTML]{6AACDE}} \color[HTML]{F1F1F1} \color{black} 0.485 & {\cellcolor[HTML]{9DB7CB}} \color[HTML]{000000} \color{black} 0.772 & {\cellcolor[HTML]{94B5CE}} \color[HTML]{000000} \color{black} 22.493 & {\cellcolor[HTML]{67ABDF}} \color[HTML]{F1F1F1} \color{black} 18.494 & {\cellcolor[HTML]{ACBAC5}} \color[HTML]{000000} \color{black} 2.856 & {\cellcolor[HTML]{C0B4B4}} \color[HTML]{000000} \color{black} 2.983 & {\cellcolor[HTML]{88B2D2}} \color[HTML]{000000} \color{black} 2.3 \\
\color{black} T & \color{black} 1/730 & {\cellcolor[HTML]{94B5CE}} \color[HTML]{000000} \color{black} 3.859 & {\cellcolor[HTML]{5AA8E4}} \color[HTML]{F1F1F1} \bfseries \color{black} 5.984 & {\cellcolor[HTML]{BEBEBE}} \color[HTML]{000000} \color{black} 0.454 & {\cellcolor[HTML]{C98D8E}} \color[HTML]{F1F1F1} \color{black} 0.754 & {\cellcolor[HTML]{BFBBBB}} \color[HTML]{000000} \color{black} 23.028 & {\cellcolor[HTML]{5AA8E4}} \color[HTML]{F1F1F1} \bfseries \color{black} 18.459 & {\cellcolor[HTML]{72ADDB}} \color[HTML]{F1F1F1} \color{black} 2.831 & {\cellcolor[HTML]{5AA8E4}} \color[HTML]{F1F1F1} \bfseries \color{black} 2.942 & {\cellcolor[HTML]{C0B6B6}} \color[HTML]{000000} \color{black} 4.7 \\
\color{black} JSU & \color{black} 1/365 & {\cellcolor[HTML]{9FB7CA}} \color[HTML]{000000} \color{black} 3.865 & {\cellcolor[HTML]{66ABE0}} \color[HTML]{F1F1F1} \color{black} 6.129 & {\cellcolor[HTML]{D16566}} \color[HTML]{F1F1F1} \color{black} 0.422 & {\cellcolor[HTML]{D35D5F}} \color[HTML]{F1F1F1} \color{black} 0.729 & {\cellcolor[HTML]{D35D5F}} \color[HTML]{F1F1F1} \color{black} 24.218 & {\cellcolor[HTML]{D35F60}} \color[HTML]{F1F1F1} \color{black} 18.997 & {\cellcolor[HTML]{B3BBC2}} \color[HTML]{000000} \color{black} 2.859 & {\cellcolor[HTML]{8DB3D1}} \color[HTML]{000000} \color{black} 2.961 & {\cellcolor[HTML]{D35D5F}} \color[HTML]{F1F1F1} \color{black} 13.3 \\
\color{black} Normal & \color{black} 1/365 & {\cellcolor[HTML]{C3A5A6}} \color[HTML]{000000} \color{black} 3.898 & {\cellcolor[HTML]{5DA9E3}} \color[HTML]{F1F1F1} \color{black} 6.022 & {\cellcolor[HTML]{A5B9C7}} \color[HTML]{000000} \color{black} 0.460 & {\cellcolor[HTML]{D35D5F}} \color[HTML]{F1F1F1} \color{black} 0.745 & {\cellcolor[HTML]{C1B2B2}} \color[HTML]{000000} \color{black} 23.130 & {\cellcolor[HTML]{BCBEBF}} \color[HTML]{000000} \color{black} 18.723 & {\cellcolor[HTML]{BFBBBB}} \color[HTML]{000000} \color{black} 2.865 & {\cellcolor[HTML]{D35D5F}} \color[HTML]{F1F1F1} \color{black} 3.016 & {\cellcolor[HTML]{ABBAC5}} \color[HTML]{000000} \color{black} 3.4 \\
\color{black} T & \color{black} 1/365 & {\cellcolor[HTML]{5AA8E4}} \color[HTML]{F1F1F1} \bfseries \color{black} 3.826 & {\cellcolor[HTML]{5AA8E4}} \color[HTML]{F1F1F1} \color{black} 5.992 & {\cellcolor[HTML]{CE7576}} \color[HTML]{F1F1F1} \color{black} 0.441 & {\cellcolor[HTML]{D35D5F}} \color[HTML]{F1F1F1} \color{black} 0.737 & {\cellcolor[HTML]{C59D9E}} \color[HTML]{F1F1F1} \color{black} 23.404 & {\cellcolor[HTML]{91B4CF}} \color[HTML]{000000} \color{black} 18.607 & {\cellcolor[HTML]{5AA8E4}} \color[HTML]{F1F1F1} \bfseries \color{black} 2.821 & {\cellcolor[HTML]{65AAE0}} \color[HTML]{F1F1F1} \color{black} 2.947 & {\cellcolor[HTML]{C69899}} \color[HTML]{F1F1F1} \color{black} 6.6 \\
\bottomrule
\end{tabular}

%% file: tables/electricity_price/04_epf_ablation_batch.tex
\begin{tabular}{llrrrrrrrrr}
\toprule
Distribution & Batch Size & MAE & RMSE & COV50 & COV80 & IS50 & IS80 & CRPS & LS & Time (min) \\
\midrule
\color{black} Normal & \color{black} 1 & {\cellcolor[HTML]{C4A2A3}} \color[HTML]{F1F1F1} \color{black} 3.941 & {\cellcolor[HTML]{5CA9E3}} \color[HTML]{F1F1F1} \color{black} 6.145 & {\cellcolor[HTML]{6AACDE}} \color[HTML]{F1F1F1} \bfseries \color{black} 0.497 & {\cellcolor[HTML]{7BAFD7}} \color[HTML]{F1F1F1} \bfseries \color{black} 0.788 & {\cellcolor[HTML]{5AA8E4}} \color[HTML]{F1F1F1} \bfseries \color{black} 21.822 & {\cellcolor[HTML]{8BB3D2}} \color[HTML]{000000} \color{black} 18.842 & {\cellcolor[HTML]{C2ADAD}} \color[HTML]{000000} \color{black} 2.898 & {\cellcolor[HTML]{C0B3B3}} \color[HTML]{000000} \color{black} 2.981 & {\cellcolor[HTML]{9AB6CC}} \color[HTML]{000000} \color{black} 1.4 \\
\color{black} T & \color{black} 1 & {\cellcolor[HTML]{ADBAC4}} \color[HTML]{000000} \color{black} 3.918 & {\cellcolor[HTML]{5AA8E4}} \color[HTML]{F1F1F1} \bfseries \color{black} 6.109 & {\cellcolor[HTML]{8CB3D1}} \color[HTML]{000000} \color{black} 0.473 & {\cellcolor[HTML]{94B5CE}} \color[HTML]{000000} \color{black} 0.776 & {\cellcolor[HTML]{8AB3D2}} \color[HTML]{000000} \color{black} 22.339 & {\cellcolor[HTML]{BDBEBE}} \color[HTML]{000000} \color{black} 18.923 & {\cellcolor[HTML]{BFBABA}} \color[HTML]{000000} \color{black} 2.893 & {\cellcolor[HTML]{5AA8E4}} \color[HTML]{F1F1F1} \bfseries \color{black} 2.953 & {\cellcolor[HTML]{C89192}} \color[HTML]{F1F1F1} \color{black} 3.1 \\
\color{black} JSU & \color{black} 1 & {\cellcolor[HTML]{5AA8E4}} \color[HTML]{F1F1F1} \bfseries \color{black} 3.877 & {\cellcolor[HTML]{D35D5F}} \color[HTML]{F1F1F1} \color{black} 8.985 & {\cellcolor[HTML]{D06D6F}} \color[HTML]{F1F1F1} \color{black} 0.419 & {\cellcolor[HTML]{D16566}} \color[HTML]{F1F1F1} \color{black} 0.716 & {\cellcolor[HTML]{CC7C7E}} \color[HTML]{F1F1F1} \color{black} 23.673 & {\cellcolor[HTML]{5AA8E4}} \color[HTML]{F1F1F1} \bfseries \color{black} 18.764 & {\cellcolor[HTML]{5AA8E4}} \color[HTML]{F1F1F1} \bfseries \color{black} 2.855 & {\cellcolor[HTML]{A0B7C9}} \color[HTML]{000000} \color{black} 2.971 & {\cellcolor[HTML]{D35D5F}} \color[HTML]{F1F1F1} \color{black} 5.4 \\
\midrule
\color{black} Normal & \color{black} 7 & {\cellcolor[HTML]{D35D5F}} \color[HTML]{F1F1F1} \color{black} 3.977 & {\cellcolor[HTML]{5FA9E2}} \color[HTML]{F1F1F1} \color{black} 6.167 & {\cellcolor[HTML]{7BAFD7}} \color[HTML]{F1F1F1} \color{black} 0.491 & {\cellcolor[HTML]{8CB3D1}} \color[HTML]{000000} \color{black} 0.779 & {\cellcolor[HTML]{7AAFD8}} \color[HTML]{F1F1F1} \color{black} 22.172 & {\cellcolor[HTML]{D35D5F}} \color[HTML]{F1F1F1} \color{black} 19.086 & {\cellcolor[HTML]{D35D5F}} \color[HTML]{F1F1F1} \color{black} 2.928 & {\cellcolor[HTML]{D35D5F}} \color[HTML]{F1F1F1} \color{black} 3.004 & {\cellcolor[HTML]{5AA8E4}} \color[HTML]{F1F1F1} \color{black} 0.7 \\
\color{black} T & \color{black} 7 & {\cellcolor[HTML]{C88F90}} \color[HTML]{F1F1F1} \color{black} 3.951 & {\cellcolor[HTML]{5CA8E3}} \color[HTML]{F1F1F1} \color{black} 6.132 & {\cellcolor[HTML]{ADBAC4}} \color[HTML]{000000} \color{black} 0.467 & {\cellcolor[HTML]{A5B9C7}} \color[HTML]{000000} \color{black} 0.770 & {\cellcolor[HTML]{A4B8C8}} \color[HTML]{000000} \color{black} 22.621 & {\cellcolor[HTML]{B7BDC1}} \color[HTML]{000000} \color{black} 18.914 & {\cellcolor[HTML]{C69B9B}} \color[HTML]{F1F1F1} \color{black} 2.904 & {\cellcolor[HTML]{8FB4D0}} \color[HTML]{000000} \color{black} 2.966 & {\cellcolor[HTML]{AEBAC4}} \color[HTML]{000000} \color{black} 1.7 \\
\color{black} JSU & \color{black} 7 & {\cellcolor[HTML]{B9BDC0}} \color[HTML]{000000} \color{black} 3.924 & {\cellcolor[HTML]{C3A5A6}} \color[HTML]{000000} \color{black} 7.782 & {\cellcolor[HTML]{D06D6F}} \color[HTML]{F1F1F1} \color{black} 0.411 & {\cellcolor[HTML]{D35D5F}} \color[HTML]{F1F1F1} \color{black} 0.710 & {\cellcolor[HTML]{D35D5F}} \color[HTML]{F1F1F1} \color{black} 24.047 & {\cellcolor[HTML]{CF6F71}} \color[HTML]{F1F1F1} \color{black} 19.055 & {\cellcolor[HTML]{BEBDBD}} \color[HTML]{000000} \color{black} 2.892 & {\cellcolor[HTML]{CB8081}} \color[HTML]{F1F1F1} \color{black} 2.995 & {\cellcolor[HTML]{C59E9F}} \color[HTML]{F1F1F1} \color{black} 2.7 \\
\bottomrule
\end{tabular}

%% file: tables/electricity_price/05_epf_ablation_model_selection.tex
\begin{tabular}{llrrrrrrrrr}
\toprule
Distribution & Model Selection & MAE & RMSE & COV50 & COV80 & IS50 & IS80 & CRPS & LS & Time (min) \\
\midrule
\color{black} Normal & \color{black} Local RSS & {\cellcolor[HTML]{C98B8C}} \color[HTML]{F1F1F1} \color{black} 3.941 & {\cellcolor[HTML]{5CA9E3}} \color[HTML]{F1F1F1} \color{black} 6.145 & {\cellcolor[HTML]{6AACDE}} \color[HTML]{F1F1F1} \color{black} 0.497 & {\cellcolor[HTML]{7BAFD7}} \color[HTML]{F1F1F1} \color{black} 0.788 & {\cellcolor[HTML]{5AA8E4}} \color[HTML]{F1F1F1} \bfseries \color{black} 21.822 & {\cellcolor[HTML]{80B0D6}} \color[HTML]{F1F1F1} \color{black} 18.842 & {\cellcolor[HTML]{C3A8A9}} \color[HTML]{000000} \color{black} 2.898 & {\cellcolor[HTML]{D35D5F}} \color[HTML]{F1F1F1} \color{black} 2.981 & {\cellcolor[HTML]{5AA8E4}} \color[HTML]{F1F1F1} \color{black} 1.4 \\
\color{black} T & \color{black} Local RSS & {\cellcolor[HTML]{BDBEBE}} \color[HTML]{000000} \color{black} 3.918 & {\cellcolor[HTML]{5AA8E4}} \color[HTML]{F1F1F1} \bfseries \color{black} 6.109 & {\cellcolor[HTML]{8CB3D1}} \color[HTML]{000000} \color{black} 0.473 & {\cellcolor[HTML]{94B5CE}} \color[HTML]{000000} \color{black} 0.776 & {\cellcolor[HTML]{8AB3D2}} \color[HTML]{000000} \color{black} 22.339 & {\cellcolor[HTML]{A7B9C7}} \color[HTML]{000000} \color{black} 18.923 & {\cellcolor[HTML]{C0B5B5}} \color[HTML]{000000} \color{black} 2.893 & {\cellcolor[HTML]{72ADDB}} \color[HTML]{F1F1F1} \color{black} 2.953 & {\cellcolor[HTML]{8CB3D1}} \color[HTML]{000000} \color{black} 3.1 \\
\color{black} JSU & \color{black} Local RSS & {\cellcolor[HTML]{5AA8E4}} \color[HTML]{F1F1F1} \bfseries \color{black} 3.877 & {\cellcolor[HTML]{D35D5F}} \color[HTML]{F1F1F1} \color{black} 8.985 & {\cellcolor[HTML]{D06D6F}} \color[HTML]{F1F1F1} \color{black} 0.419 & {\cellcolor[HTML]{D16566}} \color[HTML]{F1F1F1} \color{black} 0.716 & {\cellcolor[HTML]{CC7E7F}} \color[HTML]{F1F1F1} \color{black} 23.673 & {\cellcolor[HTML]{5AA8E4}} \color[HTML]{F1F1F1} \bfseries \color{black} 18.764 & {\cellcolor[HTML]{5AA8E4}} \color[HTML]{F1F1F1} \bfseries \color{black} 2.855 & {\cellcolor[HTML]{C59C9D}} \color[HTML]{F1F1F1} \color{black} 2.971 & {\cellcolor[HTML]{ADBAC4}} \color[HTML]{000000} \color{black} 5.4 \\
\midrule
\color{black} Normal & \color{black} Global LL & {\cellcolor[HTML]{C69B9B}} \color[HTML]{F1F1F1} \color{black} 3.934 & {\cellcolor[HTML]{5FA9E2}} \color[HTML]{F1F1F1} \color{black} 6.175 & {\cellcolor[HTML]{6AACDE}} \color[HTML]{F1F1F1} \bfseries \color{black} 0.498 & {\cellcolor[HTML]{73AEDA}} \color[HTML]{F1F1F1} \bfseries \color{black} 0.789 & {\cellcolor[HTML]{5EA9E3}} \color[HTML]{F1F1F1} \color{black} 21.870 & {\cellcolor[HTML]{D35D5F}} \color[HTML]{F1F1F1} \color{black} 19.177 & {\cellcolor[HTML]{D35D5F}} \color[HTML]{F1F1F1} \color{black} 2.925 & {\cellcolor[HTML]{D1686A}} \color[HTML]{F1F1F1} \color{black} 2.979 & {\cellcolor[HTML]{C98C8D}} \color[HTML]{F1F1F1} \color{black} 16.7 \\
\color{black} T & \color{black} Global LL & {\cellcolor[HTML]{A1B8C9}} \color[HTML]{000000} \color{black} 3.907 & {\cellcolor[HTML]{5AA8E4}} \color[HTML]{F1F1F1} \color{black} 6.112 & {\cellcolor[HTML]{94B5CE}} \color[HTML]{000000} \color{black} 0.473 & {\cellcolor[HTML]{A5B9C7}} \color[HTML]{000000} \color{black} 0.774 & {\cellcolor[HTML]{87B2D3}} \color[HTML]{000000} \color{black} 22.307 & {\cellcolor[HTML]{68ABDF}} \color[HTML]{F1F1F1} \color{black} 18.794 & {\cellcolor[HTML]{9AB6CC}} \color[HTML]{000000} \color{black} 2.878 & {\cellcolor[HTML]{5AA8E4}} \color[HTML]{F1F1F1} \bfseries \color{black} 2.949 & {\cellcolor[HTML]{CF6F71}} \color[HTML]{F1F1F1} \color{black} 27.4 \\
\color{black} JSU & \color{black} Global LL & {\cellcolor[HTML]{D35D5F}} \color[HTML]{F1F1F1} \color{black} 3.962 & {\cellcolor[HTML]{C3A8A8}} \color[HTML]{000000} \color{black} 7.741 & {\cellcolor[HTML]{CE7576}} \color[HTML]{F1F1F1} \color{black} 0.424 & {\cellcolor[HTML]{D06D6F}} \color[HTML]{F1F1F1} \color{black} 0.722 & {\cellcolor[HTML]{D35D5F}} \color[HTML]{F1F1F1} \color{black} 24.075 & {\cellcolor[HTML]{D26264}} \color[HTML]{F1F1F1} \color{black} 19.165 & {\cellcolor[HTML]{CE7678}} \color[HTML]{F1F1F1} \color{black} 2.916 & {\cellcolor[HTML]{C1B0B0}} \color[HTML]{000000} \color{black} 2.967 & {\cellcolor[HTML]{D35D5F}} \color[HTML]{F1F1F1} \color{black} 37.4 \\
\bottomrule
\end{tabular}

%% file: references.bib
@article{nowotarski2018recent,
  title     = {Recent advances in electricity price forecasting: A review of probabilistic forecasting},
  author    = {Nowotarski, Jakub and Weron, Rafa{\l}},
  year      = {2018},
  month     = {1},
  journal   = {Renewable and Sustainable Energy Reviews},
  publisher = {Elsevier},
  volume    = {81},
  pages     = {1548--1568},
  doi       = {10.1016/j.rser.2017.05.234}
}

@article{lago2018forecasting,
  title     = {Forecasting day-ahead electricity prices in Europe: The importance of considering market integration},
  author    = {Lago, Jesus and De Ridder, Fjo and Vrancx, Peter and De Schutter, Bart},
  year      = {2018},
  month     = {2},
  journal   = {Applied energy},
  publisher = {Elsevier},
  volume    = {211},
  pages     = {890--903},
  doi       = {10.1016/j.apenergy.2017.11.098}
}

@article{pedregosa2011scikit,
  title={Scikit-learn: Machine learning in Python},
  author={Pedregosa, Fabian and Varoquaux, Ga{\"e}l and Gramfort, Alexandre and Michel, Vincent and Thirion, Bertrand and Grisel, Olivier and Blondel, Mathieu and Prettenhofer, Peter and Weiss, Ron and Dubourg, Vincent and others},
  journal={the Journal of machine Learning research},
  volume={12},
  pages={2825--2830},
  year={2011},
  publisher={JMLR. org}
}

@article{eilers1996flexible,
  title     = {Flexible smoothing with B-splines and penalties},
  author    = {Eilers, Paul HC and Marx, Brian D},
  year      = {1996},
  month     = {5},
  journal   = {Statistical science},
  publisher = {Institute of Mathematical Statistics},
  volume    = {11},
  number    = {2},
  pages     = {89--121},
  doi       = {10.1214/ss/1038425655}
}

@article{cox1987parameter,
  title     = {Parameter orthogonality and approximate conditional inference},
  author    = {Cox, David Roxbee and Reid, Nancy},
  year      = {1987},
  month     = {9},
  journal   = {Journal of the Royal Statistical Society: Series B (Methodological)},
  publisher = {Wiley Online Library},
  volume    = {49},
  number    = {1},
  pages     = {1--18}
}

@misc{garza2022statsforecast,
  title        = {{StatsForecast}: Lightning fast forecasting with statistical and econometric models},
  author       = {Azul Garza AND Max Mergenthaler Canseco AND Cristian Challú AND Kin G. Olivares},
  year         = {2022},
  howpublished = {{PyCon} Salt Lake City, Utah, US 2022}
}

@article{tay2023elastic,
  title     = {Elastic net regularization paths for all generalized linear models},
  author    = {Tay, J Kenneth and Narasimhan, Balasubramanian and Hastie, Trevor},
  year      = {2023},
  journal   = {Journal of statistical software},
  publisher = {NIH Public Access},
  volume    = {106},
  number    = {1},
  doi       = {10.18637/jss.v106.i01}
}

@article{ziel2018day,
  title     = {Day-ahead electricity price forecasting with high-dimensional structures: Univariate vs. multivariate modeling frameworks},
  author    = {Ziel, Florian and Weron, Rafa{\l}},
  year      = {2018},
  month     = {2},
  journal   = {Energy Economics},
  publisher = {Elsevier},
  volume    = {70},
  pages     = {396--420},
  doi       = {10.1016/j.eneco.2017.12.016}
}

@article{muschinski2022cholesky,
  title     = {Cholesky-based multivariate Gaussian regression},
  author    = {Muschinski, Thomas and Mayr, Georg J and Simon, Thorsten and Umlauf, Nikolaus and Zeileis, Achim},
  year      = {2022},
  month     = {3},
  journal   = {Econometrics and Statistics},
  publisher = {Elsevier},
  volume    = {29},
  pages     = {261--281},
  doi       = {10.1016/j.ecosta.2022.03.001}
}

@inproceedings{thielmann2024neural,
  title        = {Neural additive models for location scale and shape: A framework for interpretable neural regression beyond the mean},
  author       = {Thielmann, Anton Frederik and Kruse, Ren{\'e}-Marcel and Kneib, Thomas and S{\"a}fken, Benjamin},
  year         = {2024},
  booktitle    = {International Conference on Artificial Intelligence and Statistics},
  publisher    = {PMLR},
  volume       = {238},
  pages        = {1783--1791},
  editor       = {Dasgupta, Sanjoy and Mandt, Stephan and Li, Yingzhen},
  organization = {PMLR}
}

@article{umlauf2024scalable,
  title     = {Scalable estimation for structured additive distributional regression},
  author    = {Umlauf, Nikolaus and Seiler, Johannes and Wetscher, Mattias and Simon, Thorsten and Lang, Stefan and Klein, Nadja},
  year      = {2024},
  month     = {8},
  journal   = {Journal of Computational and Graphical Statistics},
  publisher = {Taylor \& Francis},
  pages     = {1--17}
}

@article{maerz2019xgboostlss,
  title     = {XGBoostLSS--An extension of XGBoost to probabilistic forecasting},
  author    = {M{\"a}rz, Alexander},
  year      = {2019},
  journal   = {arXiv preprint arXiv:1907.03178},
  publisher = {Cornell University},
  volume    = {abs/1907.03178},
  doi       = {10.48550/arxiv.1907.03178}
}

@article{cevid2022distributional,
  title   = {Distributional random forests: Heterogeneity adjustment and multivariate distributional regression},
  author  = {Cevid, Domagoj and Michel, Loris and N{\"a}f, Jeffrey and B{\"u}hlmann, Peter and Meinshausen, Nicolai},
  year    = {2022},
  journal = {Journal of Machine Learning Research},
  volume  = {23},
  number  = {333},
  pages   = {1--79}
}

@incollection{dette2013least,
  title     = {Least Squares Estimation in High Dimensional Sparse Heteroscedastic Models},
  author    = {Dette, Holger and Wagener, Jens},
  year      = {2013},
  booktitle = {Robustness and Complex Data Structures: Festschrift in Honour of Ursula Gather},
  publisher = {Springer},
  pages     = {135--147},
  doi       = {10.1007/978-3-642-35494-6_9},
  isbn      = {978-3642354939}
}

@article{friedman2010regularization,
  title     = {Regularization paths for generalized linear models via coordinate descent},
  author    = {Friedman, Jerome and Hastie, Trevor and Tibshirani, Robert},
  year      = {2010},
  journal   = {Journal of statistical software},
  publisher = {NIH Public Access},
  volume    = {33},
  number    = {1},
  pages     = {1},
  doi       = {10.18637/jss.v033.i01}
}

@article{friedman2007pathwise,
  title     = {Pathwise coordinate optimization},
  author    = {Friedman, Jerome and Hastie, Trevor and H{\"o}fling, Holger and Tibshirani, Robert},
  year      = {2007},
  month     = {12},
  journal   = {The Annals of Applied Statistics},
  publisher = {Institute of Mathematical Statistics},
  volume    = {1},
  number    = {2},
  doi       = {10.1214/07-aoas131}
}

@article{hirsch2025online,
  title={Online Multivariate Regularized Distributional Regression for High-dimensional Probabilistic Electricity Price Forecasting},
  author={Hirsch, Simon},
  journal={arXiv preprint arXiv:2504.02518},
  year={2025}
}

@article{messner2019online,
  title     = {Online adaptive lasso estimation in vector autoregressive models for high dimensional wind power forecasting},
  author    = {Messner, Jakob W and Pinson, Pierre},
  year      = {2019},
  month     = {10},
  journal   = {International Journal of Forecasting},
  publisher = {Elsevier},
  volume    = {35},
  number    = {4},
  pages     = {1485--1498},
  doi       = {10.1016/j.ijforecast.2018.02.001}
}

@article{ziel2016iteratively,
  title     = {Iteratively reweighted adaptive lasso for conditional heteroscedastic time series with applications to ar--arch type processes},
  author    = {Ziel, Florian},
  year      = {2016},
  month     = {8},
  journal   = {Computational Statistics \& Data Analysis},
  publisher = {Elsevier},
  volume    = {100},
  pages     = {773--793},
  doi       = {10.1016/j.csda.2015.11.016}
}

@article{angelosante2010online,
  title     = {Online adaptive estimation of sparse signals: Where RLS meets the $\ell_1$-norm},
  author    = {Angelosante, Daniele and Bazerque, Juan Andr{\'e}s and Giannakis, Georgios B},
  year      = {2010},
  month     = {7},
  journal   = {IEEE Transactions on signal Processing},
  publisher = {IEEE},
  volume    = {58},
  number    = {7},
  pages     = {3436--3447},
  doi       = {10.1109/tsp.2010.2046897}
}

@article{werge2022adavol,
  title     = {Adavol: An adaptive recursive volatility prediction method},
  author    = {Werge, Nicklas and Wintenberger, Olivier},
  year      = {2022},
  month     = {7},
  journal   = {Econometrics and Statistics},
  publisher = {Elsevier},
  volume    = {23},
  pages     = {19--35},
  doi       = {10.1016/j.ecosta.2021.01.004}
}

@article{hendrych2018self,
  title     = {Self-weighted recursive estimation of GARCH models},
  author    = {Hendrych, Radek and Cipra, Tom{\'a}{\v{s}}},
  year      = {2018},
  month     = {2},
  journal   = {Communications in Statistics-Simulation and Computation},
  publisher = {Taylor \& Francis},
  volume    = {47},
  number    = {2},
  pages     = {315--328}
}

@book{haykin2014adaptive,
  title     = {Adaptive filter theory},
  author    = {Haykin, Simon},
  year      = {2014},
  publisher = {Pearson},
  edition   = {5}
}

@article{kock2016consistent,
  title     = {Consistent and conservative model selection with the adaptive lasso in stationary and nonstationary autoregressions},
  author    = {Kock, Anders Bredahl},
  year      = {2016},
  month     = {2},
  journal   = {Econometric Theory},
  publisher = {Cambridge University Press},
  volume    = {32},
  number    = {1},
  pages     = {243--259},
  doi       = {10.1017/s0266466615000304}
}

@article{kim2012consistent,
  title     = {Consistent model selection criteria on high dimensions},
  author    = {Kim, Yongdai and Kwon, Sunghoon and Choi, Hosik},
  year      = {2012},
  month     = {3},
  journal   = {The Journal of Machine Learning Research},
  publisher = {JMLR. org},
  volume    = {13},
  number    = {1},
  pages     = {1037--1057}
}

@article{cipra2018robust,
  title     = {Robust recursive estimation of garch models},
  author    = {Cipra, Tom{\'a}{\v{s}} and Hendrych, Radek},
  year      = {2018},
  month     = {12},
  journal   = {Kybernetika},
  publisher = {Institute of Information Theory and Automation AS CR},
  volume    = {54},
  number    = {6},
  pages     = {1138--1155},
  doi       = {10.14736/kyb-2018-6-1138}
}

@article{shi2016primer,
  title     = {A primer on coordinate descent algorithms},
  author    = {Shi, Hao-Jun Michael and Tu, Shenyinying and Xu, Yangyang and Yin, Wotao},
  year      = {2016},
  month     = {9},
  journal   = {arXiv preprint arXiv:1610.00040},
  publisher = {Cornell University},
  doi       = {10.48550/arxiv.1610.00040}
}

@inproceedings{yang2023online,
  title        = {Online Linearized LASSO},
  author       = {Yang, Shuoguang and Yan, Yuhao and Zhu, Xiuneng and Sun, Qiang},
  year         = {2023},
  booktitle    = {International Conference on Artificial Intelligence and Statistics},
  publisher    = {PMLR},
  volume       = {206},
  pages        = {7594--7610},
  editor       = {Ruiz, Francisco J. R. and Dy, Jennifer G. and van de Meent, Jan-Willem},
  organization = {PMLR}
}

@inproceedings{yang2010online,
  title     = {Online learning for group lasso},
  author    = {Yang, Haiqin and Xu, Zenglin and King, Irwin and Lyu, Michael R},
  year      = {2010},
  month     = {6},
  journal   = {International Conference on Machine Learning},
  booktitle = {Proceedings of the 27th International Conference on Machine Learning (ICML-10)},
  publisher = {Omnipress},
  pages     = {1191--1198},
  editor    = {Fürnkranz, Johannes and Joachims, Thorsten}
}

@article{petropoulos2022forecasting,
  title     = {Forecasting: theory and practice},
  author    = {Petropoulos, Fotios and Apiletti, Daniele and Assimakopoulos, Vassilios and Babai, Mohamed Zied and Barrow, Devon K and Taieb, Souhaib Ben and Bergmeir, Christoph and Bessa, Ricardo J and Bijak, Jakub and Boylan, John E and others},
  year      = {2022},
  month     = {7},
  journal   = {International Journal of Forecasting},
  publisher = {Elsevier},
  volume    = {38},
  number    = {3},
  pages     = {705--871},
  doi       = {10.1016/j.ijforecast.2021.11.001}
}

@article{monti2018adaptive,
  title     = {Adaptive regularization for lasso models in the context of nonstationary data streams},
  author    = {Monti, Ricardo P and Anagnostopoulos, Christoforos and Montana, Giovanni},
  year      = {2018},
  month     = {7},
  journal   = {Statistical Analysis and Data Mining: The ASA Data Science Journal},
  publisher = {Wiley Online Library},
  volume    = {11},
  number    = {5},
  pages     = {237--247}
}

@article{ziel2022m5,
  title     = {M5 competition uncertainty: Overdispersion, distributional forecasting, GAMLSS, and beyond},
  author    = {Ziel, Florian},
  year      = {2022},
  month     = {10},
  journal   = {International Journal of Forecasting},
  publisher = {Elsevier},
  volume    = {38},
  number    = {4},
  pages     = {1546--1554},
  doi       = {10.1016/j.ijforecast.2021.09.008}
}

@article{stasinopoulos2018gamlss,
  title     = {GAMLSS: A distributional regression approach},
  author    = {Stasinopoulos, Mikis D and Rigby, Robert A and Bastiani, Fernanda De},
  year      = {2018},
  month     = {3},
  journal   = {Statistical Modelling},
  publisher = {SAGE Publications Sage India: New Delhi, India},
  volume    = {18},
  number    = {3-4},
  pages     = {248--273}
}

@article{rigby2005generalized,
  title     = {Generalized additive models for location, scale and shape},
  author    = {Rigby, Robert A and Stasinopoulos, Mikis D},
  year      = {2005},
  month     = {4},
  journal   = {Journal of the Royal Statistical Society Series C: Applied Statistics},
  publisher = {Oxford University Press},
  volume    = {54},
  number    = {3},
  pages     = {507--554}
}

@article{stasinopoulos2008generalized,
  title   = {Generalized additive models for location scale and shape (GAMLSS) in R},
  author  = {Stasinopoulos, Mikis D and Rigby, Robert A},
  year    = {2008},
  journal = {Journal of Statistical Software},
  volume  = {23},
  pages   = {1--46}
}

@article{hofner2014gamboostlss,
  title     = {gamboostLSS: An R package for model building and variable selection in the GAMLSS framework},
  author    = {Hofner, Benjamin and Mayr, Andreas and Schmid, Matthias},
  year      = {2014},
  month     = {7},
  journal   = {arXiv preprint arXiv:1407.1774},
  publisher = {Cornell University},
  doi       = {10.48550/arxiv.1407.1774}
}

@article{groll2019lasso,
  title     = {LASSO-type penalization in the framework of generalized additive models for location, scale and shape},
  author    = {Groll, Andreas and Hambuckers, Julien and Kneib, Thomas and Umlauf, Nikolaus},
  year      = {2019},
  month     = {12},
  journal   = {Computational Statistics \& Data Analysis},
  publisher = {Elsevier},
  volume    = {140},
  pages     = {59--73},
  doi       = {10.1016/j.csda.2019.06.005}
}

@article{cesa2021online,
  title     = {Online learning algorithms},
  author    = {Cesa-Bianchi, Nicol{\`o} and Orabona, Francesco},
  year      = {2021},
  month     = {3},
  journal   = {Annual review of statistics and its application},
  publisher = {Annual Reviews},
  volume    = {8},
  pages     = {165--190},
  doi       = {10.1146/annurev-statistics-040620-035329}
}

@misc{gneiting2008probabilistic,
  title     = {Probabilistic forecasting},
  author    = {Gneiting, Tilmann},
  year      = {2008},
  journal   = {Journal of the Royal Statistical Society Series A: Statistics in Society},
  publisher = {Oxford University Press},
  volume    = {171},
  number    = {2},
  pages     = {319--321}
}

@article{bracher2021evaluating,
  title     = {Evaluating epidemic forecasts in an interval format},
  author    = {Bracher, Johannes and Ray, Evan L and Gneiting, Tilmann and Reich, Nicholas G},
  year      = {2021},
  month     = {2},
  journal   = {PLoS computational biology},
  publisher = {Public Library of Science San Francisco, CA USA},
  volume    = {17},
  number    = {2},
  pages     = {e1008618},
  doi       = {10.1371/journal.pcbi.1008618}
}

@article{gneiting2011making,
  title     = {Making and evaluating point forecasts},
  author    = {Gneiting, Tilmann},
  year      = {2011},
  month     = {6},
  journal   = {Journal of the American Statistical Association},
  publisher = {Taylor \& Francis},
  volume    = {106},
  number    = {494},
  pages     = {746--762}
}

@article{gneiting2014probabilistic,
  title     = {Probabilistic forecasting},
  author    = {Gneiting, Tilmann and Katzfuss, Matthias},
  year      = {2014},
  month     = {1},
  journal   = {Annual Review of Statistics and Its Application},
  publisher = {Annual Reviews},
  volume    = {1},
  number    = {1},
  pages     = {125--151},
  doi       = {10.1146/annurev-statistics-062713-085831}
}

@article{priouret2005recursive,
  title     = {On recursive estimation for time varying autoregressive processes},
  author    = {Priouret, P and Moulines, E and Roueff, Fran{\c{c}}ois},
  year      = {2005},
  month     = {12},
  journal   = {Annals of Statistics},
  publisher = {Institute of Mathematical Statistics},
  volume    = {33},
  number    = {6},
  pages     = {2610--2654},
  doi       = {10.1214/009053605000000624}
}

@article{dahlhaus2007recursive,
  title     = {A recursive online algorithm for the estimation of time-varying ARCH parameters},
  author    = {Dahlhaus, Rainer and Subba Rao, Suhasini},
  year      = {2007},
  month     = {5},
  journal   = {Bernoulli},
  publisher = {Bernoulli Society for Mathematical Statistics and Probability},
  volume    = {13},
  number    = {2},
  pages     = {389–422},
  doi       = {10.3150/07-bej5009},
  issue     = {2}
}

@article{alvarez2021probabilistic,
  title     = {Probabilistic load forecasting based on adaptive online learning},
  author    = {{\'A}lvarez, Ver{\'o}nica and Mazuelas, Santiago and Lozano, Jos{\'e} A},
  year      = {2021},
  month     = {7},
  journal   = {IEEE Transactions on Power Systems},
  publisher = {IEEE},
  volume    = {36},
  number    = {4},
  pages     = {3668--3680},
  doi       = {10.1109/tpwrs.2021.3050837}
}

@article{mayr2012generalized,
  title     = {Generalized additive models for location, scale and shape for high dimensional data—a flexible approach based on boosting},
  author    = {Mayr, Andreas and Fenske, Nora and Hofner, Benjamin and Kneib, Thomas and Schmid, Matthias},
  year      = {2012},
  month     = {1},
  journal   = {Journal of the Royal Statistical Society Series C: Applied Statistics},
  publisher = {Oxford University Press},
  volume    = {61},
  number    = {3},
  pages     = {403--427}
}

@article{ziel2021gamlss,
  title     = {gamlss. lasso: Extra Lasso-Type Additive Terms for GAMLSS},
  author    = {Ziel, Florian and Muniain, P and Stasinopoulos, Mikis D},
  year      = {2021},
  month     = {4},
  journal   = {R package version},
  booktitle = {CRAN: Contributed Packages},
  publisher = {The R Foundation},
  pages     = {1--0},
  doi       = {10.32614/cran.package.gamlss.lasso}
}

@article{klein2024distributional,
  title     = {Distributional regression for data analysis},
  author    = {Klein, Nadja},
  year      = {2024},
  month     = {4},
  journal   = {Annual Review of Statistics and Its Application},
  publisher = {Annual Reviews},
  volume    = {11},
  pages     = {321--346},
  doi       = {10.1146/annurev-statistics-040722-053607}
}

@article{kneib2023rage,
  title     = {Rage against the mean--a review of distributional regression approaches},
  author    = {Kneib, Thomas and Silbersdorff, Alexander and S{\"a}fken, Benjamin},
  year      = {2023},
  month     = {4},
  journal   = {Econometrics and Statistics},
  publisher = {Elsevier},
  volume    = {26},
  pages     = {99--123},
  doi       = {10.1016/j.ecosta.2021.07.006}
}

@article{marcjasz2023distributional,
  title     = {Distributional neural networks for electricity price forecasting},
  author    = {Marcjasz, Grzegorz and Narajewski, Micha{\l} and Weron, Rafa{\l} and Ziel, Florian},
  year      = {2023},
  month     = {9},
  journal   = {Energy Economics},
  publisher = {Elsevier},
  volume    = {125},
  pages     = {106843},
  doi       = {10.1016/j.eneco.2023.106843}
}

@article{2020SciPy-NMeth,
  title     = {{{SciPy} 1.0: Fundamental Algorithms for Scientific Computing in Python}},
  author    = {Virtanen, Pauli and Gommers, Ralf and Oliphant, Travis E. and Haberland, Matt and Reddy, Tyler and Cournapeau, David and Burovski, Evgeni and Peterson, Pearu and Weckesser, Warren and Bright, Jonathan and {van der Walt}, St{\'e}fan J. and Brett, Matthew and Wilson, Joshua and Millman, K. Jarrod and Mayorov, Nikolay and Nelson, Andrew R. J. and Jones, Eric and Kern, Robert and Larson, Eric and Carey, C J and Polat, {\.I}lhan and Feng, Yu and Moore, Eric W. and {VanderPlas}, Jake and Laxalde, Denis and Perktold, Josef and Cimrman, Robert and Henriksen, Ian and Quintero, E. A. and Harris, Charles R. and Archibald, Anne M. and Ribeiro, Ant{\^o}nio H. and Pedregosa, Fabian and {van Mulbregt}, Paul and {SciPy 1.0 Contributors}},
  year      = {2020},
  month     = {2},
  journal   = {Nature Methods},
  publisher = {Nature Portfolio},
  volume    = {17},
  number    = {3},
  pages     = {261--272},
  doi       = {10.1038/s41592-019-0686-2},
  adsurl    = {https://rdcu.be/b08Wh}
}

@article{2020NumPy-Array,
  title     = {Array programming with {NumPy}},
  author    = {Harris, Charles R. and Millman, K. Jarrod and van der Walt, Stéfan J and Gommers, Ralf and Virtanen, Pauli and Cournapeau, David and Wieser, Eric and Taylor, Julian and Berg, Sebastian and Smith, Nathaniel J. and Kern, Robert and Picus, Matti and Hoyer, Stephan and van Kerkwijk, Marten H. and Brett, Matthew and Haldane, Allan and Fernández del Río, Jaime and Wiebe, Mark and Peterson, Pearu and Gérard-Marchant, Pierre and Sheppard, Kevin and Reddy, Tyler and Weckesser, Warren and Abbasi, Hameer and Gohlke, Christoph and Oliphant, Travis E.},
  year      = {2020},
  month     = {9},
  journal   = {Nature},
  publisher = {Springer Science and Business Media LLC},
  volume    = {585},
  number    = {7825},
  pages     = {357–362},
  doi       = {10.1038/s41586-020-2649-2}
}

@inproceedings{lam2015numba,
  title     = {Numba: A llvm-based python jit compiler},
  author    = {Lam, Siu Kwan and Pitrou, Antoine and Seibert, Stanley},
  year      = {2015},
  month     = {11},
  journal   = {LLVM '15},
  booktitle = {Proceedings of the Second Workshop on the LLVM Compiler Infrastructure in HPC},
  publisher = {ACM},
  pages     = {1--6},
  editor    = {Finkel, Hal}
}

@software{zanetta_scoringrules_2024,
  title  = {Scoringrules: a python library for probabilistic forecast evaluation},
  author = {Francesco Zanetta and Sam Allen},
  year   = {2024}
}

@misc{hong2016probabilistic,
  title     = {Probabilistic energy forecasting: Global energy forecasting competition 2014 and beyond},
  author    = {Hong, Tao and Pinson, Pierre and Fan, Shu and Zareipour, Hamidreza and Troccoli, Alberto and Hyndman, Rob J},
  year      = {2016},
  month     = {7},
  journal   = {International Journal of forecasting},
  publisher = {Elsevier},
  volume    = {32},
  number    = {3},
  pages     = {896--913},
  doi       = {10.1016/j.ijforecast.2016.02.001}
}

@article{diebold2015comparing,
  title     = {Comparing predictive accuracy, twenty years later: A personal perspective on the use and abuse of Diebold--Mariano tests},
  author    = {Diebold, Francis X},
  year      = {2015},
  month     = {1},
  journal   = {Journal of Business \& Economic Statistics},
  publisher = {Taylor \& Francis},
  volume    = {33},
  number    = {1},
  pages     = {1--1}
}

@article{diebold2002comparing,
  title     = {Comparing predictive accuracy},
  author    = {Diebold, Francis X and Mariano, Robert S},
  year      = {1995},
  month     = {7},
  journal   = {Journal of Business \& economic statistics},
  publisher = {Taylor \& Francis},
  volume    = {13},
  number    = {3},
  pages     = {253--263},
  doi       = {10.1198/073500102753410444}
}

@article{angelopoulos2023conformal,
  title     = {Conformal pid control for time series prediction},
  author    = {Angelopoulos, Anastasios and Candes, Emmanuel and Tibshirani, Ryan J},
  year      = {2023},
  month     = {7},
  journal   = {Advances in neural information processing systems},
  booktitle = {Advances in Neural Information Processing Systems 36: Annual Conference on Neural Information Processing Systems 2023, NeurIPS 2023, New Orleans, LA, USA, December 10 - 16, 2023},
  publisher = {Cornell University},
  volume    = {36},
  pages     = {23047--23074},
  doi       = {10.48550/arxiv.2307.16895},
  editor    = {Oh, Alice and Naumann, Tristan and Globerson, Amir and Saenko, Kate and Hardt, Moritz and Levine, Sergey}
}

@book{stasinopoulos2024generalized,
  title     = {Generalized Additive Models for Location, Scale and Shape: A Distributional Regression Approach, with Applications},
  author    = {Stasinopoulos, Mikis D and Kneib, Thomas and Klein, Nadja and Mayr, Andreas and Heller, Gillian Z},
  year      = {2024},
  publisher = {Cambridge University Press},
  volume    = {56}
}

@article{viehmann2017state,
  title     = {State of the German short-term power market},
  author    = {Viehmann, Johannes},
  year      = {2017},
  month     = {5},
  journal   = {Zeitschrift f{\"u}r Energiewirtschaft},
  publisher = {Springer},
  volume    = {41},
  number    = {2},
  pages     = {87--103},
  doi       = {10.1007/s12398-017-0196-9}
}

@article{wright2015coordinate,
  title     = {Coordinate descent algorithms},
  author    = {Wright, Stephen J},
  year      = {2015},
  month     = {3},
  journal   = {Mathematical programming},
  publisher = {Springer},
  volume    = {151},
  number    = {1},
  pages     = {3--34},
  doi       = {10.1007/s10107-015-0892-3}
}

@article{epftoolbox,
  title     = {Forecasting day-ahead electricity prices: A review of state-of-the-art algorithms, best practices and an open-access benchmark},
  author    = {Jesus Lago and Grzegorz Marcjasz and Bart {De Schutter} and Rafał Weron},
  year      = {2021},
  month     = {4},
  journal   = {Applied Energy},
  publisher = {Elsevier BV},
  volume    = {293},
  pages     = {116983},
  doi       = {https://doi.org/10.1016/j.apenergy.2021.116983}
}

@article{dutot2024adaptive,
  title     = {Adaptive probabilistic forecasting of French electricity spot prices},
  author    = {Dutot, Gr{\'e}goire and Zaffran, Margaux and F{\'e}ron, Olivier and Goude, Yannig},
  year      = {2024},
  month     = {5},
  journal   = {arXiv preprint arXiv:2405.15359},
  publisher = {Cornell University}
}

@inproceedings{zaffran2022adaptive,
  title        = {Adaptive conformal predictions for time series},
  author       = {Zaffran, Margaux and F{\'e}ron, Olivier and Goude, Yannig and Josse, Julie and Dieuleveut, Aymeric},
  year         = {2022},
  month        = {2},
  journal      = {arXiv (Cornell University)},
  booktitle    = {International Conference on Machine Learning},
  publisher    = {Cornell University},
  volume       = {162},
  pages        = {25834--25866},
  doi          = {10.48550/arxiv.2202.07282},
  editor       = {Chaudhuri, Kamalika and Jegelka, Stefanie and Le Song and Szepesvári, Csaba and Niu, Gang and Sabato, Sivan},
  organization = {PMLR}
}

@article{wintenberger2024stochastic,
  title     = {Stochastic online convex optimization. Application to probabilistic time series forecasting},
  author    = {Wintenberger, Olivier},
  year      = {2024},
  journal   = {Electronic Journal of Statistics},
  publisher = {The Institute of Mathematical Statistics and the Bernoulli Society},
  volume    = {18},
  number    = {1},
  pages     = {429--464},
  doi       = {10.1214/23-ejs2208}
}

@article{rugamer2024semi,
  title     = {Semi-structured distributional regression},
  author    = {R{\"u}gamer, David and Kolb, Chris and Klein, Nadja},
  year      = {2024},
  journal   = {The American Statistician},
  publisher = {Taylor \& Francis},
  volume    = {78},
  number    = {1},
  pages     = {88--99}
}

@book{francq2019garch,
  title     = {GARCH models: structure, statistical inference and financial applications},
  author    = {Francq, Christian and Zakoian, Jean-Michel},
  year      = {2019},
  month     = {6},
  publisher = {John Wiley \& Sons}
}

@article{brusaferri2024line,
  title     = {On-line conformalized neural networks ensembles for probabilistic forecasting of day-ahead electricity prices},
  author    = {Brusaferri, Alessandro and Ballarino, Andrea and Grossi, Luigi and Laurini, Fabrizio},
  year      = {2024},
  month     = {4},
  journal   = {arXiv preprint arXiv:2404.02722},
  publisher = {Cornell University},
  volume    = {abs/2404.02722},
  doi       = {10.48550/arxiv.2404.02722}
}

@inproceedings{bhatnagar2023improved,
  title        = {Improved online conformal prediction via strongly adaptive online learning},
  author       = {Bhatnagar, Aadyot and Wang, Huan and Xiong, Caiming and Bai, Yu},
  year         = {2023},
  month        = {2},
  journal      = {arXiv (Cornell University)},
  booktitle    = {International Conference on Machine Learning},
  publisher    = {Cornell University},
  volume       = {202},
  pages        = {2337--2363},
  doi          = {10.48550/arxiv.2302.07869},
  editor       = {Krause, Andreas and Brunskill, Emma and Cho, KyungHyun and Engelhardt, Barbara and Sabato, Sivan and Scarlett, Jonathan},
  organization = {PMLR}
}

@article{gibbs2024conformal,
  title   = {Conformal inference for online prediction with arbitrary distribution shifts},
  author  = {Gibbs, Isaac and Cand{\`e}s, Emmanuel J},
  year    = {2024},
  journal = {Journal of Machine Learning Research},
  volume  = {25},
  number  = {162},
  pages   = {1--36}
}

@article{vilmarest2024viking,
  title     = {Viking: variational Bayesian variance tracking},
  author    = {Vilmarest, Joseph de and Wintenberger, Olivier},
  year      = {2024},
  month     = {5},
  journal   = {Statistical Inference for Stochastic Processes},
  publisher = {Springer},
  volume    = {27},
  number    = {3},
  pages     = {1--22},
  doi       = {10.1007/s11203-024-09312-7}
}

@article{kupiec1995techniques,
  title={Techniques for Verifying the Accuracy of Risk Measurement Models},
  author={Kupiec, Paul H},
  journal={The Journal of Derivatives},
  volume={3},
  number={2},
  pages={73--84},
  year={1995},
  publisher={Portfolio Management Research}
}
